%% file: main.tex
\documentclass[10pt,twocolumn,letterpaper]{article}

\usepackage[pagenumbers]{wacv}


\definecolor{wacvblue}{rgb}{0.21,0.49,0.74}
\usepackage[
    pagebackref,
    breaklinks,
    colorlinks,
    allcolors=wacvblue
]{hyperref}

\def\wacvPaperID{}
\def\confName{}
\def\confYear{}

\title{
When Should a Low-Light Pipeline Generate?
Auditable Risk Control for Selective
Generative Enhancement
}

\author{
Qiyi Xie\textsuperscript{1,2},
Wei Chen Ng\textsuperscript{2,3},
Suyang Xi\textsuperscript{4}
Xicheng Wang\textsuperscript{1},
Ziyao Li\textsuperscript{3},
Bohong Xu\textsuperscript{1},
Wenyi Liu\textsuperscript{1},
Jian Ding Tan\textsuperscript{2,3,*}
\\[1mm]
\textsuperscript{1}School of Engineering and Applied Science, University of Pennsylvania\\
\textsuperscript{2}S.M.A.R.T. Nexus Center of Excellence, Xiamen University Malaysia\\
\textsuperscript{3}School of Artificial Intelligence and Robotics, Xiamen University Malaysia\\
\textsuperscript{4}Emory College of Arts and Sciences Computer Science, Emory University\\
\\[1mm]
{\ttfamily\small xieqiyi@engineering.upenn.edu,
mcs2509006@xmu.edu.my,
sxi5@emory.edu,}\\
{\ttfamily\small wangxc6@engineering.upenn.edu, EEE2409020@xmu.edu.my,
xu35@engineering.upenn.edu,}\\
{\ttfamily\small liu50@engineering.upenn.edu, jianding.tan@xmu.edu.my}
\\[1mm]
\textsuperscript{*}Corresponding author
}

\begin{document}

\maketitle

\input{sec/0_abstract}

\input{sec/1_intro}
\input{sec/2_relatedworks}
\input{sec/3_framework}
\input{sec/4_evidence1}
\input{sec/5_evidence2}
\input{sec/6_evidence3}
\input{sec/7_discussion}
\input{sec/8_conclusion}

{
    \small
    \bibliographystyle{ieeenat_fullname}
    \bibliography{main}
}

\end{document}


\maketitle


\section{Reading map, settings, and notation}
\label{supp:readingmap}

This document is the evidence appendix of the submitted paper. Every analysis
in it is derived from artifacts frozen before it was assembled: no additional
model fitting and no final-test rerun were performed, and its tables are
readings, or arithmetic re-analyses, of those frozen records.
Sec.~\ref{supp:artifact} states which of them are released.
Table~\ref{tab:readingmap} is the map; Table~\ref{tab:settings}
decodes the internal setting identifiers and the
symbols the main text leaves unexpanded.

Two conventions hold throughout. A bare ``Sec.~3.4'', ``Table~1'' or
``Fig.~4'' means the submitted paper; parts of this document are cited as
Sec.~\ref{supp:readingmap} and Table~\ref{tab:readingmap}. And the unit of
experiment---one (dataset, expert pair, routing direction) triple with its own
pre-registration, certificate and endpoints---carries three names for one
thing: \emph{setting}, used throughout here and by the main paper in Table~3's
\emph{Setting} column and in the Fig.~4 caption; \emph{dataset line}
(Sec.~3.1; ``per-line'' in Sec.~3.4, ``each line'' in Sec.~3.6); and the key
\texttt{line} in the frozen records. There are eight, in
Table~\ref{tab:settings}(a).

\begin{table*}[t]\centering
\caption{Reading map. Block (A) is every place the submitted paper sends the
reader here, in its own words; block (B) collects the additional analyses this
appendix carries, listed against the main-text anchor that raises each
question. Two entries of block (B), Secs.~\ref{supp:greyzones}
and~\ref{supp:provenance}, appear in block (A) as well, because the pointers of
Sec.~3.5 and Sec.~3.6 land in them without naming them.}
\label{tab:readingmap}
\scriptsize
\setlength{\tabcolsep}{3pt}
\begin{tabular}{@{}p{1.9cm}p{8.1cm}p{5.9cm}@{}}
\toprule
Main-paper anchor & What it promises, or what a reader arrives looking for & Answered in \\
\midrule
\multicolumn{3}{@{}l}{\emph{(A) Where the main text sends the reader to the supplementary}}\\
\midrule
Abstract & ``per-image records and a consumption ledger will be released'' & Sec.~\ref{supp:artifact}; Sec.~\ref{supp:ledger} \\
Sec.~3.4 & ``per-line counts in the supplementary'': harm recurrence in $\lambda$ & Sec.~\ref{supp:rules} \\
Sec.~3.5 & ``details of the retired certificate designs'' & Sec.~\ref{supp:rules}; Sec.~\ref{supp:greyzones}(i) \\
Sec.~3.6 & ``the LOLv2-synthetic deviations are detailed in the supplementary'' & Sec.~\ref{supp:greyzones}(ii); Sec.~\ref{supp:provenance}; Sec.~\ref{supp:premises} \\
Sec.~3.6 & ``the supplementary consumption ledger'', role-membership hashes & Sec.~\ref{supp:ledger} \\
Sec.~4, Fig.~3 & ``protocol in the supplementary'': the local degradation probe & Sec.~\ref{supp:dose} \\
Table~2 caption & ``cue-by-cue tables are in the supplementary'' & Sec.~\ref{supp:premises} \\
Sec.~5, Fig.~4 & ``detailed sweep results'' / ``full sweeps are provided'' & Sec.~\ref{supp:frontier}, Sec.~\ref{supp:frontier_table} \\
Sec.~5 & ``the registered $\varepsilon$ sweep (supplementary)'' & Sec.~\ref{supp:eps} \\
Table~4 caption & ``the full per-condition grid'' & Sec.~\ref{supp:stress} \\
Sec.~6 & ``the full regeneration audit, gate criteria, and amendment history'' & Sec.~\ref{supp:gsad} \\
Sec.~7 & ``same-pipeline comparisons with representative LLIE methods'' & Sec.~\ref{supp:sota}, Sec.~\ref{supp:sota_published} \\
\midrule
\multicolumn{3}{@{}l}{\emph{(B) Additional supporting analyses}}\\
\midrule
Sec.~3.3 & do the router's components earn their place; is it just capacity? & Sec.~\ref{supp:ablation} \\
Sec.~7 & ``requires running both experts, which increases inference cost'' & Sec.~\ref{supp:cost} \\
Sec.~6 & the SICE setting in full: roles, pair selection, endpoint chain & Sec.~\ref{supp:sice} \\
Fig.~6 & further qualitative cases, and how the crops were chosen & Sec.~\ref{supp:cases} \\
Sec.~3.6 & what was registered when; was any endpoint consumed twice? & Sec.~\ref{supp:provenance} \\
Sec.~3.6, Sec.~5 & the construction/validation boundary, setting by setting & Sec.~\ref{supp:greyzones} \\
Table~2 & ``area at which the oracle already wins 2.5\% (saturates near 20\%)'' & Sec.~\ref{supp:premises}; Sec.~\ref{supp:pareto} \\
Sec.~2.1 & no-reference metrics ranking outputs above ground truth & Sec.~\ref{supp:noref} \\
Sec.~1, Sec.~3.1 & why route regions, rather than pick the better whole image? & Sec.~\ref{supp:imgsel} \\
Sec.~7 & ``the main certificate uses only $n=30$ calibration images'' & Sec.~\ref{supp:statscope} \\
Sec.~7 & ``marginally over exchangeable deployment images'': what that assumes here & Sec.~\ref{supp:statscope} \\
Sec.~4, Sec.~5 & raw routing on LOLv2-real validation: $-0.0051$ in one place, $-0.024$ in another & Sec.~\ref{supp:reconcile} \\
Sec.~3.1, Eq.~(1) & colour alignment, halo masks, greedy composition under budget & Sec.~\ref{supp:composite} \\
Sec.~2.1 & ``consider DISTS \dots\ as an alternative'' to LPIPS & Sec.~\ref{supp:dists} \\
\bottomrule
\end{tabular}
\end{table*}

\paragraph{Read the identifier, not the prefix.}
\texttt{v1\_} is a \emph{run-family} prefix, not the LOL-v1 dataset: only
\texttt{v1} and \texttt{v1\_cid} are LOL-v1, \texttt{v1\_syn\_full} is
LOLv2-synthetic, and \texttt{v1\_real\_full} together with all three
\texttt{v1\_cgr} settings are LOLv2-real. Table~3 decodes the expert pairs but
not the datasets, so a reader who takes the prefix for a dataset name puts the
three modern-pair rows on the wrong benchmark. Fig.~4b adds a second scheme,
naming only its two LOL-v1 bars by pair number (``pair~1'' $=$ \texttt{v1},
``pair~2'' $=$ \texttt{v1\_cid}) and its other two by dataset, as all three
series of Fig.~4a are; \texttt{v1\_cid} appears in 4b only. ``Pair~2'' is a
third thing in Sec.~\ref{supp:premises}, where the ``$+3.8\%$ pair~2'' is an
older Retinexformer$+$Diff-Retinex exploration on LOL-v1 (\texttt{v1\_dr}) with
no certificate, endpoint or table row.

\begin{table*}[t]\centering
\caption{Settings and notation. (a)~Every internal setting identifier used here
and in Table~3, with its dataset, its frozen expert pair and where it surfaces
in the submitted paper; the \emph{base} is returned whenever no override is
admitted, the \emph{candidate} supplies the regional overrides
$y_{\mathrm{ins}}$ of Eq.~(1) (\texttt{router/direction.py}), and $R$/$G$ mark
the restoration (deterministic) and generative (stochastic, deployment seed~1)
expert. (b)~Symbols and terms, with the file or script that fixes each.}
\label{tab:settings}
\scriptsize
\setlength{\tabcolsep}{3pt}
\begin{tabular}{@{}p{1.7cm}p{2.2cm}p{1.4cm}p{1.4cm}p{6.3cm}p{2.0cm}@{}}
\multicolumn{6}{@{}l}{\textbf{(a) Settings.} Rf $=$ Retinexformer, DiffRet $=$ Diff-Retinex (main-paper abbreviations)}\\
\toprule
Identifier & Dataset & Base & Candidate & Where it appears in the submitted paper & Final test \\
\midrule
\texttt{v1} & LOL-v1 & DiffLL ($G$) & Rf ($R$) & Tab.~1 row~1; Tab.~3 ``v1''; Fig.~4a,b (``pair~1''); Fig.~6a & once, $n{=}15$ \\
\texttt{v1\_syn\_full} & LOLv2-synthetic & Rf ($R$) & DiffRet ($G$) & Tab.~1 row~2; Tab.~3 ``v1 syn''; Fig.~4a,b; Fig.~6b & once, $n{=}100$ \\
\texttt{v1\_real\_full} & LOLv2-real & DiffRet ($G$) & Rf ($R$) & Tab.~1 row~3; Tab.~3 ``v1 real''; Fig.~4a,b; Secs.~4,~5 (Sec.~\ref{supp:reconcile}) & once, $n{=}100$ \\
\texttt{v1\_cid} & LOL-v1 & CIDNet ($R$) & DiffLL ($G$) & Tab.~3 ``v1 cid''; Fig.~4b (``pair~2''); the LTT repair of Sec.~5 & never consumed \\
\texttt{v1\_cgr} & LOLv2-real & CIDNet ($R$) & GSAD ($G$) & Tab.~3 ``v1 cgr (degraded basin)''; Sec.~4, Sec.~6; Fig.~6c & never consumed \\
\texttt{v1\_cgr\_det} & LOLv2-real & CIDNet ($R$) & GSAD ($G$) & Tab.~3 ``v1 cgr det-path regeneration''; Sec.~4 (``reduced this to 4/65'') & never (basin gate) \\
\texttt{v1\_cgr\_off} & LOLv2-real & GSAD ($G$) & CIDNet ($R$) & Tab.~1 modern-pair row ($\dagger$: disclosed audit outcome, not a selection-independent endpoint); Tab.~3 ``v1 cgr official state''; Secs.~4--6; Fig.~6c & once, $n{=}100$ \\
\texttt{sice} (\texttt{sice\_cd}) & SICE Part2, $-1$EV & CIDNet ($R$) & DiffLL ($G$) & Sec.~5; Sec.~6 terminal test; Fig.~6d & once, $n{=}40$ \\
\bottomrule
\end{tabular}

\vspace{4pt}
\begin{tabular}{@{}p{2.6cm}p{9.6cm}p{3.1cm}@{}}
\multicolumn{3}{@{}l}{\textbf{(b) Notation and terms}}\\
\toprule
Term & Definition as the code and the records fix it & Traceable to \\
\midrule
$r_{H,s}$, and the scoring chain & Per-scale robust normalizer: the median of $|U_{p,s}|$ over the nonzero ($>10^{-8}$) label entries of the training role outside its 30-image internal holdout, fixed once per setting. Training predicts $u_{p,s}=U_{p,s}/r_{H,s}$; deployment inverts it and applies the per-scale isotonic map $g_s$ of the calibration fit half, which returns the score to LPIPS units---so $\lambda$ is in LPIPS units, and a largest knot below $\lambda^\star$ forces a constructive abstention. On \texttt{v1}: $7.93{\times}10^{-4}$, $3.54{\times}10^{-3}$, $5.21{\times}10^{-3}$, $5.61{\times}10^{-3}$ for $s=16,32,64,128$. & \texttt{router/dataset.py} (\texttt{compute\_norms}), \texttt{pipeline.py}; \texttt{router/v1/}\allowbreak\texttt{norms.json} \\
$\tilde\lambda_i$ (safe-suffix score), $\lambda^\star$ & $\tilde\lambda_i$ is the smallest threshold above which calibration image $i$ stays unharmed at every more conservative threshold, and $+\infty$ (stored as \texttt{null}) if none exists; $\lambda^\star$ is the $\lceil(1-\alpha)(n+1)\rceil$-th order statistic of those scores. On \texttt{v1}: $n=30$, one $\tilde\lambda_i=+\infty$, $k=28$, $\lambda^\star=0.01974$. & Eq.~(3); \texttt{certificate\_}\allowbreak\texttt{v2\_v1.json} \\
quality mode $\equiv$ ``raw routing''; certified mode & Table~1's ``Quality-mode gain'' and the ``raw routing'' of Sec.~4 are the same uncertified $\lambda{=}0$ runs---\emph{not} the ``quality gate'' of Table~\ref{tab:premises} and Table~\ref{tab:pareto}, which thresholds the isotonic score at the retired policy-level $\lambda{=}0.0215$. Two documented flags govern the quality mode: the ordering score (raw normalized or isotonic) and a learned verifier filter ($p_{\mathrm{viol}}>0.5$ dropped, disabled where a run had to match its own validation configuration); the records store outcomes, not invocations. The certified mode admits $\tilde U>\lambda^\star$ only, never uses the verifier, and returns the base expert when nothing is admitted. & \texttt{stage4\_final\_test.py} (\texttt{-{}-quality-raw}, \texttt{-{}-no-verifier}); \texttt{stage3\_eval\_}\allowbreak\texttt{router.py} \\
capture & Realized mean gain over the mean oracle headroom at the 20\% budget, same split. Certified capture on \texttt{v1} at $\alpha{=}0.10$ is $0.003157/0.010635=29.7\%$; Table~2's $57.2\%$ is the quality-gate policy just above and $46.6\%$ the raw $>0$ policy, on that same denominator---both are tabulated in Sec.~\ref{supp:premises}. & \texttt{alpha\_utility\_}\allowbreak\texttt{frontier.json}; \texttt{router\_eval\_v1}\allowbreak\texttt{[\_conformal].json} \\
\bottomrule
\end{tabular}
\end{table*}

\section{Released process data}
\label{supp:artifact}

The abstract promises that ``per-image records and a consumption ledger will
be released''. On acceptance we release, in a public repository, the process data behind the numbers reported here: the per-image
records of every consumed final test, the certificates with their calibration
curves, the split manifests with their role-membership hashes, the consumption
ledger of Sec.~\ref{supp:ledger}, and the read-only checker that re-derives the
printed table values from them by arithmetic alone. These are the same
artifacts every table in this supplementary was computed from, so the
re-analyses reported here can be repeated without a GPU. Settings are
identified as in Sec.~\ref{supp:readingmap}; five of the eight reached a final
test and so have a per-image record.

Two things the release does not carry. The benchmark images are referenced by
identifier and are not redistributed, so a third party obtains LOL-v1, LOLv2
and SICE from their original sources under those datasets' own terms; and no
expert weights are included, only the freeze manifests naming each upstream
repository and its pinned commit. Two cautions travel with the records: each is
a single frozen expert realisation---stochastic experts ran at seed~1 for the
final tests, and Sec.~\ref{supp:gsad} documents a process-state hazard under
which even the same nominal seed can land elsewhere---and SICE values are not
comparable with the pixel-paired LOL settings (Sec.~\ref{supp:sice}). 

\section{Same-pipeline SOTA context}
\label{supp:sota}
Table~\ref{tab:sota} is the same-pipeline comparison referenced in the
main text (Sec.~7): official per-dataset weights, official test sets,
full-resolution inference, one metric stack, and no GT-mean brightness
correction for any system. Published GT-mean numbers are transcribed in
Sec.~\ref{supp:sota_published} with their convention labelled.

\begin{table*}[t]\centering
\caption{Same-pipeline SOTA context (official test sets, official
per-dataset weights, no GT-mean for any system, one metric stack;
PSNR$\uparrow$ / SSIM$\uparrow$ / LPIPS$\downarrow$). Context, not
leaderboard: the contribution is the certificate and audit protocol,
which are expert-agnostic. The certified modern-pair output on LOLv2-real
(LPIPS 0.1130, Table~1 of the main text) sits below same-pipeline GSAD.}
\label{tab:sota}
\scriptsize
\setlength{\tabcolsep}{4pt}
\begin{tabular}{lccc}
\toprule
System & LOL-v1 (15) & LOLv2-real (100) & LOLv2-syn (100)\\
\midrule
All-$R$ (Retinexformer~\cite{retinexformer}) & 25.15 / 0.843 / 0.1314 & 22.79 / 0.839 / 0.1707 & 25.67 / 0.928 / 0.0592\\
All-$G$ (DiffLL~\cite{diffll} / Diff-Retinex~\cite{diffretinex}) & 26.32 / 0.844 / 0.1183 & 21.28 / 0.837 / 0.1613 & 22.23 / 0.893 / 0.0898\\
Router, quality & \textbf{26.50} / 0.849 / 0.1118 & 21.60 / 0.838 / 0.1607 & 25.54 / 0.928 / 0.0595\\
Router, certified & 26.36 / 0.846 / 0.1156 & 21.28 / 0.837 / 0.1613 & 25.67 / 0.929 / 0.0578\\
Oracle@20\% (upper bound) & 26.47 / 0.848 / 0.1084 & 21.50 / 0.840 / 0.1492 & 25.69 / 0.930 / 0.0536\\
\midrule
HVI-CIDNet~\cite{cidnet} & 23.81 / \textbf{0.857} / \textbf{0.0857} & \textbf{23.90} / \textbf{0.866} / 0.1219 & 25.13 / \textbf{0.939} / \textbf{0.0450}\\
GSAD~\cite{gsad} & 22.85 / 0.849 / 0.1036 & 20.05 / 0.846 / \textbf{0.1131} & 24.46 / 0.929 / 0.0505\\
ModalFormer~\cite{modalformer} & 23.72 / 0.854 / 0.1147 & 21.25 / 0.849 / 0.1467 & \textbf{25.77} / 0.920 / 0.0761\\
\bottomrule
\end{tabular}
\end{table*}

\section{Published numbers (GT-mean convention)}
\label{supp:sota_published}
Table~\ref{tab:sota_published} transcribes, without re-evaluation, the
numbers that the three modern external methods of Table~\ref{tab:sota}
report in their own papers or official README, with the evaluation
convention each one uses. All three published headline rows apply GT-mean
brightness correction (the output is rescaled to the ground-truth mean
before scoring), which our systems never use and which is unavailable at
deployment because it requires the ground truth; ModalFormer additionally
reports with $\times8$ flip/rotation self-ensembling. HVI-CIDNet's README
also publishes no-GT-mean numbers, which is what makes the bit-exact
pipeline-equivalence check of the main text possible (LOLv2-syn
$25.1294/0.9388/0.0450$ vs.\ our $25.13/0.9388/0.0450$; LOL-v1 agrees to
rounding except for $9\times10^{-4}$ in SSIM; LOLv2-real, bestPSNR
weights, is $-0.21$\,dB PSNR against ours and is recorded as such). The reading: the
3--5\,dB PSNR gap between this table and Table~\ref{tab:sota} is the
convention, not the models, and the two tables must not be mixed.

\begin{table*}[t]\centering
\caption{Published numbers of the external methods under their own
evaluation conventions (PSNR$\uparrow$ / SSIM$\uparrow$ / LPIPS$\downarrow$;
transcribed, not re-run). GT-mean = output brightness rescaled to the
ground-truth mean before scoring; TTA = $\times$8 flip/rotation
self-ensemble; Res.\ = evaluation resolution as stated by the method
(not recorded in our transcription). `---' = not reported by the method
or not transcribed in our records. Compare with the no-GT-mean same-pipeline
numbers of Table~\ref{tab:sota}.}
\label{tab:sota_published}
\scriptsize
\setlength{\tabcolsep}{3pt}
\begin{tabular}{llcccll}
\toprule
Method & Source & LOL-v1 & LOLv2-real & LOLv2-syn & GT-mean & Res.\\
\midrule
HVI-CIDNet~\cite{cidnet} & paper, best variant per dataset & 27.71 / 0.876 / 0.0791 & 28.14 / 0.892 / 0.1008 & 29.37 / 0.950 / 0.0403 & yes & ---\\
HVI-CIDNet~\cite{cidnet} & README (w\_perc / bestPSNR / w\_perc) & 23.8091 / 0.8574 / 0.0856 & 24.1106 / 0.8675 / 0.1162 & 25.1294 / 0.9388 / 0.0450 & no & ---\\
GSAD~\cite{gsad} & paper, Table~1 & 27.839 / 0.877 / 0.091 & 28.818 / 0.895 / 0.095 & 28.670 / 0.944 / 0.047 & yes & ---\\
ModalFormer~\cite{modalformer} & paper, Table~1 & 27.97 / 0.897 / --- & 29.33 / 0.915 / --- & 30.15 / 0.951 / --- & yes, $+$$\times$8 TTA & ---\\
\bottomrule
\end{tabular}
\end{table*}

\paragraph{The experts of Table~\ref{tab:sota}.}
Table~\ref{tab:sota_published_experts} repeats the transcription for the
three of the five frozen experts that serve as a fixed base or candidate in
Table~\ref{tab:sota}, so that the convention argument covers the systems
those rows are built on and not only the modern external ones. Retinexformer settles the question on its own, because its
official README publishes both conventions for the same weights: the
GT-mean row sits $+2.02$, $+4.91$ and $+3.37$\,dB above the no-GT-mean row
on LOL-v1, LOLv2-real and LOLv2-syn respectively. The several-dB step is
therefore measurable inside a single frozen model, with no comparison
across models and no appeal to our own pipeline required.

\paragraph{A second pipeline-equivalence check.}
The no-GT-mean row also provides a check independent of the HVI-CIDNet one
above, and on an expert rather than an external baseline. Retinexformer's
published no-GT-mean PSNR is $25.16 / 22.80 / 25.67$ against the
$25.15 / 22.79 / 25.67$ our pipeline measures in Table~\ref{tab:sota}:
within $0.01$\,dB on all three settings, with SSIM within $0.002$. DiffLL's
published LOL-v1 PSNR, $26.336$, likewise sits $0.02$\,dB from the $26.32$
the All-$G$ entry records on that setting, where the generative expert is
DiffLL. This is also why we label the first row as carrying no GT-mean: the
README introduces the second table as a separate test setting that follows
KinD, LLFlow and the diffusion models and rescales the output to the mean of
the ground truth, and a no-GT-mean pipeline reproduces the first table but
not the second.

\paragraph{What is missing, and what must not be compared.}
Three cells are left empty rather than filled. Diff-Retinex scores LOL-v1
only; the LOLv2 weights we run are the authors' later release, described in
their repository as extended weights for further research and never scored
in the paper. DiffLL reports no LOLv2-syn. And DiffLL's published LOLv2-real
PSNR, $28.857$, has no counterpart in Table~\ref{tab:sota}, because the
generative entry on that setting is Diff-Retinex and not DiffLL; we never
ran DiffLL on that setting and so cannot check that value here, and it must
not be read against our LOLv2-real column. The LPIPS column is likewise not
on a common footing: on the same LOL-v1 test set DiffLL prints $0.217$ and
Diff-Retinex $0.048$, a factor of four apart, and neither source states its
backbone network. Only PSNR and SSIM carry the convention comparison; the
LPIPS values are transcribed because they are published, not because they
are comparable.

\begin{table*}[t]\centering
\caption{Published numbers of the three frozen experts appearing in
Table~\ref{tab:sota}, under their own evaluation conventions (PSNR$\uparrow$ / SSIM$\uparrow$ / LPIPS$\downarrow$;
transcribed, not re-run). Retinexformer publishes both conventions for the
same weights: the first row is read from the results figure of its official
README and the second from that README's own table for the setting it labels
as following KinD, LLFlow and the diffusion models, which rescales the output
to the ground-truth mean. Diff-Retinex's PSNR and SSIM are its Table~2 and its
LPIPS its Table~1; its LOL-v1 test split is the 15-image one. TTA = self-ensemble; `opt-in, not passed' = the official
repository offers self-ensemble as a command-line flag that the commands
given for these numbers do not pass; `none in code' = the official
evaluation script contains no self-ensemble path. `---' = not reported by
the method or not transcribed in our records. The LPIPS column is not
comparable across rows (see text). Compare with the no-GT-mean same-pipeline
numbers of Table~\ref{tab:sota}.}
\label{tab:sota_published_experts}
\scriptsize
\setlength{\tabcolsep}{3pt}
\begin{tabular}{llcccll}
\toprule
Method & Source & LOL-v1 & LOLv2-real & LOLv2-syn & GT-mean & TTA\\
\midrule
Retinexformer~\cite{retinexformer} & README results figure & 25.16 / 0.845 / --- & 22.80 / 0.840 / --- & 25.67 / 0.930 / --- & no & opt-in, not passed\\
Retinexformer~\cite{retinexformer} & README, KinD/LLFlow setting & 27.18 / 0.850 / --- & 27.71 / 0.856 / --- & 29.04 / 0.939 / --- & yes & opt-in, not passed\\
DiffLL~\cite{diffll} & paper, Table~1 & 26.336 / 0.845 / 0.217 & 28.857 / 0.876 / 0.207 & --- & not stated & none in code\\
Diff-Retinex~\cite{diffretinex} & paper, Table~2 / Table~1 & 21.98 / 0.863 / 0.048 & --- & --- & not stated & none in code\\
\bottomrule
\end{tabular}
\end{table*}


\section{Whole-image expert selection}
\label{supp:imgsel}

\paragraph{The construction.}
The whole-image alternative to regional routing---take the better expert
for the image as a whole---is the final-test script's own pre-registered
bound: lines 220--222 of \path{scripts/stage4_final_test.py}, whose
comment marks it so, take \texttt{all\_g} when its whole-image $\lpips$
against the ground truth is $\le$ that of \texttt{all\_r}, and line 252
emits it unconditionally. The two newest final tests store the block;
the three earlier ones predate that line, so we recompute those
by the identical rule---arithmetic over the frozen per-image records
(Sec.~\ref{supp:artifact}), no re-inference---which reproduces both
stored blocks exactly (\texttt{picked} included, 140/140 images).
\emph{Whole-image} here is a selection granularity, not the main text's
\emph{image-level} harm control.

\paragraph{Neither granularity dominates.}
Table~\ref{tab:imgsel} sets that pick beside the fixed experts, the
certified router and the window oracle at a 20\% area budget; every cell
is a stored \texttt{table} block bar the three recomputed picks, shared
entries match Table~\ref{tab:sota}, and fixed base minus certified
gives main-text Table~1's certified-gain column to rounding. The oracle
holds the better bound on two settings, the pick on three; they are not
nested---the oracle stops at the first greedy step past a 20\% area
budget, and edits only feathered windows from a fixed multi-scale grid,
while the pick edits 0\% or 100\%---so the crossing is expected. The certificate itself
tracks that split. It opens override area on many images in exactly the
two settings where the oracle holds the larger bound: LOL-v1
(11/15 images touched, certified gain $+0.00275$) and LOLv2-synthetic
(25/100, $+0.00140$); areas are in Sec.~\ref{supp:composite}, touched
and harmed counts in Sec.~\ref{supp:statscope}. Where the pick dominates
it either abstains---on LOLv2-real and SICE the certified area is $0$ on
every image and the gain exactly zero, CI $[0,0]$, output bit-identical
to the base---or barely intervenes: 8/100 images on the modern pair,
$+0.00023$ $[+0.00007,+0.00044]$.

\paragraph{An oracle, and its proxy is fidelity-blind.}
The pick is computed \emph{against the ground truth}---an unattainable
bound, not a policy we could have run. It is also unflattering: it
beats the deployed certified router on all five
settings, by $+0.0033$, $+0.0026$, $+0.0134$, $+0.0128$ and $+0.0229$
$\lpips$, with the certified output better on only 40\%, 11\%, 0\%, 4\%
and 0\% of images. Its only deployable proxy is a no-reference score, shown
fidelity-blind in Sec.~\ref{supp:noref}. Table~\ref{tab:imgsel_nr}
repeats the construction with each of six frozen no-reference metrics in
place of the ground truth (each metric's own better direction; ties to
All-$G$, as at line 221). Seventeen of the eighteen selectors fail to
improve on the certified router---three tie it exactly, being constant
on LOLv2-real---and on LOLv2-synthetic all
six are worse than the fixed base ($0.0592$), from NIQE $0.0654$ to
MANIQA $0.0850$. On LOL-v1, BRISQUE ($0.1170$) and MUSIQ ($0.1177$) do
beat the fixed base ($0.1183$) but not the certified router ($0.1156$);
three of the eighteen beat the base in all. The one improvement on the
certified router is NIMA on LOLv2-real ($0.1604$ vs.\ $0.1613$), the
panel's one abstaining setting. None was pre-registered: point estimates
only.

The reading: the three settings where the swap beats the window oracle
are those where the certificate, seeing no ground truth, either refuses
to route at all (LOLv2-real, SICE) or opens area on 8 of 100 images
(modern pair).

\begin{table*}[t]\centering
\caption{Whole-image expert selection against regional routing on the
five once-only final tests (PSNR$\uparrow$ / SSIM$\uparrow$ /
LPIPS$\downarrow$; all values $D_{\mathrm{test}}$, unlike the validation
premises of Table~2). $\dagger$ = computed against the ground truth, an
unattainable bound and not a deployable policy; the pick minimises
whole-image LPIPS, so its PSNR and SSIM are carried along, not
optimised. Bold marks the better of the two $\dagger$ bounds. The
modern-pair row is main-text Table~1's disclosed audit setting---its
official test set informed expert-pair selection---and the SICE row
restates the selector of Sec.~\ref{supp:sice}. Rf:
Retinexformer~\cite{retinexformer}; DiffRet: Diff-Retinex.}
\label{tab:imgsel}
\scriptsize
\setlength{\tabcolsep}{4pt}
\begin{tabular}{llcccccc}
\toprule
Setting ($n$) & Metric & All-$R$ & All-$G$ & Fixed base & Whole-image pick$^\dagger$ & Router, certified & Oracle@20\%$^\dagger$\\
\midrule
LOL-v1 (15) & PSNR & 25.15 & 26.32 & 26.32 & 26.69 & 26.36 & 26.47\\
base $G$: DiffLL~\cite{diffll}$|$Rf & SSIM & 0.843 & 0.844 & 0.844 & 0.850 & 0.846 & 0.848\\
 & LPIPS & 0.1314 & 0.1183 & 0.1183 & 0.1123 & 0.1156 & \textbf{0.1084}\\
\midrule
LOLv2-syn (100) & PSNR & 25.67 & 22.23 & 25.67 & 25.89 & 25.67 & 25.69\\
base $R$: Rf$|$DiffRet~\cite{diffretinex} & SSIM & 0.928 & 0.893 & 0.928 & 0.930 & 0.929 & 0.930\\
 & LPIPS & 0.0592 & 0.0898 & 0.0592 & 0.0553 & 0.0578 & \textbf{0.0536}\\
\midrule
LOLv2-real (100) & PSNR & 22.79 & 21.28 & 21.28 & 22.50 & 21.28 & 21.50\\
base $G$: DiffRet~\cite{diffretinex}$|$Rf & SSIM & 0.839 & 0.837 & 0.837 & 0.845 & 0.837 & 0.840\\
 & LPIPS & 0.1707 & 0.1613 & 0.1613 & \textbf{0.1479} & 0.1613 & 0.1492\\
\midrule
LOLv2-real, modern pair (100) & PSNR & 23.89 & 19.98 & 19.98 & 23.11 & 19.98 & 20.00\\
base $G$: GSAD~\cite{gsad}$|$CIDNet~\cite{cidnet} & SSIM & 0.866 & 0.845 & 0.845 & 0.869 & 0.845 & 0.847\\
 & LPIPS & 0.1220 & 0.1132 & 0.1132 & \textbf{0.1002} & 0.1130 & 0.1104\\
\midrule
SICE $-1$EV (40) & PSNR & 15.92 & 16.20 & 15.92 & 16.56 & 15.92 & 15.97\\
base $R$: CIDNet~\cite{cidnet}$|$DiffLL~\cite{diffll} & SSIM & 0.520 & 0.523 & 0.520 & 0.534 & 0.520 & 0.523\\
 & LPIPS & 0.2534 & 0.2605 & 0.2534 & \textbf{0.2305} & 0.2534 & 0.2390\\
\bottomrule
\end{tabular}
\end{table*}

\begin{table}[t]\centering
\caption{Deployable whole-image selectors: Table~\ref{tab:imgsel}'s
construction with each frozen no-reference metric in place of the ground
truth, scored by the resulting mean LPIPS$\downarrow$ (parenthesis:
overrides of the fixed base expert). \texttt{noref\_panel\_*.json}
exists for these three settings only.}
\label{tab:imgsel_nr}
\scriptsize
\setlength{\tabcolsep}{2.5pt}
\begin{tabular}{lccc}
\toprule
Selector & LOL-v1 (15) & LOLv2-real (100) & LOLv2-syn (100)\\
\midrule
Fixed base expert & 0.1183 & 0.1613 & 0.0592\\
Router, certified & 0.1156 & 0.1613 & 0.0578\\
\midrule
MUSIQ & 0.1177 (2) & 0.1613 (0) & 0.0792 (80)\\
NIQE & 0.1314 (15) & 0.1716 (91) & 0.0654 (16)\\
BRISQUE & 0.1170 (6) & 0.1637 (93) & 0.0676 (38)\\
MANIQA & 0.1187 (2) & 0.1613 (0) & 0.0850 (92)\\
CLIP-IQA+ & 0.1218 (6) & 0.1613 (0) & 0.0784 (74)\\
NIMA & 0.1237 (7) & 0.1604 (3) & 0.0828 (83)\\
\midrule
Ground-truth pick$^\dagger$ & 0.1123 & 0.1479 & 0.0553\\
\bottomrule
\end{tabular}
\end{table}

\section{Deployed-rule $\alpha$--utility frontier}
\label{supp:frontier}
\begin{figure*}[t]
  \centering
  \includegraphics[width=\textwidth]{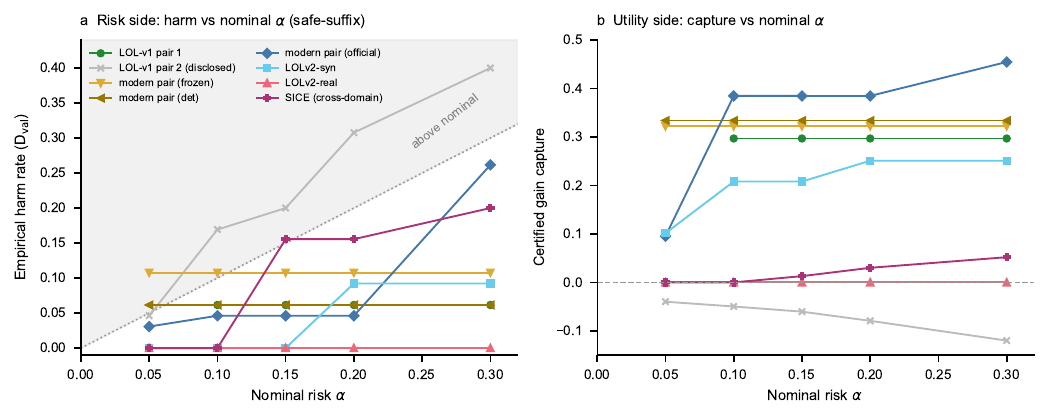}
  \caption{Deployed-rule $\alpha$--utility frontier, all eight settings
  ($D_{\mathrm{val}}$, $\varepsilon=0$): empirical harm (left) and
  certified capture (right) vs.\ nominal risk under the safe-suffix rule.}
  \label{fig:frontier}
\end{figure*}

Figure~\ref{fig:frontier} and Table~\ref{tab:frontier} give the full
eight-setting, five-$\alpha$ grid under both rules (safe-suffix and valid
LTT), recomputed from the frozen certificates with zero refitting
(\path{outputs/logs/stage3/alpha_utility_frontier.json}). The 8 of 39
evaluable cells whose point estimate exceeds nominal are v1\_cid at
$\alpha\in\{0.10,0.15,0.20,0.30\}$ (16.9/20.0/30.8/40.0\%), v1\_cgr at
$\alpha\in\{0.05,0.10\}$ (10.8\%), v1\_cgr-det at $\alpha=0.05$ (6.2\%) and
SICE at $\alpha=0.15$ (7/45 $=$ 15.6\%); every exceedance lies within its
Clopper--Pearson 95\% interval of nominal. Were the bound exactly tight,
18.2 exceedances would be expected over the 39 cells---a magnitude
benchmark rather than a test, since cells within a setting share
images---and none of the four LOL-family settings carrying a consumed final
test exceeds at any $\alpha$. v1\_cid's capture at the
valid-LTT point is $-0.040$ (clipped to 0 in Fig.~3b of the main paper);
its thin mean headroom on $D_{\mathrm{val}}$ (${\approx}{+}0.005$) goes
unrealized. The two abstentions are constructive: the deployed isotonic
score clamps at its largest knot, and that ceiling lies below the
certified threshold in exactly two settings ($0.00472<\lambda^\star{=}0.00563$
on LOLv2-real; $0.04289<0.04307$ on SICE), so no override window is
admissible at $\alpha=0.10$.

\section{Deployed-rule $\alpha$--utility frontier, full table}
\label{supp:frontier_table}
Table~\ref{tab:frontier} is the full grid behind Fig.~\ref{fig:frontier}:
eight settings $\times$ five nominal risk levels $\alpha$, under the deployed
safe-suffix rule and the valid Learn-then-Test (LTT) rule, read off the
frozen per-$\alpha$ $\lambda^\star$ and the frozen $D_{\mathrm{val}}$
curves with zero refitting
(\path{outputs/logs/stage3/alpha_utility_frontier.md}). Harm is the
$D_{\mathrm{val}}$ count of images where the routed output is worse than
the fixed expert ($\varepsilon=0$), capture is the certified share of the
oracle gain, and mean gain is the mean per-image LPIPS improvement over the
fixed expert; ``abstains'' means no $\lambda^\star$ can be certified at
that $\alpha$, so the routed system equals the fixed expert (harm 0,
capture 0). The reading: valid LTT abstains at $\alpha=0.05$ on all
seven LOL-family settings (SICE certifies under both rules) and, wherever it
certifies, is never less conservative than safe-suffix
($\lambda^\star_{\mathrm{LTT}}\ge\lambda^\star_{\mathrm{ss}}$ in every
cell); v1\_cid is the one setting whose safe-suffix validation harm exceeds
nominal at every $\alpha\ge0.10$ (up to $26/65$ at $\alpha=0.30$) with
negative capture throughout, while its LTT certificate stays at or below
nominal for $\alpha\le0.20$.

\begin{table*}[t]\centering
\caption{Deployed-rule $\alpha$--utility frontier, all eight settings and
five $\alpha$ levels ($D_{\mathrm{val}}$, $\varepsilon=0$). Safe-suffix
(ss) is the deployed rule; valid LTT is the auxiliary certificate. Harm is
given as count/$n$ (\%). Negative capture on v1\_cid is shown unclipped
(Fig.~3b of the main paper clips it to 0).}
\label{tab:frontier}
\scriptsize
\setlength{\tabcolsep}{3pt}
\begin{tabular}{llcccccc}
\toprule
Setting & $\alpha$ & $\lambda^\star$ (ss) & harm (ss) & capture (ss) & mean gain (ss) & $\lambda^\star$ (LTT) & harm (LTT)\\
\midrule
v1 (Rf+DiffLL) & 0.05 & abstains & --- & --- & --- & abstains & ---\\
 & 0.10 & 0.01974 & 4/65 (6.2) & 0.297 & $+$0.00316 & abstains & ---\\
 & 0.15 & 0.01974 & 4/65 (6.2) & 0.297 & $+$0.00316 & 0.02359 & 1/65 (1.5)\\
 & 0.20 & 0.01974 & 4/65 (6.2) & 0.297 & $+$0.00316 & 0.01974 & 4/65 (6.2)\\
 & 0.30 & 0.01974 & 4/65 (6.2) & 0.297 & $+$0.00316 & 0.01974 & 4/65 (6.2)\\
\midrule
v1\_cid (CIDNet+DiffLL) & 0.05 & 0.01717 & 3/65 (4.6) & $-$0.040 & $-$0.00018 & abstains & ---\\
 & 0.10 & 0.01333 & 11/65 (16.9) & $-$0.050 & $-$0.00023 & 0.01717 & 3/65 (4.6)\\
 & 0.15 & 0.01076 & 13/65 (20.0) & $-$0.060 & $-$0.00028 & 0.01717 & 3/65 (4.6)\\
 & 0.20 & 0.0082 & 20/65 (30.8) & $-$0.079 & $-$0.00036 & 0.01333 & 11/65 (16.9)\\
 & 0.30 & 0.00435 & 26/65 (40.0) & $-$0.120 & $-$0.00055 & 0.00563 & 23/65 (35.4)\\
\midrule
v1\_cgr (degraded basin) & 0.05 & 0.00691 & 7/65 (10.8) & 0.322 & $+$0.00145 & abstains & ---\\
 & 0.10 & 0.00691 & 7/65 (10.8) & 0.322 & $+$0.00145 & 0.00691 & 7/65 (10.8)\\
 & 0.15 & 0.00691 & 7/65 (10.8) & 0.322 & $+$0.00145 & 0.00691 & 7/65 (10.8)\\
 & 0.20 & 0.00691 & 7/65 (10.8) & 0.322 & $+$0.00145 & 0.00691 & 7/65 (10.8)\\
 & 0.30 & 0.00691 & 7/65 (10.8) & 0.322 & $+$0.00145 & 0.00691 & 7/65 (10.8)\\
\midrule
v1\_cgr det-path & 0.05 & 0.00691 & 4/65 (6.2) & 0.334 & $+$0.00151 & abstains & ---\\
 & 0.10 & 0.00691 & 4/65 (6.2) & 0.334 & $+$0.00151 & 0.00691 & 4/65 (6.2)\\
 & 0.15 & 0.00691 & 4/65 (6.2) & 0.334 & $+$0.00151 & 0.00691 & 4/65 (6.2)\\
 & 0.20 & 0.00691 & 4/65 (6.2) & 0.334 & $+$0.00151 & 0.00691 & 4/65 (6.2)\\
 & 0.30 & 0.00691 & 4/65 (6.2) & 0.334 & $+$0.00151 & 0.00691 & 4/65 (6.2)\\
\midrule
v1\_cgr official basin & 0.05 & 0.01846 & 2/65 (3.1) & 0.095 & $+$0.00074 & abstains & ---\\
 & 0.10 & 0.01333 & 3/65 (4.6) & 0.385 & $+$0.00298 & 0.01846 & 2/65 (3.1)\\
 & 0.15 & 0.01204 & 3/65 (4.6) & 0.385 & $+$0.00298 & 0.01846 & 2/65 (3.1)\\
 & 0.20 & 0.01204 & 3/65 (4.6) & 0.385 & $+$0.00298 & 0.01333 & 3/65 (4.6)\\
 & 0.30 & 0.01076 & 17/65 (26.2) & 0.455 & $+$0.00352 & 0.01204 & 3/65 (4.6)\\
\midrule
v1\_syn (Rf+DiffRet) & 0.05 & 0.01717 & 0/65 (0.0) & 0.101 & $+$0.00066 & abstains & ---\\
 & 0.10 & 0.01333 & 0/65 (0.0) & 0.208 & $+$0.00136 & 0.01717 & 0/65 (0.0)\\
 & 0.15 & 0.01333 & 0/65 (0.0) & 0.208 & $+$0.00136 & 0.01717 & 0/65 (0.0)\\
 & 0.20 & 0.00563 & 6/65 (9.2) & 0.251 & $+$0.00164 & 0.01333 & 0/65 (0.0)\\
 & 0.30 & 0.00563 & 6/65 (9.2) & 0.251 & $+$0.00164 & 0.00563 & 6/65 (9.2)\\
\midrule
v1\_real (Rf+DiffRet) & 0.05 & 0.00563 & 0/65 (0.0) & 0.000 & $+$0.00000 & abstains & ---\\
 & 0.10 & 0.00563 & 0/65 (0.0) & 0.000 & $+$0.00000 & 0.00563 & 0/65 (0.0)\\
 & 0.15 & 0.00563 & 0/65 (0.0) & 0.000 & $+$0.00000 & 0.00563 & 0/65 (0.0)\\
 & 0.20 & 0.00563 & 0/65 (0.0) & 0.000 & $+$0.00000 & 0.00563 & 0/65 (0.0)\\
 & 0.30 & 0.00563 & 0/65 (0.0) & 0.000 & $+$0.00000 & 0.00563 & 0/65 (0.0)\\
\midrule
SICE (CIDNet+DiffLL) & 0.05 & 0.04307 & 0/45 (0.0) & 0.000 & $+$0.00000 & 0.04307 & 0/45 (0.0)\\
 & 0.10 & 0.04307 & 0/45 (0.0) & 0.000 & $+$0.00000 & 0.04307 & 0/45 (0.0)\\
 & 0.15 & 0.0392 & 7/45 (15.6) & 0.013 & $+$0.00018 & 0.04307 & 0/45 (0.0)\\
 & 0.20 & 0.02759 & 7/45 (15.6) & 0.030 & $+$0.00041 & 0.0392 & 7/45 (15.6)\\
 & 0.30 & 0.01598 & 9/45 (20.0) & 0.052 & $+$0.00071 & 0.01985 & 6/45 (13.3)\\
\bottomrule
\end{tabular}
\end{table*}


\section{Operating-point reconciliation on LOLv2-real}
\label{supp:reconcile}

The main text gives two validation numbers for the same 65 LOLv2-real
images that differ by a factor of $4.6$: raw routing ``already showed
significant harm on validation ($-0.0051$, 95\% CI $[-0.0069,-0.0035]$)''
(Sec.~4), and ``every threshold that allows routing is harmful on
validation (61/65 images; mean gain $-0.024$)'' (Sec.~5). They are not two
estimates of one quantity but two \emph{policies} on one frozen image set,
and the gap is why the certificate abstains here. Records below are under
\path{outputs/logs/stage3/}.

The $-0.0051$ policy is the uncalibrated router: raw per-window scores,
admit everything above $0$, fill the 20\% area budget greedily in score
order (\path{router_eval_v1_real_full_raw_diff_retinex.json}, the row of
Table~\ref{tab:premisescross}). The $-0.024$ policy applies the isotonic
deployment map to the same predictions and admits every window whose
calibrated score exceeds a candidate $\lambda$. Of its 25 grid points in
$[0,0.03]$ only five admit anything: the isotonic map clamps at its largest
knot, so no calibrated score can exceed $0.004715$ while
$\lambda^\star=0.00563$. The ceiling audit flags the setting
\path{infeasible_by_construction}; Sec.~\ref{supp:frontier} states this as
$0.00472<\lambda^\star$.

Recomputing all 25 points from the frozen per-image curves
(\path{certificate_v2_v1_real_full.json}, \path{val_curves_full_grid})
gives Table~\ref{tab:reconcile}.
The five activatable points ($\lambda\le0.00435$) are numerically identical
on all 65 images --- 61 harmed, mean gain $-0.023566$, mean $\lpips$
$0.167673$, rounding to the main text's $-0.024$ and $61/65$ --- because
the budget binds before the threshold does. The curves carry $\lpips$ only,
but this policy's coverage is in the budget sweep
(\path{budget_pareto_v1_real_full.json}, \path{final_area}), whose
per-image $\lpips$ at a 20\% budget is identical to the $\lambda{=}0$ point
on all 65: mean realized area $0.19999$, range $0.19973$--$0.20000$.

\begin{table*}[t]\centering
\caption{The two LOLv2-real validation operating points on one frozen image
set ($D_\mathrm{val}$, $n{=}65$, Diff-Retinex base with Retinexformer
overrides, 20\% area budget, $\lpips\downarrow$). ``Activated'': images
receiving at least one override window; ``harmed'': routed output worse
than the fixed expert (0.14411). Gains are paired means over all 65 images
(paired-bootstrap 95\% CIs, $20{,}000$ resamples); area is realized union
coverage (Sec.~\ref{supp:composite}). Row~2 covers the five grid points
below the isotonic ceiling $0.004715$, all numerically identical; row~3,
the twenty at or above $\lambda^\star$, which all return the base image.}
\label{tab:reconcile}
\scriptsize
\setlength{\tabcolsep}{4pt}
\begin{tabular}{llccllc}
\toprule
Policy & Threshold & Activated & Harmed & Mean gain vs.\ fixed (CI) & Mean $\lpips$ & Mean area\\
\midrule
Raw score, no isotonic (main text Sec.~4) & $>0$ (raw) & 52/65 & 45/65 & $-0.0051$ $[-0.0069,-0.0035]$ & 0.14923 & 0.1443\\
Isotonic score, any activatable point & $\lambda\le0.00435$ (5 pts) & 65/65 & 61/65 & $-0.0236$ $[-0.0295,-0.0179]$ & 0.16767 & 0.19999\\
Isotonic score, deployed certificate & $\lambda^\star=0.00563$ & 0/65 & 0/65 & $0.0000$ $[0.0000,0.0000]$ & 0.14411 & 0.0000\\
\bottomrule
\end{tabular}
\end{table*}

The reading: nothing lies between touching little and losing a little and
touching everything and losing a lot. The raw policy declines 13 of the 65
images, covers 14.4\% of pixels and still loses $0.0051$; every threshold
the certificate can score activates all 65 at $20.0\%$ and loses $0.0236$
--- harm grows $4.6\times$ where coverage grows $1.4\times$. The
certificate refuses both, because its risk rule lives on the calibrated
scale, which admits no threshold here. And since it admits nothing, the
LOLv2-real row of Table~1 is the fixed expert by construction: certified
harm exactly $0.0\%$, certified gain printed ``0 (abstains)'', the frozen
record carrying that interval as exactly $[0,0]$, not a narrow interval
near it.


\section{Calibration-rule baselines and harm non-monotonicity}
\label{supp:rules}
Table~\ref{tab:rules} reports the six-rule comparison at $\alpha=0.10$
(selection on the conformal calibration half only; evaluation on
$D_{\mathrm{val}}$), recomputed from the frozen certificate scores
(\path{outputs/logs/stage3/rule_baselines_nonmono.json}; 96/96
built-in consistency checks reproduce the deployed thresholds bit-exactly).
Per-image harm is non-monotone in $\lambda$ (a $0\to1$ recurrence after a
first unharmed grid point) on 4/30 LOL-v1, 3/30 v1\_cid, 2/30 v1\_cgr-off
and 22/76 SICE calibration images (validation: 7/65, 9/65, 3/65, 13/45;
v1\_cgr 4/65, v1\_cgr-det 3/65, LOLv2-syn 1/65, LOLv2-real 0/65). On LOL-v1
the four recurring calibration images are exactly those whose
$\tilde\lambda_i$ equals the deployed $\lambda^\star=0.01974$, and the
largest post-safe excess is $2.8\times$ the DiffLL seed-noise floor
($0.00116$ vs.\ $0.00042$). On SICE the overrides released by the naive
rules buy essentially nothing (mean gains $-0.0001$ for plug-in and
$+0.0008$ for first-safe conformal) while harming 11--13\% of validation
images. Valid LTT abstains on LOL-v1 at
$\delta=0.05$: even the most conservative grid point harms one
calibration image, so no point passes the test ($p=0.184>\delta$); at
$\alpha=0.15$ the walk certifies $\lambda^\star=0.0236$.

\begin{table*}[t]\centering
\caption{Six calibration rules at $\alpha=0.10$: $\lambda^\star$\,/\,
$D_{\mathrm{val}}$ harm count (rate). ``first-safe scan'' = first grid
point with zero calibration harm; ``plug-in'' = least conservative grid
point with empirical calibration harm $\le\alpha$; ``retired'' = the
$(V{+}1)/(N{+}1)$ scan retired by the audit; ``first-safe conformal'' =
conformal quantile over first-unharmed scores; safe-suffix = deployed;
LTT = valid Learn-then-Test walk. Where harm is monotone the rules land on
the same or adjacent grid points with identical validation harm.}
\label{tab:rules}
\scriptsize
\setlength{\tabcolsep}{2.5pt}
\begin{tabular}{lcccccc}
\toprule
Setting & first-safe scan & plug-in & retired & first-safe conformal & safe-suffix & valid LTT\\
\midrule
v1 (Rf+DiffLL) & abstains & 0.0197\,/\,4/65 & 0.0197\,/\,4/65 & 0.0197\,/\,4/65 & 0.0197\,/\,4/65 & abstains\\
v1\_syn (Rf+DiffRet) & 0.0172\,/\,0/65 & 0.0133\,/\,0/65 & 0.0133\,/\,0/65 & 0.0133\,/\,0/65 & 0.0133\,/\,0/65 & 0.0172\,/\,0/65\\
v1\_real (Rf+DiffRet) & 0.0056\,/\,0/65 & 0.0056\,/\,0/65 & 0.0056\,/\,0/65 & 0.0056\,/\,0/65 & 0.0056\,/\,0/65 & 0.0056\,/\,0/65\\
v1\_cid (CIDNet+DiffLL) & 0.0172\,/\,3/65 & \textbf{0.0082\,/\,20/65 (30.8\%)} & 0.0133\,/\,11/65 (16.9\%) & \textbf{0.0108\,/\,13/65 (20.0\%)} & 0.0133\,/\,11/65 (16.9\%) & \textbf{0.0172\,/\,3/65 (4.6\%)}\\
v1\_cgr (degraded basin) & 0.0069\,/\,7/65 & 0.0069\,/\,7/65 & 0.0069\,/\,7/65 & 0.0069\,/\,7/65 & 0.0069\,/\,7/65 & 0.0069\,/\,7/65\\
v1\_cgr det-path & 0.0069\,/\,4/65 & 0.0069\,/\,4/65 & 0.0069\,/\,4/65 & 0.0069\,/\,4/65 & 0.0069\,/\,4/65 & 0.0069\,/\,4/65\\
v1\_cgr official basin & 0.0185\,/\,2/65 & 0.0120\,/\,3/65 & 0.0133\,/\,3/65 & 0.0120\,/\,3/65 & 0.0133\,/\,3/65 & 0.0185\,/\,2/65\\
SICE (CIDNet+DiffLL) & 0.0431\,/\,0/45 & \textbf{0.0411\,/\,5/45 (11.1\%)} & 0.0431\,/\,0/45 & \textbf{0.0199\,/\,6/45 (13.3\%)} & 0.0431\,/\,0/45 & 0.0431\,/\,0/45\\
\bottomrule
\end{tabular}
\end{table*}

\section{$\varepsilon$-Sensitivity of the Certificate}
\label{supp:eps}

The main tables use the strict harm event ($\varepsilon=0$: routed LPIPS
above the fixed expert's by any amount). Pre-registered amendment A2
re-reads the frozen $D_{\mathrm{val}}$ calibration curves ($n=65$,
$\alpha=0.10$, $\delta=0.05$) with $\varepsilon$ set, per setting, to the
stochastic expert's own seed-to-seed noise floor (p95 $|\Delta\lpips|$
between seeds on the first half of $D_{\mathrm{cal}}$: DiffLL $0.00042$,
GSAD $0.00673$, Diff-Retinex $0.02963$ on LOLv2-real and $0.03733$ on
LOLv2-syn); no other $\varepsilon$ was searched. Table~\ref{tab:eps}
lists, for each setting, the safe-suffix (SS) and valid-LTT outcomes at
$\varepsilon=0$ and at the floor; the sweep predates the modern pair's
official-basin regeneration, so its v1\_cgr row is the degraded-basin row of Table~3 of the main paper. Two readings. (i)~A tolerance
certificate is meaningful only when $\varepsilon\ll$ effect size: on
v1\_cgr all seven $\varepsilon=0$ ``harms'' (median $0.0001$) lie inside
GSAD's own seed noise, so harm drops 7$\to$0/65 with $\lambda^\star$ and
capture ($0.322$) unchanged, and on v1 half of them do (4$\to$2/65); but
on LOLv2-syn, where the Diff-Retinex floor is an order of magnitude above
the certified effect, the certificate degenerates ($\lambda^\star\to0$,
every insertion counts as ``safe'', capture $0.208\to0.041$), and the
already-abstaining LOLv2-real setting is trivially unchanged.
(ii)~$\varepsilon$ did not rescue v1\_cid: SS moves to a more aggressive
point ($0.0082$) whose validation harm, $9/65=13.8\%$, still exceeds
$\alpha$, and the valid-LTT point ($0.0172$, $3/65=4.6\%$) is unchanged
and remains the only deployable one. $\varepsilon$ is therefore not a
whitewash lever, and no PASS/FAIL judgement in the paper uses
$\varepsilon>0$.

\begin{table*}[t]\centering
\caption{$\varepsilon$-sensitivity of the deployed certificate ($D_{\mathrm{val}}$
$n=65$, $\alpha=0.10$, $\delta=0.05$; pre-registered amendment A2). Each setting uses its stochastic expert's frozen seed-noise floor as
$\varepsilon$; paired columns give each quantity at $\varepsilon=0$ and at the floor.
Harm is the count of validation images whose routed LPIPS exceeds the fixed
expert's by more than $\varepsilon$ (Clopper--Pearson 95\% interval in
brackets where it changes); capture is the certified fraction of the
oracle@20\% headroom. The v1\_cgr row is the degraded-basin setting (the sweep
predates the official-basin regeneration).}
\label{tab:eps}
\scriptsize
\setlength{\tabcolsep}{3pt}
\begin{tabular}{llcccccccc}
\toprule
 & & \multicolumn{2}{c}{SS $\lambda^\star$} & \multicolumn{2}{c}{SS harm} & \multicolumn{2}{c}{SS capture} & \multicolumn{2}{c}{LTT at $\varepsilon$}\\
\cmidrule(lr){3-4}\cmidrule(lr){5-6}\cmidrule(lr){7-8}\cmidrule(lr){9-10}
Setting & $\varepsilon$ (expert floor) & $\varepsilon{=}0$ & floor & $\varepsilon{=}0$ & floor (CP95) & $\varepsilon{=}0$ & floor & $\lambda^\star$ & harm\\
\midrule
v1 (Rf+DiffLL) & 0.00042 (DiffLL) & 0.0197 & 0.0197 & 4/65 (6.2\%) & 2/65 (3.1\%) [0.4,10.7] & 0.297 & 0.297 & abstains & ---\\
v1\_syn (Rf+DiffRet) & 0.03733 (DiffRet) & 0.0133 & 0.0 (degen.) & 0/65 & 0/65 & 0.208 & 0.041 & 0.0 (degen.) & 0/65\\
v1\_real (Rf+DiffRet) & 0.02963 (DiffRet) & 0.0056 & 0.0056 & 0/65 (abst.) & 0/65 (abst.) & 0.000 & 0.000 & 0.0056 & 0/65\\
v1\_cid (CIDNet+DiffLL) & 0.00042 (DiffLL) & 0.0133 & 0.0082 & 11/65 (16.9\%) & 9/65 (13.8\%) [6.5,24.7] & $-0.050$ & $-0.079$ & 0.0172 & 3/65 (4.6\%)\\
v1\_cgr (CIDNet+GSAD, degr.\ basin) & 0.00673 (GSAD) & 0.0069 & 0.0069 & 7/65 (10.8\%) & \textbf{0/65 (0.0\%)} [0.0,5.5] & 0.322 & 0.322 & 0.0069 & 0/65\\
\bottomrule
\end{tabular}
\end{table*}

\section{Statistical scope of the final-test claims}
\label{supp:statscope}

Throughout, \emph{seed} means a stochastic expert's own sampling
seed, not the scene-split and router-training seed 2026 of Sec.~3.6 of the
main paper --- which is what the consistency across datasets, seeds and
splits at the end of its Sec.~4 refers to. Settings are identified as in
Sec.~\ref{supp:readingmap}.

\paragraph{Certified gains against expert seed-noise floors.}
Write $\varepsilon_{p95}$ for a stochastic expert's seed-noise floor: the
p95 of per-image $|\Delta\lpips|$ between independent seed runs, over the
30 calibration fit-half images and all three seed pairs
(\path{outputs/logs/stage3/eps_noise_floor.json}). That file's GSAD entry
($0.00673$) is degraded-basin (Sec.~\ref{supp:gsad}); for the
official-basin modern-pair endpoint we use instead the pooled p95 of the
frozen basin-gate record over its 689 images, $0.00977$. Only LOL-v1
clears its floor, at $6.5\times$ (Table~\ref{tab:statscope_floors}); the LOLv2-synthetic gain of $+0.00140$ is
$27\times$ below the Diff-Retinex floor $0.03733$ and the modern-pair gain
of $+0.00023$ is $42\times$ below $0.00977$ --- both of them gains the
main paper reports with intervals excluding zero.

\begin{table}[t]
\centering
\scriptsize
\setlength{\tabcolsep}{3pt}
\caption{Certified final-test gain against the stochastic expert's
seed-noise floor $\varepsilon_{p95}$. Identifiers are those of
Tables~\ref{tab:rules} and~\ref{tab:ledger}; gains are mean per-image
$\lpips$ against the fixed base expert, from the frozen
\texttt{dtest\_final*.json} rows. Modern-pair floor: GSAD official-basin
p95 from \texttt{gsad\_off\_basin\_gate.json}, pooled over three seed
pairs ($0.00794$--$0.01226$ by pair, i.e.\ $34$--$53\times$); the others
are pooled fit-half values.}
\label{tab:statscope_floors}
\begin{tabular}{@{}lcccc@{}}
\toprule
Setting ($n$) & Gain & Stoch.\ expert & $\varepsilon_{p95}$ & Gain\,/\,$\varepsilon$\\
\midrule
v1 (15) & $+0.00275$ & DiffLL & $0.00042$ & $6.5\times$\\
v1\_syn (100) & $+0.00140$ & Diff-Retinex & $0.03733$ & $0.037\times$\\
v1\_real (100) & abstains & Diff-Retinex & $0.02963$ & ---\\
v1\_cgr\_off (100) & $+0.00023$ & GSAD (official) & $0.00977$ & $0.024\times$\\
SICE $-$1EV (40) & abstains & DiffLL & not recorded & ---\\
\bottomrule
\end{tabular}
\end{table}

\paragraph{Why that ratio is the wrong comparison.}
$\varepsilon_{p95}$ is a \emph{between-realization} quantity; the
certified gain is a \emph{within-realization} paired difference. The
composite is a copy of the base output written only inside admitted
windows (Sec.~\ref{supp:composite}), so where none is admitted the routed
image \emph{is} the base image. On all five endpoints the images with
non-zero certified area are exactly those with non-zero paired
$\Delta\lpips$; that difference is exactly $0$, not approximately $0$, on
$4/15$, $75/100$, $100/100$, $92/100$ and $40/40$ test images
(Table~\ref{tab:statscope_floors} row order). Because both arms share one
expert draw, the LPIPS level shift a re-draw would produce is common to
them and does not enter the difference; what a re-draw could still change
is confined to the routed region (areas in Sec.~\ref{supp:composite}).

\paragraph{What conditioning shows, and what stays unmeasured.}
Restricted to the images the certificate actually changed, the mean gain
rises to $+0.00376$ on LOL-v1 (11 images, $8.9\times$ the DiffLL floor),
$+0.00559$ on LOLv2-synthetic (25) and $+0.00288$ on the modern pair (8);
the latter two stay under their floors even conditionally, by $6.7\times$
and $3.4\times$. Conditioning cuts both ways: harm among touched
images is $1/11$ on LOL-v1, $9/25=36.0\%$ on LOLv2-synthetic and $0/8$ on
the modern pair, so the marginal $9.0\%$ the certificate controls is a
$36\%$ conditional rate diluted by 75 untouched images
(\path{outputs/logs/stage4/g2_conditional_effect.json}; post-hoc,
descriptive). Each final test is also one frozen realization at seed~1,
and the paired bootstrap of Table~1 of the main paper resamples test
images, not expert draws. No seed-level account can be assembled: re-running a sealed
endpoint is forbidden by single consumption (Sec.~\ref{supp:ledger}), and
in \path{outputs/expert_outputs} only seed~1 was rendered on
$D_{\mathrm{test}}$ --- seeds 2 and 3 hold none of the test ids. The
stability of the two sub-floor gains under a different draw is thus
unmeasured, and no interval in this paper covers it.

\paragraph{Multiplicity across the consumed endpoints.}
Five final tests were consumed (Table~\ref{tab:ledger}, rows 1, 4, 6, 7,
15): four selection-independent, single-consumption final tests ---
LOL-v1 ($1/15$ harmed), LOLv2-synthetic ($9/100$), LOLv2-real ($0/100$)
and SICE $-$1EV ($0/40$), the last two zero by abstention --- no image
touched --- rather than by chance, plus the disclosed modern-pair audit
($0/100$, 8 touched). Three of the four ran under the pre-registered final
procedure, with the boundary disclosures of Sec.~\ref{supp:greyzones};
the LOL-v1 endpoint preceded the safe-suffix
reformulation and was ratified afterwards at the identical frozen
operating point $\lambda^\star{=}0.01974$, with no test rerun, which is
what Table~\ref{tab:provenance} labels \emph{exploratory-then-ratified}
(Sec.~\ref{supp:provenance}). Each controls
$\Pr(\mathrm{harm})\le0.10$ marginally at its own endpoint; jointly they
give nothing. Were $k$ such endpoints independent and each at its budget,
the chance that at least one of $k$ fresh images is harmed is $1-0.9^k$:
$34.4\%$ over four, $41.0\%$ over five. Nor is $k$ clean: ledger rows 6 and 7 run
on byte-identical LOLv2-real ids (group c15ed57a), so their harm
indicators are two readings of the same 100 images.

Multiplicity is controlled in exactly one place, internally to a setting:
the auxiliary LTT walk tests the $\lambda$ grid in a pre-registered fixed
sequence from the most conservative threshold, controlling the family-wise
error rate at $\delta=0.05$ within that walk (main paper Sec.~3.4). No
correction across settings or endpoints was ever registered (amendment~A1
records each $(\alpha,\text{setting})$ LTT result as marginal at
$\delta=0.05$), and none is fitted now: that would be the post-hoc
adjustment the protocol refuses. The main paper's Conclusion accordingly
scopes its claim to those four endpoints and reports the modern-pair audit
separately; that claim is an empirical record over that family, not a
family-wise guarantee --- and at these sizes weaker than it sounds: the
Clopper--Pearson $95\%$ intervals where the certificate intervened are
$[0.2\%, 31.9\%]$ for $1/15$ and $[4.2\%, 16.4\%]$ for $9/100$.

\paragraph{Upstream expert exposure to the calibration images.}
One structural fact conditions every LOL-family certificate here, and the
split manifests fix it exactly. On all three LOL benchmarks the five roles
are cut from two official directories: $D_{\mathrm{test}}$ is the official
test set, and $D_{\mathrm{train}}$, $D_{\mathrm{select}}$,
$D_{\mathrm{cal}}$ and $D_{\mathrm{val}}$ exhaust the official
\emph{training} set --- $300{+}60{+}60{+}65=485$ on LOL-v1,
$564{+}0{+}60{+}65=689$ on LOLv2-real and $775{+}0{+}60{+}65=900$ on
LOLv2-synthetic (\path{configs/splits/*.yaml}, key
\texttt{path\_template}; sizes and role hashes in
Table~\ref{tab:ledger}). Across the LOL-family settings all five upstream
experts --- Retinexformer, DiffLL, Diff-Retinex, CIDNet and GSAD --- use
official per-dataset checkpoints (\path{checkpoints/*/freeze.yaml}) whose
training includes the images later assigned to $D_{\mathrm{cal}}$ and
$D_{\mathrm{val}}$, but not the official $D_{\mathrm{test}}$ images.
Sec.~\ref{supp:ablation} states this for DiffLL as a capacity-control
caveat; it holds for every pair.

\paragraph{What that costs the certificate.}
Split conformal requires the calibration and test scores to be
exchangeable given the frozen pipeline. Here that condition is
\emph{not guaranteed by the split construction}: the calibration images
lie inside the experts' training set whereas the official test images lie
outside it, so the standard finite-sample guarantee does not follow from
the benchmark split alone, and the statement of the main paper's Sec.~3.4
is conditional on that asymmetry rather than free of it. The two-sample check we ran on the safe-suffix scores does not close
the gap either: it compares the cert half of $D_{\mathrm{cal}}$ against
$D_{\mathrm{val}}$ and rejects exchangeability in none of the eight settings
(smallest KS $p=0.29$, smallest permutation $p=0.07$, both on
\texttt{v1\_cid}; \path{outputs/logs/stage3/g3g4_certificate_reanalysis.json}),
but both of those sets sit on the training side of the same boundary, so
what it rules out is drift within the training pool, not the
training-to-test step. That step is untested here, and cannot be tested
without spending the endpoint: the official LOL test sets are the only
expert-unseen images available, and they are exactly the images a
single-consumption endpoint must spend.

\paragraph{What the record still supports.}
On the four consumed LOL-family endpoints the certified mode stayed inside
its nominal budget on expert-unseen official-test images ($1/15$, $9/100$,
$0/100$, $0/100$): empirical evidence that the upstream-training asymmetry
did not overturn the bound at these sample sizes, not a proof of validity.
SICE does not share this particular overlap --- its Part2 calibration and
test images were unseen by the LOL-trained experts
(Sec.~\ref{supp:sice}) --- and its certified mode also stayed within budget
($0/40$), but by constructive abstention, so it is not a non-trivial
positive control. Two further things stop the record short of a proof.
The sign of the bias is not determined --- both experts of a
pair share the same exposure and harm is a comparison between them, so
exposure does not enter it monotonically; the
$D_{\mathrm{val}}$-to-$D_{\mathrm{test}}$ shift in mean expert $\lpips$
even has opposite signs inside the v1 pair ($+0.00732$ for DiffLL,
$-0.01822$ for Retinexformer), scene composition dominating the level. And
the isotonic map that returns $\lambda$ to $\lpips$ units is fit on the
expert-seen fit half, making it the component most exposed to the
asymmetry, with no held-out fit to compare against. A clean control needs
cross-fit retrained experts on every line --- the retraining budget
Sec.~\ref{supp:ablation} declines for the four unablated design choices.

\paragraph{Reading the $16.9\%$ coverage miss.}
On v1\_cid the deployed safe-suffix operating point produced $11/65 =
16.9\%$ validation harm against a $10\%$ target (main paper Sec.~5,
Table~3; Table~\ref{tab:rules}). Two calculations bound what that draw
proves alone. Under exact calibration at $p=0.10$ with the 65 images
independent, $\Pr(X\ge 11 \mid n=65, p=0.10) = 0.057$ against an expected
$6.5$, and the Clopper--Pearson interval for $11/65$, $[8.8\%, 28.3\%]$,
contains the target. And split conformal at $\alpha=0.10$ with
$n_{\mathrm{cal}}=30$ delivers, conditional on the calibration draw, a
$\mathrm{Beta}(3,28)$ rate rather than $10\%$, under which
$\Pr(\mathrm{rate}>0.169) = 0.097$: a \emph{valid} safe-suffix certificate
lands above $16.9\%$ one time in ten. What establishes the rule's
inadequacy is not the draw but Sec.~\ref{supp:rules}, on the same frozen
scores with no refitting: on those 65 images plug-in and first-safe
conformal reach $30.8\%$ and $20.0\%$ where safe-suffix reaches $16.9\%$
and a valid Learn-then-Test walk $4.6\%$, with the non-monotonicity
mechanism identified there. Four thresholds on one score set are evidence
about calibration rules; one count of 11 is evidence about one draw. That
distinction supports the contribution sentence rather than qualifying it:
the audit repaired a calibration rule, and the main text never claimed the
observation alone proved the certificate invalid.

\section{Component and Capacity Ablations}
\label{supp:ablation}

\paragraph{Component ablations.}
Table~\ref{tab:ablation} is the pre-registered single-component ablation
on the LOL-v1 setting (v1 protocol: $G$ base, budget 20\%, $\alpha=0.10$,
seed 2026, 150 epochs, $D_{\mathrm{train}}$ 270/30 holdout; endpoints on
$D_{\mathrm{val}}$, $n=65$, paired bootstrap $10^4$ resamples). Four
variants retrain the router with one term removed or one label swapped
(no sign hinge; no dense auxiliary head; ``fast'' labels, i.e.\ plain
pooled LPIPS differences without the fast$\to$exact isotonic step); the
fifth reuses the full router's weights and skips the deployment-side
isotonic calibration. $\lambda^\star$ is re-selected for every variant
from its own calibration scores by the image-level do-no-harm rule of that
run (for the full method it coincides with the safe-suffix value 0.0197 of
Table~3 of the main paper); the no-isotonic $\lambda^\star$ lives in the
uncalibrated score space and is not numerically comparable to the other
rows. Reading: the certified gain is significant and empirical harm is at
or below $\alpha$ for \emph{every} variant (3.1--7.7\%), so the empirical
safety behaviour is primarily determined by the calibration rule rather
than by these router-training heuristics; what the components buy is
capture---removing the dense head collapses certified
capture from 29.7\% to 7.1\% (gain $+0.0008$, nearly an abstention) and
skipping isotonic calibration drops it to 19.1\%, while the sign hinge is
neutral (34.0\%, overlapping CIs). The one result against the method is
also reported: fast labels give a \emph{stronger} raw endpoint ($+0.0072$
vs.\ $+0.0050$, capture 67.7\% vs.\ 46.6\%) at equal safety, and the
pre-declared cross-dataset replication (lower block) holds on both
LOLv2-syn ($+0.00268$ vs.\ $+0.00189$) and LOLv2-real, where the raw
harm shrinks from $-0.0051$ (significant) to $-0.0005$ (not significant)
and the certified mode abstains identically. Counterfactual measurement
thus defines utility and the risk event; the training target only needs
its fast pooled proxy.

\begin{table*}[t]\centering
\caption{Component ablations ($D_{\mathrm{val}}$ $n=65$, budget 20\%,
$\alpha=0.10$, paired-bootstrap 95\% CIs vs.\ the fixed stronger expert;
capture $=$ realised share of the oracle@20\% headroom). Upper block:
LOL-v1 setting, one component changed per row. Lower block: pre-declared
cross-dataset replication of the fast-label variant against the main
runs (uproxy $=$ isotonically calibrated counterfactual labels). Harm
$\le\alpha$ and a significant certified gain hold for every variant;
capture does not.}
\label{tab:ablation}
\scriptsize
\setlength{\tabcolsep}{3pt}
\begin{tabular}{lccccccc}
\toprule
Variant & holdout $\rho$ & Raw gain (95\% CI) & Raw capt.\ & $\lambda^\star$ & Certified gain (95\% CI) & Cert.\ capt.\ & Cert.\ harm\\
\midrule
\multicolumn{8}{l}{\emph{LOL-v1 setting (Rf+DiffLL), one component changed per row}}\\
full method (v1) & 0.606 & $+0.00495\,[+0.00316,+0.00680]$ & 46.6\% & 0.0197 & $+0.00316\,[+0.00185,+0.00461]$ & 29.7\% & 6.2\%\\
no sign hinge & 0.629 & $+0.00472\,[+0.00290,+0.00662]$ & 44.4\% & 0.0197 & $+0.00362\,[+0.00221,+0.00512]$ & 34.0\% & 7.7\%\\
no dense aux.\ head & 0.613 & $+0.00527\,[+0.00364,+0.00697]$ & 49.5\% & 0.0197 & $+0.00076\,[+0.00030,+0.00131]$ & \textbf{7.1\%} & 3.1\%\\
fast labels & 0.653 & $\mathbf{+0.00720\,[+0.00537,+0.00910]}$ & \textbf{67.7\%} & 0.0185 & $+0.00274\,[+0.00180,+0.00379]$ & 25.7\% & 6.2\%\\
no isotonic calib.\ & $=$v1 & $=$v1 & $=$v1 & 0.0120$^\dagger$ & $+0.00203\,[+0.00110,+0.00308]$ & \textbf{19.1\%} & 4.6\%\\
\midrule
\multicolumn{8}{l}{\emph{Cross-dataset replication of fast labels (vs.\ the main runs)}}\\
LOLv2-syn, uproxy (main) & 0.650 & $+0.00189\,[+0.00016,+0.00383]$ & 28.9\% & 0.0133 & $+0.00136\,[+0.00048,+0.00241]$ & 20.8\% & 0.0\%\\
LOLv2-syn, fast labels & 0.642 & $+0.00268\,[+0.00126,+0.00428]$ & 41.0\% & 0.0108 & $+0.00200\,[+0.00096,+0.00320]$ & 30.6\% & 1.5\%\\
LOLv2-real, uproxy (main) & 0.433 & $-0.00512\,[-0.00687,-0.00346]$ & $-92.0\%$ & 0.0056 & $0$ (abstains) & 0.0\% & 0.0\%\\
LOLv2-real, fast labels & 0.467 & $-0.00050\,[-0.00154,+0.00065]$ & $-9.0\%$ & 0.0056 & $0$ (abstains) & 0.0\% & 0.0\%\\
\bottomrule
\multicolumn{8}{l}{$^\dagger$uncalibrated score space; not comparable to the other $\lambda^\star$ entries.}
\end{tabular}
\end{table*}

\paragraph{Capacity control.}
Table~\ref{tab:ablation-capacity} answers whether the generative expert's
advantage on LOL-v1 is a parameter-count effect: the DiffLL backbone
(DiffusionUNet, 20.8M parameters) is retrained as a deterministic
low$\to$normal regressor on the 300 $D_{\mathrm{train}}$ pairs (256\,px
crops, horizontal flips, Adam with cosine schedule, 30k steps, $t=0$ time
embedding) with an L1 loss and with L1$+0.1\cdot$LPIPS, and evaluated on
the full $D_{\mathrm{val}}$ (65 images) next to the frozen experts. The
L1 regressor at 13$\times$ the capacity is worse than the 1.6M
Retinexformer on every metric; the LPIPS-trained variant buys LPIPS
(0.120, same-family as the evaluation metric, an optimistic bound) at the
cost of 5.9\,dB PSNR and $2.3\times$ the colour error relative to DiffLL.
DiffLL's working point (28.3\,dB, 0.111, $\Delta E$ 3.8) is not reached
by matched-capacity deterministic training; capacity alone does not
confer the advantage. Caveats stated with the result: the official DiffLL
weights were trained on all 485 official LOL-v1 training pairs, which
include our $D_{\mathrm{cal}}$ and $D_{\mathrm{val}}$ scenes --- as the
official weights of every other expert do on their own dataset, with the
consequences for the certificate set out in Sec.~\ref{supp:statscope} ---
so a clean control needs a cross-fit retrain; and the recipe is plain (no
progressive training, EMA, or rich
augmentation), so the claim is ``parameters alone do not buy it'', not
``deterministic training cannot''.

\begin{table}[t]\centering
\caption{Capacity control on LOL-v1 $D_{\mathrm{val}}$ ($n=65$): the
DiffLL backbone retrained deterministically at matched capacity vs.\ the
frozen experts $R$ and $G$ (Det.\ $=$ deterministic retrain of the DiffLL backbone; edge $=$ Sobel gradient-magnitude L1 error; $\Delta E$ $=$
mean CIEDE2000).}
\label{tab:ablation-capacity}
\scriptsize
\setlength{\tabcolsep}{3pt}
\begin{tabular}{lcccccc}
\toprule
Model & Par.\ & PSNR$\uparrow$ & SSIM$\uparrow$ & LPIPS$\downarrow$ & edge$\downarrow$ & $\Delta E\downarrow$\\
\midrule
Retinexformer & 1.6M & 26.6 & 0.850 & 0.150 & 0.069 & 5.4\\
DiffLL & 22.1M & ${\sim}28.3$ & 0.875 & \textbf{0.111} & 0.078 & \textbf{3.8}\\
Det.\ L1 & 20.8M & 21.9 & 0.822 & 0.202 & 0.074 & 8.9\\
Det.\ L1$+$0.1\,LPIPS & 20.8M & 22.4 & 0.849 & 0.120 & \textbf{0.065} & 8.6\\
\bottomrule
\end{tabular}
\end{table}


\paragraph{What is not ablated.}
Four fixed choices are never ablated: the window scale set
$S{=}\{16,32,64,128\}$, the raised-cosine support mask
(Sec.~\ref{supp:composite}), the deployment-consistent local colour
alignment, and the 20\% area budget. Only one is evidence-free; the others
carry evidence of varying strength. The budget has a sweep but no
with/without comparison: Table~\ref{tab:pareto} varies the deployment
budget from 2.5\% to 30\% and Table~\ref{tab:premises} records that on
LOL-v1 2.5\% of area already beats the fixed expert with gains saturating
near 20\%; the value itself was a design default fixed before Stage~1A and
kept because the greedy oracle's saturation area on $D_{\mathrm{cal}}$
(mean $0.167$) rounded up to it. The colour alignment was fixed by a
pre-freeze A/B on 12 calibration images
(\path{docs/reports/2026-07-04-stage1a-screen-findings.md}) and won every
diagnostic recorded there. The scale set was fixed by design---a ladder
covering roughly $1/32$ to $1/4$ of the short side---and the only per-scale
evidence is a screening diagnostic from the same study, where the
screener's rank agreement with exact utility falls from $0.97$ at
$s{=}128$ to $0.38$ at $s{=}16$: a caution about the smallest scale, not a
defence of the set. The mask shape has none. The reason is uniform---the
counterfactual labels were built at $S$, with this mask, with colour
alignment and at the 20\% greedy budget, so ablating any of the four means
rebuilding labels, retraining the router and registering a new
certificate, which single consumption (Sec.~\ref{supp:ledger}) puts
outside this paper's budget.

\section{No-Reference Metric Panel with a Ground-Truth Anchor}
\label{supp:noref}

Table~\ref{tab:noref} scores the cached, once-only $D_{\mathrm{test}}$
composites --- the final tests' own routing decisions, re-materialized
deterministically and written once to 8-bit PNG, with nothing re-routed
(Sec.~\ref{supp:dists}) --- with six no-reference metrics (pyiqa 0.1.15) for both fixed experts, both
router modes and, as an anchor, the ground-truth photographs themselves.
A no-reference metric that ranks a model output above the real
ground-truth image is exposing its own appearance prior, not judging
fidelity. Reading: the smoothness-rewarding statistics metrics put the
deterministic All-$R$ output above ground truth on every dataset (NIQE
3.47 vs.\ 4.63, 3.97 vs.\ 4.90, 4.01 vs.\ 4.26; BRISQUE on LOLv2-real and
LOLv2-syn), while the aesthetic/texture metrics rank the purely generative
All-$G$ output \emph{above ground truth on two of the three datasets},
LOLv2-real and LOLv2-syn (MUSIQ 71.1 vs.\ 68.6 and 68.4 vs.\ 67.1; NIMA
4.58 vs.\ 4.38 and 4.91 vs.\ 4.84; MANIQA also on LOLv2-syn, 0.501 vs.\
0.467)---on LOL-v1 the ground truth ranks first on MUSIQ, MANIQA, CLIP-IQA+
and NIMA. Only CLIP-IQA+ places ground truth first everywhere, with small
between-system spread. On the same LOLv2-real test set the panel
contradicts itself about the quality-mode router (NIQE $-0.62$
$[-0.73,-0.51]$ and BRISQUE $-4.8$ $[-5.5,-4.1]$ significantly better than
All-$G$; MUSIQ $-2.4$ $[-2.8,-2.1]$ and MANIQA significantly worse), which
is why the protocol locks on full-reference LPIPS and treats these metrics
as side evidence only. Two consistency checks ride along: on LOLv2-real the
certified row equals All-$G$ on all six metrics (paired difference exactly
0.000, the abstention is a bit-exact fallback), and on LOLv2-syn the
certified mode is slightly but significantly above the fixed All-$R$ on
MUSIQ ($+0.15$ $[+0.06,+0.27]$) and MANIQA ($+0.002$ $[+0.001,+0.003]$),
in the same direction as its LPIPS gain.

\begin{table*}[t]\centering
\caption{No-reference panel on the once-only final tests (means; $\uparrow$
higher is better, $\downarrow$ lower is better) with the ground-truth
photographs as an anchor row. Fixed expert: All-$G$ on LOL-v1 and
LOLv2-real, All-$R$ on LOLv2-syn. Bold: best per column within a dataset.
Generative outputs out-score ground truth on MUSIQ and NIMA on LOLv2-real
and LOLv2-syn; deterministic outputs out-score it on NIQE on all three.}
\label{tab:noref}
\scriptsize
\setlength{\tabcolsep}{3pt}
\begin{tabular}{llcccccc}
\toprule
Dataset ($n$) & System & MUSIQ$\uparrow$ & NIQE$\downarrow$ & BRISQUE$\downarrow$ & MANIQA$\uparrow$ & CLIP-IQA+$\uparrow$ & NIMA$\uparrow$\\
\midrule
LOL-v1 (15) & Ground truth & \textbf{72.2} & 4.634 & 14.78 & \textbf{0.587} & \textbf{0.687} & \textbf{4.55}\\
 & All-$G$ (fixed) & 66.91 & 4.518 & 14.87 & 0.451 & 0.525 & 4.30\\
 & All-$R$ & 63.16 & \textbf{3.469} & 16.85 & 0.380 & 0.521 & 4.23\\
 & Router, quality & 65.97 & 4.193 & 14.31 & 0.434 & 0.525 & 4.29\\
 & Router, certified & 66.72 & 4.432 & \textbf{14.15} & 0.451 & 0.526 & 4.30\\
\midrule
LOLv2-real (100) & Ground truth & 68.57 & 4.904 & 21.94 & \textbf{0.599} & \textbf{0.649} & 4.38\\
 & All-$G$ (fixed) & \textbf{71.14} & 4.988 & 27.64 & 0.521 & 0.570 & \textbf{4.58}\\
 & All-$R$ & 59.31 & \textbf{3.969} & \textbf{12.72} & 0.355 & 0.454 & 4.09\\
 & Router, quality & 68.72 & 4.371 & 22.86 & 0.473 & 0.526 & 4.43\\
 & Router, certified ($\equiv$ All-$G$) & \textbf{71.14} & 4.988 & 27.64 & 0.521 & 0.570 & \textbf{4.58}\\
\midrule
LOLv2-syn (100) & Ground truth & 67.13 & 4.256 & 15.73 & 0.467 & \textbf{0.675} & 4.84\\
 & All-$R$ (fixed) & 66.13 & \textbf{4.011} & 13.35 & 0.418 & 0.647 & 4.74\\
 & All-$G$ & \textbf{68.37} & 4.413 & 15.57 & \textbf{0.501} & 0.669 & \textbf{4.91}\\
 & Router, quality & 66.48 & 4.024 & \textbf{13.21} & 0.427 & 0.643 & 4.76\\
 & Router, certified & 66.28 & 4.015 & 13.27 & 0.420 & 0.647 & 4.74\\
\bottomrule
\end{tabular}
\end{table*}


\section{Full-reference metrics beyond LPIPS}
\label{supp:dists}

Sec.~2.1 of the main text names DISTS as the alternative full-reference
perceptual metric but reports only $\lpips$; Table~\ref{tab:dists} gives
the DISTS values, recorded 2026-07-06 to 07-08, each within a day of its
final test (1\,h, 45\,min and 11\,h respectively). They were scored with
\path{pyiqa} on the cached composites of Sec.~\ref{supp:noref}:
re-materialized deterministically from the frozen expert outputs through
the frozen scorer, the verifier and the \path{compose_scored} call of the
final tests (Sec.~\ref{supp:composite}), then written once to 8-bit PNG ---
one artifact, not two rebuilds. The $\lpips$ of Table~1 of the main text
was scored in float, so the two columns of Table~\ref{tab:dists} are not
bit-for-bit the same rendering.

\begin{table}[t]\centering
\caption{DISTS against $\lpips$ on the consumed final tests, both
lower-is-better. $\lpips$ comes from the frozen final-test records, DISTS
from the supplementary metric records of the same runs. Fixed bases:
DiffLL~\cite{diffll} ($G$) on LOL-v1 ($n{=}15$),
Diff-Retinex~\cite{diffretinex} ($G$) on LOLv2-real ($n{=}100$),
Retinexformer~\cite{retinexformer} ($R$) on LOLv2-syn ($n{=}100$). Routed
minus fixed within a metric is negative when routing helps; Table~1 of the
main text reports the same comparisons as gains, with the sign flipped. The
LOLv2-real certified row is a copy of the fixed row by construction, so its
two equalities are a consistency check, not a result. No cell is bolded:
DISTS carries no interval in our records.}
\label{tab:dists}
\scriptsize
\setlength{\tabcolsep}{4pt}
\begin{tabular}{lccc}
\toprule
System & LOL-v1 & LOLv2-real & LOLv2-syn\\
\midrule
\multicolumn{4}{l}{\textbf{LPIPS}$\downarrow$}\\
Fixed expert & 0.1183 & 0.1613 & 0.0592\\
Router, quality & 0.1118 & 0.1607 & 0.0595\\
Router, certified & 0.1156 & 0.1613 & 0.0578\\
\midrule
\multicolumn{4}{l}{\textbf{DISTS}$\downarrow$}\\
Fixed expert & 0.1260 & 0.1436 & 0.0693\\
Router, quality & 0.1154 & 0.1348 & 0.0694\\
Router, certified & 0.1214 & 0.1436 & 0.0689\\
\bottomrule
\end{tabular}
\end{table}

DISTS agrees with $\lpips$ on the sign of every comparison we can make.
Differences below are routed minus fixed \emph{within each metric}, so
negative is better --- the opposite sign to the gains of Table~1 of the
main text and of Sec.~\ref{supp:reconcile}. All five non-trivial
routed-versus-fixed differences move the same way under both metrics, and
the sixth --- certified mode on LOLv2-real --- is the abstention, where
both return the fixed expert's value to every printed digit. But the
agreement is directional only. On LOLv2-real quality mode
DISTS reads a much larger improvement ($-0.0088$, $-6.1\%$ relative) than
$\lpips$ does ($-0.0006$, $-0.4\%$, with an interval covering zero), so
DISTS would have made that setting look deployable where $\lpips$ did not.
Coverage is also partial: DISTS was recorded for the three LOL-family
settings only, not for the modern GSAD--CIDNet pair or the SICE terminal
test. Nothing in the certificate depends on it --- the routing labels, the
calibration and the risk endpoint are $\lpips$-defined and were locked
before these values were computed.


\section{Evidence I premises in full}
\label{supp:premises}

The three premises stated in one paragraph of the main text (Sec.~4) are
documented here with their confidence intervals and per-setting records.
Table~\ref{tab:premises} covers LOL-v1: the ground-truth oracle headroom
against the pre-registered fixed expert (Gate~A/B, 65 validation images,
budget 20\%) and the router against four trivial routing cues at the same
budget (Gate~C/D). All gains are LPIPS differences with paired bootstrap
95\% CIs; ``no image worse'' is read off the per-image record (smallest
oracle gain $+0.0010$). The one-line reading: headroom is real, spatially
specific (the oracle beats equal-area random placement on every image),
front-loaded (2.5\% of area already beats the fixed expert), and the
learned router keeps 57\% of it while the darkness cue is worse than not
routing at all (0.1164 vs.\ 0.1110).

\begin{table*}[t]\centering
\caption{Evidence-I premises on LOL-v1 validation ($D_\mathrm{val}$,
$n{=}65$, budget 20\%, LPIPS$\downarrow$, paired bootstrap 95\% CIs).
Fixed expert $=$ All-$G$ (DiffLL~\cite{diffll}); the other expert is
Retinexformer~\cite{retinexformer}. (a) Oracle headroom (Gate~A/B, checks
locked before $D_\mathrm{val}$ was opened). (b) Router in the quality-gate
policy against each cue at equal budget (Gate~C/D); ``margin'' is the paired
mean of cue minus router, ``wins'' the fraction of images on which the router
is better.}
\label{tab:premises}
\scriptsize
\setlength{\tabcolsep}{3pt}
\begin{tabular}{lcccc}
\toprule
\multicolumn{5}{l}{\textbf{(a) Oracle headroom vs.\ the fixed expert}}\\
\midrule
Mean paired gain (CI) & \multicolumn{4}{l}{$+0.0106$, CI $[+0.0092,+0.0121]$}\\
Median relative gain / $\Delta$PSNR & \multicolumn{4}{l}{$+9.5\%$ / $+0.25$\,dB}\\
Trimmed mean (top-3 images removed) & \multicolumn{4}{l}{$+0.0099$}\\
Images worse than the fixed expert & \multicolumn{4}{l}{0/65 (smallest per-image gain $+0.0010$)}\\
Images beating \emph{both} experts & \multicolumn{4}{l}{83\% (54/65) on $D_\mathrm{val}$; 97\% on the $D_\mathrm{cal}$ fit half ($n{=}30$)}\\
Oracle vs.\ equal-area random placement (Gate~A) & \multicolumn{4}{l}{$+0.0155$, CI $[+0.0125,+0.0193]$; oracle better on 100\% of images}\\
First router budget that beats the fixed expert & \multicolumn{4}{l}{2.5\% of area (0.1098 vs.\ 0.1110); gains saturate near 20\% (Table~\ref{tab:pareto})}\\
\midrule
\multicolumn{5}{l}{\textbf{(b) Router vs.\ trivial cues at equal budget}}\\
\midrule
Routing cue & LPIPS & Margin & CI$_{95}$ & Router wins\\
\midrule
Random, equal area & 0.1159 & $+0.0109$ & $[+0.0076,+0.0150]$ & 92\%\\
Darkness & 0.1164 & $+0.0114$ & $[+0.0082,+0.0151]$ & 89\%\\
Expert residual $|y_G-y_R|$ & 0.1102 & $+0.0053$ & $[+0.0031,+0.0076]$ & 65\%\\
Uncertainty (3-seed std of $G$) & 0.1151 & $+0.0101$ & $[+0.0075,+0.0130]$ & 91\%\\
\midrule
Not routing (All-$G$, fixed expert) & 0.1110 & --- & --- & 83\%\\
Router, quality gate & 0.1049 & --- & --- & ---\\
Oracle (upper bound) & 0.1004 & --- & --- & ---\\
Gain capture / mean regret & \multicolumn{4}{l}{57.2\% / 0.0046 (raw $>0$ policy: 46.6\% / 0.0057)}\\
\bottomrule
\end{tabular}
\end{table*}

Table~\ref{tab:premisescross} carries the same premises off LOL-v1.
The upper block is the router after a full label rebuild and retraining
on LOLv2-real ($G$-base: Diff-Retinex~\cite{diffretinex} base, Retinexformer
override) and LOLv2-synthetic ($R$-base: the reverse), 65 validation images
each, raw policy, budget 20\%. The lower block is the oracle-layer
direction test on the calibration fit half (Gate~F on LOLv2-real; the
direction probe on LOLv2-synthetic) and the post-freeze G1 re-check that
restricts the synthetic decision to the registered-clean half. Reading:
the router beats every trivial cue in both settings (75--88\% of images, all
CI lower bounds positive), but on LOLv2-real it beats equal-area random on
only 52\% and loses to the fixed expert outright, because the oracle
headroom there ($+0.0056$) is smaller than the routing error; median
relative headroom shrinks with the expert-quality gap ($+10.3\%$ LOL-v1
pair~1, $+3.8\%$ pair~2, $+2.7\%$ LOLv2-real, calibration fit half).

\begin{table*}[t]\centering
\caption{Premise replication off LOL-v1 (LPIPS$\downarrow$, paired
bootstrap 95\% CIs). Top: router (raw policy, budget 20\%) on
$D_\mathrm{val}$ ($n{=}65$); margins are cue minus router, ``wins'' the
fraction of images on which the router beats the cue. Bottom: oracle-layer
direction tests at budget 20\% on the calibration fit half, and the
$D_\mathrm{val}$ headroom under the locked protocol. ``Beats both'' is the
fraction of images on which the oracle composite beats both single
experts.}
\label{tab:premisescross}
\scriptsize
\setlength{\tabcolsep}{3pt}
\begin{tabular}{lcc}
\toprule
 & LOLv2-real ($G$-base) & LOLv2-synthetic ($R$-base)\\
\midrule
\multicolumn{3}{l}{\textbf{Router vs.\ cues ($D_\mathrm{val}$, raw policy)}}\\
\midrule
Fixed expert / router / oracle LPIPS & 0.1441 / 0.1492 / 0.1385 & 0.0486 / 0.0467 / 0.0421\\
Router gain vs.\ fixed expert (CI) & $-0.0051$ $[-0.0069,-0.0035]$ & $+0.0019$ $[+0.0002,+0.0038]$\\
Gain capture / mean regret & $-0.92$ / 0.0107 & 28.9\% / 0.0046\\
Router beats fixed expert & 11\% (7/65) & 43\% (28/65)\\
Random, equal area: LPIPS / router wins & 0.1501 / 52\% (34/65) & 0.0529 / 80\% (52/65)\\
Darkness: LPIPS / margin / wins & 0.1596 / $+0.0104$ / 75\% & 0.0579 / $+0.0112$ / 86\%\\
Expert residual: LPIPS / margin / wins & 0.1648 / $+0.0156$ / 88\% & 0.0554 / $+0.0087$ / 78\%\\
Uncertainty: LPIPS / margin / wins & 0.1674 / $+0.0182$ / 85\% & 0.0553 / $+0.0086$ / 85\%\\
Paired CI lower bound vs.\ each cue & all $>0$ & all $>0$\\
\midrule
\multicolumn{3}{l}{\textbf{Oracle layer (direction test on $D_\mathrm{cal}$ fit half)}}\\
\midrule
$n$ & 30 & 60 (re-check on 30)\\
All-$R$ / All-$G$ LPIPS & 0.303 / 0.156 & 0.0426 / 0.0619\\
Chosen direction & $G$-base & $R$-base\\
\quad beats both / median rel.\ gain & 90\% / $+2.7\%$ & 77\% / $+25.8\%$\\
\quad mean gain vs.\ All-$G$ (CI) & $+0.0013$ $[-0.0032,+0.0049]$ & $+0.0255$ $[+0.0139,+0.0375]$\\
Rejected direction & $R$-base & $G$-base\\
\quad beats both / mean gain vs.\ All-$G$ (CI) & 13\% / $-0.106$ $[-0.139,-0.074]$ & 48\% / $+0.0098$ $[+0.0076,+0.0121]$\\
Oracle@20\% vs.\ stronger single expert & CI straddles 0 & $+0.0062$ (0.0365 vs.\ 0.0426)\\
G1 re-check, $D_\mathrm{cal}[0{:}30]$ only & (already fit half) & $R$-base 70.0\% (21/30) vs.\ $G$-base 60.0\% (18/30)\\
\midrule
\multicolumn{3}{l}{\textbf{Oracle layer ($D_\mathrm{val}$, $n{=}65$, locked protocol)}}\\
\midrule
Oracle headroom@20\% vs.\ fixed expert & $+0.0056$ ($+3.9\%$ rel.) & $+0.0065$\\
Beats both (chosen / rejected direction) & 94\% / --- & 77\% / 35\%\\
\bottomrule
\end{tabular}
\end{table*}

\section{Dose--response probe}
\label{supp:dose}

Table~\ref{tab:dose} is the controlled information-removal probe behind
the third premise. Twenty real low-light images from the calibration fit
half were degraded inside a fixed mask (four rotating shapes, a target 10\% of
area) by five operators at three strengths ($s_1$--$s_3$; $s_0$ is the
untouched image); windows of side 32 and 64 were sampled inside the mask,
on its boundary and outside it, and the exact counterfactual generation
benefit ($-H$, $G$-base protocol) was recomputed per window. Reading:
noise flooding, blur and texture erasure raise the in-mask benefit by an
order of magnitude while the out-of-mask benefit stays flat, whereas
exposure truncation (``dark'') collapses toward zero and colour
quantisation never responds---the signal tracks \emph{recoverable}
information loss, not darkness, and is local.

\begin{table}[t]\centering
\caption{Dose--response of the exact counterfactual generation benefit
(mean over windows, LPIPS units, positive $=$ generation helps) against
degradation strength, LOL-v1 calibration fit half ($20$ images, window
sides $32/64$, $214$--$221$ in-mask and $240$ out-of-mask windows per
cell). Unsigned entries are positive.}
\label{tab:dose}
\scriptsize
\setlength{\tabcolsep}{3pt}
\begin{tabular}{lcccccc}
\toprule
 & \multicolumn{4}{c}{In-mask} & \multicolumn{2}{c}{Out-of-mask}\\
\cmidrule(lr){2-5}\cmidrule(lr){6-7}
Operator & $s_0$ & $s_1$ & $s_2$ & $s_3$ & $s_0$ & $s_3$\\
\midrule
Noise flooding & 0.0062 & 0.0321 & 0.0520 & 0.0618 & 0.0054 & 0.0057\\
Blur & 0.0032 & 0.0250 & 0.0230 & 0.0172 & 0.0045 & 0.0069\\
Texture erasure & 0.0042 & 0.0178 & 0.0187 & 0.0137 & 0.0049 & 0.0071\\
Exposure trunc.\ & 0.0055 & 0.0057 & 0.0040 & 0.0015 & 0.0063 & 0.0097\\
Colour quant.\ & 0.0012 & $-0.0058$ & 0.0008 & $-0.0001$ & 0.0062 & 0.0044\\
\bottomrule
\end{tabular}
\end{table}

The in-mask increment $s_3-s_0$ split by window side is $+0.050/+0.062$
(noise), $+0.012/+0.016$ (blur) and $+0.007/+0.013$ (texture) for
$32/64$; for the three responsive operators, boundary windows sit between the in- and out-of-mask values. The
probe measures the oracle label, not the router's prediction; the
darkness-cue result of Table~\ref{tab:premises} is the independent
deployment-side veto of the dark-shortcut reading.

\paragraph{Operator and mask settings.}
Each operator $D$ acts only inside the soft mask,
$x'=(1-M)\odot x+M\odot D(x)$; Table~\ref{tab:doseops} gives the exact
per-strength parameters. Mask shapes rotate through four
samplers---rectangle, ellipse, random-walk brush, smooth random
field---by (image, operator) index so no shape shortcut exists; binary
shapes are feathered by an 8\,px distance transform (the random field is
soft by construction), and the mask is held fixed across the three
strengths of a given (image, operator) pair. The realised soft-mask mass
averages 8.5\% of the image (7.8\% above the in-mask threshold). A
window counts as in-mask when its mean mask value exceeds 0.6, boundary
in $(0.1,0.6]$, out-of-mask at or below 0.1; at most six windows are
drawn per (image, operator, scale, zone), and the same windows are
re-scored on the undegraded expert outputs to give the $s_0$ baseline.
Mask, window and noise sampling use fixed seeds derived from 2026.

\begin{table}[t]\centering
\caption{Dose--response operators: definitions and per-strength
parameters (inputs in $[0,1]$; outputs clipped to $[0,1]$;
$\epsilon_i\sim\mathcal{N}(0,1)$ i.i.d.\ per pixel).}
\label{tab:doseops}
\scriptsize
\setlength{\tabcolsep}{2pt}
\begin{tabular}{llccc}
\toprule
Operator & Definition & $s_1$ & $s_2$ & $s_3$\\
\midrule
Noise flooding & $x+a\sqrt{x}\,\epsilon_1+b\epsilon_2$; $a/b$ & .05/.02 & .10/.04 & .20/.08\\
Blur & Gaussian, kernel ${\approx}6\sigma$ & $\sigma{=}1.5$ & 3.0 & 6.0\\
Texture erasure & $f{\times}$ down-/up-sample & $f{=}2$ & 4 & 8\\
Exposure trunc.\ & $x\cdot e$, zero below $t$; $e/t$ & .5/.01 & .25/.02 & .10/.04\\
Colour quant.\ & $L$ levels, chroma ${\times}c$ (YCrCb); $L/c$ & 16/.75 & 8/.5 & 4/.25\\
\bottomrule
\end{tabular}
\end{table}

\section{Budget--coverage curves, three bookkeeping conventions}
\label{supp:pareto}

Table~\ref{tab:pareto} gives the validation budget sweep of the
quality-gate router in the three LOL settings under the two bookkeeping
conventions that a post-freeze audit (2026-07-26) found to matter. The
\emph{published} column records the composite LPIPS at the moment the
covered area first crosses the budget; it is not guaranteed monotone and
LOLv2-synthetic is the counter-example ($0.04849\to0.04867\to0.04818$). The
\emph{per-image envelope} records, for each image, the best stopping point
on its own path below the budget---monotone by construction but
ground-truth-selected and therefore not a deployable policy. The
\emph{uniform-budget envelope} is the running minimum of the published
means over budgets, the best a single budget applied to every image can
do. The same audit found that the oracle column at 2.5\% silently fell
back to the unbudgeted oracle (the checkpoint grid starts at 5\%), so that
row is void. Under every convention the two claims used in the main text
survive: on LOL-v1, 2.5\% of area already beats the fixed expert
($0.10983<0.11103$); on LOLv2-real no budget beats it. What changes is the
shape: the frozen description of LOLv2-real as ``monotonically worsening
from 0.1464 to 0.1677'' is the first-crossing artefact, and the deployable
curve is flat at 0.14641 (per-image envelope $0.14641\to0.14591$).

\begin{table*}[t]\centering
\caption{Quality--coverage budget sweep on $D_\mathrm{val}$ ($n{=}65$ per setting, LPIPS$\downarrow$, quality-gate router) under three bookkeepings:
\emph{Pub.} $=$ value at first budget crossing (frozen, as published in the
matrix); \emph{Env.} $=$ per-image monotone envelope (GT-selected stopping
point, optimistic, not deployable); \emph{Unif.} $=$ uniform-budget
deployable envelope (running minimum of Pub.). Fixed expert: LOL-v1 All-$G$
0.11103; LOLv2-syn All-$R$ 0.04859; LOLv2-real All-$G$ 0.14411. The oracle
row at 2.5\% is void (checkpoint grid starts at 5\%). Rows above 30\% repeat
the 30\% row in every setting.}
\label{tab:pareto}
\scriptsize
\setlength{\tabcolsep}{3pt}
\begin{tabular}{l cccc cccc cccc}
\toprule
 & \multicolumn{4}{c}{LOL-v1 ($G$-base)} & \multicolumn{4}{c}{LOLv2-synthetic ($R$-base)} & \multicolumn{4}{c}{LOLv2-real ($G$-base)}\\
\cmidrule(lr){2-5}\cmidrule(lr){6-9}\cmidrule(lr){10-13}
Budget & Pub. & Env. & Unif. & Oracle & Pub. & Env. & Unif. & Oracle & Pub. & Env. & Unif. & Oracle\\
\midrule
2.5\% & 0.10983 & 0.10983 & 0.10983 & void & 0.04859 & 0.04859 & 0.04859 & void & 0.14641 & 0.14641 & 0.14641 & void\\
5\%   & 0.10877 & 0.10861 & 0.10877 & 0.10459 & 0.04849 & 0.04811 & 0.04849 & 0.04530 & 0.14910 & 0.14632 & 0.14641 & 0.13988\\
10\%  & 0.10731 & 0.10682 & 0.10731 & 0.10206 & 0.04867 & 0.04746 & 0.04849 & 0.04386 & 0.15513 & 0.14624 & 0.14641 & 0.13906\\
15\%  & 0.10605 & 0.10526 & 0.10605 & 0.10095 & 0.04818 & 0.04631 & 0.04818 & 0.04299 & 0.16157 & 0.14610 & 0.14641 & 0.13871\\
20\%  & 0.10576 & 0.10417 & 0.10576 & 0.10039 & 0.04833 & 0.04518 & 0.04818 & 0.04206 & 0.16767 & 0.14591 & 0.14641 & 0.13854\\
25\%  & 0.10576 & 0.10417 & 0.10576 & 0.10013 & 0.04833 & 0.04518 & 0.04818 & 0.04160 & 0.16767 & 0.14591 & 0.14641 & 0.13843\\
30\%  & 0.10576 & 0.10417 & 0.10576 & 0.10002 & 0.04833 & 0.04518 & 0.04818 & 0.04133 & 0.16767 & 0.14591 & 0.14641 & 0.13838\\
\midrule
Beats fixed expert & \multicolumn{4}{c}{yes at every budget} & \multicolumn{4}{c}{yes (Unif.\ 0.04859--0.04818, flat)} & \multicolumn{4}{c}{no budget}\\
\bottomrule
\end{tabular}
\end{table*}

\section{Deployment compositing algorithm}
\label{supp:composite}

Both router modes and the three map-based cue baselines compose through
\path{compose_scored} in \path{src/nerog/router/deploy.py}. Only the
quality gate can add a stage in front of it, and only on LOL-v1: a learned
verifier dropping every candidate whose predicted violation probability
exceeds $0.5$ ($0.35$\,s in Table~\ref{tab:cost}). The verifier is trained
per run and only the LOL-v1 run has one, so the other four final tests ran
with it disabled; the certified gate and the cue baselines hand over every
window in every setting. Otherwise compositing, budget accounting and color handling
are shared, so the equal-budget comparisons of Table~\ref{tab:premises}
compare scores, not compositors.

The window family and mask are those of Eq.~(1) of the main text;
\path{src/nerog/regions/windows.py} snaps the last row and column flush to
the border and builds $M$ as the outer product of two raised-cosine
profiles tapering over $\max(1,\mathrm{round}(0.25\,s/2))$ pixels per side,
so the outermost \emph{profile} sample is $0.146$ at $s{=}16$ and $0.0024$
at $s{=}128$ (the mask's corner samples are their squares). Given base
$y_\mathrm{base}$, override $y_\mathrm{insert}$, scored windows and budget
$b$ ($=0.20$ throughout):

\begin{enumerate}\itemsep0pt\parskip0pt
\item $C \leftarrow y_\mathrm{base}$ (a copy); $\mathit{cov}\leftarrow$
  all-false $H{\times}W$ mask; charged pixels $p\leftarrow0$; consecutive
  rejections $r\leftarrow0$; budget $B\leftarrow b\,H\,W$ pixels. Sort by
  score, descending; for each $(\sigma,w)$ in turn:
\item if a threshold $t$ is supplied and $\sigma \le t$, \textbf{stop} ---
  nothing later can pass;
\item let $\mathit{new}$ be the pixels of $w$'s $s\times s$ box still false
  in $\mathit{cov}$; if $p+\mathit{new} > B$, set $r \leftarrow r+1$ and
  \textbf{stop} if $r \ge 300$ (an approximation, not an exact fill; the
  LOL-v1 Gate~C cache predates this stop and reproduces only with it
  disabled), else \textbf{skip} $w$; otherwise $r \leftarrow 0$;
\item write $w$ by Eq.~(1) on its box, $y_\mathrm{insert}$ first shifted
  per color channel so its $M$-weighted mean matches the \emph{current} $C$
  there and clipped to $[0,1]$ --- the one step that can flatten local
  structure; set $\mathit{cov}[\mathrm{box}]\leftarrow\mathrm{true}$,
  $p\leftarrow p+\mathit{new}$, append $(\sigma,w)$.
\item Return $C$, the realized area $\mathrm{mean}(\mathit{cov})$ and the
  selection; if nothing was accepted --- e.g.\ an empty list --- $C$ is the
  untouched base and the area is $0$.
\end{enumerate}

In every evaluation here $t$ is unset and the caller filters first:
calibrated score ${>}0$ for the quality gate (raw only where
\path{--quality-raw} was passed, which differs by setting),
${>}\lambda^\star$ for the certified gate, no filter for the cue baselines.
Only the figure-materialization script exercises the threshold path, when
reproducing the frozen conformal composites.

Two properties govern the areas. The budget is charged on a boolean union
mask counting only newly-covered pixels (step~3), so overlapping windows
are free and the area returned is union coverage, never a sum and never
above $b$; and it is charged on the full $s\times s$ box, whose tapered rim
has almost no influence, so the accounting over-charges the generative
footprint. Coverage then typically undershoots the cap, sometimes by a lot.
Mean area at a 20\% budget on the five consumed final tests
(\path{outputs/logs/stage4/}), all images / touched subset alone (counts in
Sec.~\ref{supp:statscope}):

\begin{center}\scriptsize\setlength{\tabcolsep}{4pt}
\begin{tabular}{lcc}
\toprule
Setting ($n$) & Quality & Certified\\
\midrule
LOL-v1 (15) & 0.2000 / 0.2000 & 0.0474 / 0.0646\\
LOLv2-syn (100) & 0.1577 / 0.1877 & 0.0333 / 0.1332\\
LOLv2-real (100) & 0.2000 / 0.2000 & 0.0000 / ---\\
GSAD--CIDNet (100) & 0.1503 / 0.1503 & 0.0065 / 0.0813\\
SICE (40) & 0.1740 / 0.1740 & 0.0000 / ---\\
\bottomrule
\end{tabular}
\end{center}

\noindent The certified gate spends a small fraction of the cap, and its
two zeros are constructive abstentions (LOLv2-real,
Sec.~\ref{supp:reconcile}; SICE, Sec.~\ref{supp:sice}). These are
$D_\mathrm{test}$ means; the $5.3\%$ / 26 windows of Table~\ref{tab:cost}
is a separate measurement on its 20 $D_\mathrm{val}$ timing images.
Finally, the oracle bound and the equal-area random control drawn from the
same path share the compositing operator but not this loop: they re-score
marginal utility after each acceptance and record at the first budget
\emph{crossing} rather than staying under the cap (Sec.~\ref{supp:pareto}).

\section{Deployment cost}
\label{supp:cost}

Table~\ref{tab:cost} reports wall-clock latency and peak memory of the
single-process pipeline on an RTX~5080 laptop GPU at the native
$400\times600$ LOL-v1 resolution, averaged over the first 20 validation
images after warm-up. Because the base expert $G$ runs on the full image in
either case, the certified mode adds only the $R$ pass, the router and the
composite ($+0.10$\,s); the quality mode pays a further 0.35\,s for the
verifier, which is restricted to the high-score prefix of $3\times$ the
budget (the naive implementation that verified and composed every candidate
window took 42.97\,s per image at the same output quality: LPIPS
0.1200 vs.\ 0.1193 on the same 20 images). The certified mode therefore
adds only modest runtime overhead ($+0.10$\,s per image in this
implementation).

\begin{table}[t]\centering
\caption{Per-image deployment cost at $400\times600$ (RTX~5080 laptop,
20 $D_\mathrm{val}$ images, warm-up excluded). Seconds per component;
``over All-$G$'' is the latency added to the bare base expert. The naive
column verifies and composes every candidate window (no prefix restriction,
no incremental compositing) and is shown only to document the 42$\times$
engineering gain at unchanged quality.}
\label{tab:cost}
\scriptsize
\setlength{\tabcolsep}{3pt}
\begin{tabular}{lcccc}
\toprule
Seconds & All-$G$ bare & Quality & Certified & Naive quality\\
\midrule
$G$ pass (full image) & 0.53 & 0.53 & 0.53 & 0.56\\
$R$ pass (full image) & --- & 0.044 & 0.044 & 0.045\\
Router & --- & 0.048 & 0.040 & 0.039\\
Verifier & --- & 0.35 & --- & 9.29\\
Compose & --- & 0.049 & 0.016 & 33.04\\
\midrule
Total per image & 0.53 & 1.02 & \textbf{0.63} & 42.97\\
Over All-$G$ & --- & $+0.49$ & $+0.10$ & ---\\
Peak VRAM (MB) & $\sim$410 & 776 & 770 & 1011\\
Area / windows & --- & 17.2\% / 84 & 5.3\% / 26 & 20.0\% / 618\\
\bottomrule
\end{tabular}
\end{table}

\section{Distribution-Shift Stress Grid and the Input-Envelope Guard}
\label{supp:stress}

Table~\ref{tab:stress_full} is the full version of the stress grid
summarised in Sec.~6 of the main paper. The pre-registered grid applies a
full-image degradation to each of the 65 LOL-v1 validation inputs (clean,
plus noise/blur/darkening at three strengths: Poisson--Gaussian noise
with shot/read scales $(0.05,0.02)$, $(0.10,0.04)$, $(0.20,0.08)$;
Gaussian blur with $\sigma{=}1.5$, $3.0$, $6.0$; exposure scaling
$\times0.5$, $0.25$, $0.1$ with a $0.01$, $0.02$, $0.04$ floor cut) and
runs the \emph{frozen} LOL-v1 pipeline (v1 router, isotonic map,
verifier, $\lambda^\star{=}0.0197$ at $\alpha{=}0.10$, all calibrated on
clean $D_{\mathrm{cal}}$) with zero recalibration. The harm event is the
one used throughout the paper, LPIPS(routed) $>$ LPIPS(All-G on the
degraded input), both scored against the clean ground truth; intervals
are Clopper--Pearson 95\%. Override area is the mean image fraction on
which the certificate (or the quality mode) lets the restorer override
the generator, i.e.\ the abstention signal: an area near zero means the
certificate has withdrawn.

The guard is the three-statistic input-envelope detector of
Sec.~6 of the main paper: on the grey-scale deployment input $I$ it computes
the Laplacian energy $E{=}\operatorname{mean}[(\nabla^2 I)^2]$ and the
three statistics $S_1{=}\log_{10}E$ (log-Laplacian energy),
$S_2{=}\operatorname{mean}(I)$ (luminance) and
$S_3{=}\log_{10}[E/\operatorname{var}(I)]$ (contrast-normalised
high-frequency energy), and declares the input
out-of-envelope---forcing abstention, output $\equiv$ All-G---whenever
any statistic falls outside the min--max range observed on the 60 clean
low-light $D_{\mathrm{cal}}$ inputs ($S_1\in[-3.850,-2.215]$,
$S_2\in[0.0151,0.2745]$, $S_3\in[-1.364,0.874]$); it is label-free,
training-free, and never touches the validation set. The guard columns
reuse the frozen per-image records of the stress run (no routing is
recomputed), with the degraded inputs regenerated bit-exactly from the
registered seeds.

\textbf{Reading.} Noise collapses the calibrated scores below
$\lambda^\star$, so the certificate abstains on its own (area
0.004--0.006) and harm stays near $\alpha$ even before the guard; blur
and darkening preserve the scores (area up to 0.092, above clean) while
moving the utility landscape, and this is where the bound breaks (up to
56.9\%). The guard flags every blur and dark-s3 image and every noise
image at $s\geq2$, driving post-guard harm to 0/65 in all six of those
cells, at a clean false-alarm cost of 4/65 and a retained certified gain
of $0.00286/0.00316{=}90.5\%$. The single residual cell is dark~s1
($\times0.5$ exposure): clean low-light luminance already spans
$0.015$--$0.274$, so most halved inputs stay inside the envelope (21/65
flagged) and 13/65 (20.0\%) harm survives---an input-marginal detector
cannot separate a shift whose support overlaps the calibration
distribution. The uncertified quality mode is harmed on 24.6--98.5\% of
shifted inputs and is dominated by the certified mode in every row.

\textbf{Scope of the guard.} Two limits are not visible from the table.
First, the guard was designed after the grid it repairs: the stress grid
was run on 2026-07-08 and the detector on 2026-07-09, against the
failure modes that grid had just exposed---$S_1$ for the blur and noise
cells, $S_2$ for the darkening cells, $S_3$ as a more content-invariant
blur cue. Its design and endpoints were committed before it was run, but
the statistics were not chosen blind to the failures they close. Second,
it is evaluated only on the three families it was designed against: no
held-out shift family is ever put through it---no compression artefact, no
colour cast, no mixed corruption. The one real domain change in this paper,
SICE (Sec.~\ref{supp:sice}), is a separately calibrated setting, not a test
of this guard. The supported claim is correspondingly narrow: a label-free,
input-marginal envelope calibrated on 60 clean calibration inputs can
restore the bound on eight of the nine shifted cells. It is not evidence of
a general out-of-distribution detector, and the dark~s1 residual is the
standing reason to expect it is not one.


\begin{table*}[t]
  \centering
  \scriptsize
  \setlength{\tabcolsep}{3pt}
  \caption{Full stress grid (LOL-v1 validation, $n{=}65$ per row, frozen
  pipeline, zero recalibration) with the input-envelope guard. LPIPS
  columns are means over the 65 images; ``gain'' is All-G minus
  certified (positive $=$ certified better). Harm and detection rates are
  counts over 65, with Clopper--Pearson 95\% intervals in percent;
  ``area'' is mean override area. Post-guard harm counts harmed images
  among those the guard did not flag. Certified gain retained after the
  guard is defined only on the clean row: $0.00316\to0.00286$ (90.5\%).}
  \label{tab:stress_full}
  \begin{tabular}{@{}lrrr lr lr ll@{}}
    \toprule
    & \multicolumn{3}{c}{Mean LPIPS $\downarrow$} & \multicolumn{2}{c}{Certified mode} & \multicolumn{2}{c}{Quality mode} & \multicolumn{2}{c}{Input-envelope guard} \\
    \cmidrule(lr){2-4}\cmidrule(lr){5-6}\cmidrule(lr){7-8}\cmidrule(lr){9-10}
    Condition & All-G & Cert. & Gain & Harm [95\% CI] & Area & Harm & Area & Detected & Post-guard harm [95\% CI] \\
    \midrule
    clean    & 0.1111 & 0.1080 & $+0.0032$ & 4/65 (6.2\%) [1.7,15.0]    & 0.043 & 12/65 (18.5\%) & 0.179 & 4/65 (6.2\%)   & 3/65 (4.6\%) [1.0,12.9] \\
    \midrule
    noise s1 & 0.3951 & 0.3954 & $-0.0002$ & 5/65 (7.7\%) [2.5,17.0]    & 0.006 & 56/65 (86.2\%) & 0.184 & 33/65 (50.8\%) & 1/65 (1.5\%) [0.0,8.3] \\
    noise s2 & 0.6142 & 0.6145 & $-0.0003$ & 3/65 (4.6\%) [1.0,12.9]    & 0.004 & 63/65 (96.9\%) & 0.192 & 65/65 (100\%)  & 0/65 (0.0\%) [0.0,5.5] \\
    noise s3 & 0.7230 & 0.7245 & $-0.0015$ & 7/65 (10.8\%) [4.4,20.9]   & 0.005 & 64/65 (98.5\%) & 0.195 & 65/65 (100\%)  & 0/65 (0.0\%) [0.0,5.5] \\
    \midrule
    blur s1  & 0.2818 & 0.2862 & $-0.0044$ & 37/65 (56.9\%) [44.0,69.2] & 0.050 & 60/65 (92.3\%) & 0.162 & 65/65 (100\%)  & 0/65 (0.0\%) [0.0,5.5] \\
    blur s2  & 0.4470 & 0.4483 & $-0.0013$ & 27/65 (41.5\%) [29.4,54.4] & 0.028 & 53/65 (81.5\%) & 0.137 & 65/65 (100\%)  & 0/65 (0.0\%) [0.0,5.5] \\
    blur s3  & 0.6252 & 0.6251 & $+0.0001$ & 7/65 (10.8\%) [4.4,20.9]   & 0.012 & 16/65 (24.6\%) & 0.089 & 65/65 (100\%)  & 0/65 (0.0\%) [0.0,5.5] \\
    \midrule
    dark s1  & 0.2400 & 0.2361 & $+0.0038$ & 18/65 (27.7\%) [17.3,40.2] & 0.087 & 23/65 (35.4\%) & 0.173 & 21/65 (32.3\%) & 13/65 (20.0\%) [11.1,31.8] \\
    dark s2  & 0.5514 & 0.5481 & $+0.0033$ & 10/65 (15.4\%) [7.6,26.5]  & 0.092 & 19/65 (29.2\%) & 0.167 & 56/65 (86.2\%) & 3/65 (4.6\%) [1.0,12.9] \\
    dark s3  & 0.8415 & 0.8414 & $+0.0001$ & 21/65 (32.3\%) [21.2,45.1] & 0.031 & 25/65 (38.5\%) & 0.117 & 65/65 (100\%)  & 0/65 (0.0\%) [0.0,5.5] \\
    \bottomrule
  \end{tabular}
\end{table*}

\section{SICE cross-domain setting in full}
\label{supp:sice}

\paragraph{Data and roles.}
SICE Part2 has 229 multi-exposure sequences; one seed-2026 shuffle
assigned them to $D_{\mathrm{select}}$\,40 / $D_{\mathrm{cal}}$\,104
(pre-registered fit-half 28 + cert-half 76) / $D_{\mathrm{val}}$\,45 /
$D_{\mathrm{test}}$\,40, mutually exclusive, and the router was trained
on the 360 Part1 train sequences (minimum-exposure frame). The $-1$EV
input frame follows the dataset's own index file where it is defined
(7 frames $\to$ frame 3, 9 frames $\to$ frame 4; 210 sequences) and a
project proxy (frame $\lfloor N/2\rfloor$; 19 sequences) otherwise;
images are resized to short side 400 and centre-cropped to long side
$\min(600,\cdot)$; all expert weights are the frozen LOL-family
checkpoints (CIDNet LOLv2-real, DiffLL LOL-v1; no SICE fine-tuning).
The ground truth is the sequence-level multi-exposure-fusion reference,
so \emph{SICE LPIPS values are not comparable with the pixel-paired LOL settings} and are tabled separately.

\paragraph{Table~\ref{tab:sice}.}
Panel (a) is the pre-registered pair diagnostic on the 40 selection
images only (whole-image selection-oracle gain, $\mathrm{mean}\,\min(A,B)$
vs.\ the stronger single expert, and per-image win rates): the rule
``largest gain with both win rates $\ge25\%$'' selects CIDNet+DiffLL
(gain $0.0264=9.2\%$, a whole-image ceiling rather than a routing gain)
and CIDNet+GSAD has gain exactly 0---with the caveat that those GSAD
outputs came from the deterministic wrapper path later found to run in
the degraded basin on LOLv2-real (Sec.~\ref{supp:gsad}), so the
rejection concerns those outputs on those 40 images, not the domain.
Panel (b) is the endpoint chain from the pre-registration and the W3
verdict, including the one exercise of the grid-cap clause (cap
$0.03\to0.045$ after 11/76 infinite scores), the certificate
$\lambda^\star=0.04307$, the validation audit (harm 0/45, capture 0),
and the post-freeze feasibility audit showing that zero coverage is
both constructive (largest isotonic knot $0.04289<\lambda^\star$) and
correct (every feasible threshold is harmful or gainless). Panel (c) is
the sealed one-shot test: certified output is identical to fixed CIDNet
on 40/40 images, so validation-time abstention generalises exactly,
while raw mode's $+0.0010$ has a CI crossing zero; under the
interpretation lock nothing was recalibrated or rerun.

\begin{table*}[t]
\centering
\scriptsize
\setlength{\tabcolsep}{3pt}
\caption{The SICE Part2 $-1$EV setting in full. (a)~Pair selection on
$D_{\mathrm{select}}$ (40 images; mean LPIPS of the five single experts:
CIDNet 0.2884, DiffLL 0.3083, Diff-Retinex 0.3821, Retinexformer 0.3947,
GSAD 0.4123). (b)~Pre-registered endpoint chain. (c)~One-shot
$D_{\mathrm{test}}$ ($n=40$). SICE LPIPS uses sequence-level fusion ground
truth and is not comparable with the LOL tables.}
\label{tab:sice}
\begin{tabular}{@{}llccccc@{}}
\multicolumn{7}{@{}l}{\textbf{(a) Pair selection on $D_{\mathrm{select}}$ (LPIPS$\downarrow$)}}\\
\toprule
Pair $A$+$B$ & & Stronger single & $\mathrm{mean}\min(A,B)$ & Gain & Rel.\ gain & Wins $A$/$B$ \\
\midrule
\textbf{CIDNet+DiffLL} & \textbf{selected} & 0.2884 & 0.2620 & 0.0264 & 9.2\% & 55\%/45\% \\
Diff-Retinex+Retinexformer & & 0.3821 & 0.3568 & 0.0252 & 6.6\% & 52\%/48\% \\
GSAD+Retinexformer & & 0.3947 & 0.3750 & 0.0198 & 5.0\% & 42\%/57\% \\
Diff-Retinex+GSAD & & 0.3821 & 0.3635 & 0.0185 & 4.9\% & 60\%/40\% \\
DiffLL+GSAD & win-rate clause fails & 0.3083 & 0.3013 & 0.0070 & 2.3\% & 82\%/18\% \\
Diff-Retinex+DiffLL & win-rate clause fails & 0.3083 & 0.3021 & 0.0063 & 2.0\% & 12\%/88\% \\
CIDNet+Retinexformer & win-rate clause fails & 0.2884 & 0.2832 & 0.0052 & 1.8\% & 85\%/15\% \\
DiffLL+Retinexformer & win-rate clause fails & 0.3083 & 0.3042 & 0.0042 & 1.4\% & 78\%/22\% \\
CIDNet+Diff-Retinex & win-rate clause fails & 0.2884 & 0.2850 & 0.0035 & 1.2\% & 95\%/5\% \\
CIDNet+GSAD & rejected & 0.2884 & 0.2884 & 0.0000 & 0.0\% & 100\%/0\% \\
\bottomrule
\end{tabular}

\vspace{4pt}
\begin{tabular}{@{}p{0.11\textwidth}p{0.09\textwidth}p{0.60\textwidth}p{0.13\textwidth}@{}}
\multicolumn{4}{@{}l}{\textbf{(b) Endpoint chain} (R-base: CIDNet base, DiffLL overrides; $\alpha=0.10$, $\delta=0.05$, $\varepsilon=0$, budget 20\%)}\\
\toprule
Endpoint & Split ($n$) & Observation & Verdict \\
\midrule
E1 direction & $D_{\mathrm{select}}$ (40) & R-base: base 0.2884, oracle@20\% 0.2707, headroom \textbf{0.01772}; G-base: 0.3083 / 0.2892 / 0.01911; both $\ge0.002$; absolute-quality clause picks R-base & PASS \\
Router & $D_{\mathrm{train}}$ (360) & seed 2026, batch 1 (variable image sizes), 150 epochs, lr $3{\times}10^{-4}$; hold-out Spearman 0.585 & --- \\
E2 cues & $D_{\mathrm{val}}$ (45) & router beats darkness / residual / uncertainty / random on 80.0 / 73.3 / 77.8 / 80.0\% (each $\ge70\%$); raw router 0.2484 vs.\ base 0.2503 ($+0.0018$, 68.9\% images better, capture 13.4\%); random 0.2544; oracle@20\% 0.2366 & PASS \\
E3 certificate, round 1 & cert-half (76) & grid cap 0.03 $<$ max calibrated score 0.0429 $\Rightarrow$ 11/76 $\tilde\lambda_i=+\infty$, certificate abstains (grid artefact) & clause exercised \\
E3 certificate, round 2 & cert-half (76) & cap 0.045: $\lambda^\star=\mathbf{0.04307}$ ($k=70$/76, $n_\infty=0$); LTT: $0.045$ $V{=}0$ $p{=}3.3{\times}10^{-4}$, $0.04307$ $V{=}0$, $0.04113$ $V{=}7$ $p{=}0.505$ $\Rightarrow$ LTT $\lambda^\star=0.04307$ & issued \\
E3 audit & $D_{\mathrm{val}}$ (45) & harm \textbf{0/45}, CP95 $[0,0.0787]$; capture 0; certified $\equiv$ CIDNet (0.2503 $=$ 0.2503) & PASS (tier $k\le5$) \\
Feasibility (post-freeze) & fit-half knots & max isotonic knot $0.04289<\lambda^\star$: zero coverage is constructive; first feasible $\lambda=0.04113$: 17/45 active, gain $-0.00009$; best-gain $\lambda=0.00824$: $+0.00214$, harm 14/45 $=31\%$ & abstention correct \\
E4 upgrade & $D_{\mathrm{val}}$ (45) & certified gain 0 (no overrides); raw $+0.0018$, capture 13.4\%, significance not tested & FAIL (certified) \\
$\alpha$-sweep (frozen) & $D_{\mathrm{val}}$ (45) & safe-suffix $\lambda^\star$ at $\alpha=.05/.10/.15/.20/.30$: 0.04307 / 0.04307 / 0.0392 / 0.02759 / 0.01598; harm 0/45 at $\le0.10$; 7/45 (15.6\%) at 0.15 and 0.20, 9/45 (20\%) at 0.30 & context only \\
Replay gate & $D_{\mathrm{val}}$ (45) & frozen executor replays raw $l_{\mathrm{router}}$ and certified $\lambda=0.04307$ curve with $\max|\Delta|=0$ (45/45, tolerance $10^{-6}$) & PASS \\
E5 unlock & --- & E2$\wedge$E3(PASS); frozen command, run once; $D_{\mathrm{test}}$ ids had zero overlap with any prior expert output & consumed \\
\bottomrule
\end{tabular}

\vspace{4pt}
\begin{tabular}{@{}lccc@{}}
\multicolumn{4}{@{}l}{\textbf{(c) One-shot $D_{\mathrm{test}}$, $n=40$} (frozen command; interpretation lock: no recalibration, no rerun)}\\
\toprule
System & PSNR$\uparrow$ & SSIM$\uparrow$ & LPIPS$\downarrow$ \\
\midrule
Fixed CIDNet (base, All-R) & 15.92 & 0.520 & 0.2534 \\
Fixed DiffLL (All-G) & 16.20 & 0.523 & 0.2605 \\
Whole-image selector (GT-referenced) & 16.56 & 0.534 & 0.2305 \\
Router raw & 15.97 & 0.520 & 0.2524 \\
\textbf{Router certified} & \textbf{15.92} & \textbf{0.520} & \textbf{0.2534} \\
Window oracle@20\% (upper bound) & 15.97 & 0.523 & 0.2390 \\
\midrule
\multicolumn{4}{@{}l}{Certified $\equiv$ CIDNet on 40/40 images: capture 0, harm 0/40, paired difference identically 0.}\\
\multicolumn{4}{@{}l}{Raw vs.\ fixed: $+0.0010$, CI95 $[-0.0043,+0.0053]$ (crosses zero), 65\% of images improve.}\\
\bottomrule
\end{tabular}
\end{table*}

\section{GSAD process-state bistability: investigation summary}
\label{supp:gsad}
\paragraph{The modern pair at a glance.}
Table~\ref{tab:pairtimeline} joins what the main text spreads across its
Secs.~4--6: three CIDNet~\cite{cidnet}+GSAD~\cite{gsad} settings, two GSAD
process states, one routing-direction flip, one consumed final test. The flip
is what the unchanged registered \emph{direction} rule returned once GSAD had
been regenerated in the official basin, where it beats CIDNet on both LPIPS
and PSNR over the disjoint 125-image $D_{\mathrm{cal}}\cup D_{\mathrm{val}}$
pool. Neither failing state consumed a final test, so exactly one
pre-registered final-test endpoint was consumed on this pair; the basin gates
and probes of Table~\ref{tab:gsad} read those same images without consuming an
endpoint, and are itemized as non-endpoint reads in Sec.~\ref{supp:ledger}.
The last row is a \emph{different} pair, included only because its 11/65 is
routinely read as the modern pair's 7/65.

\begin{table*}[t]
\centering
\scriptsize
\setlength{\tabcolsep}{3pt}
\caption{The CIDNet~\cite{cidnet}+GSAD~\cite{gsad} modern pair in one view.
Rows 1--3 are that pair; row~4 is a \emph{different} pair
(\texttt{v1\_cid}, CIDNet+DiffLL~\cite{diffll}), whose 16.9\% safe-suffix
failure is the one repaired by valid LTT in the main text and is frequently
mistaken for row~1's 10.8\%. $\lambda^\star$ is the deployed safe-suffix
operating point ($\alpha{=}0.10$, $\varepsilon{=}0$), fitted on the 30/30
calibration split; harm is measured on the $n{=}65$ validation set. Every
cell is joined from Tables~\ref{tab:gsad} and~\ref{tab:provenance} and the
frozen certificate records; nothing here is newly computed.}
\label{tab:pairtimeline}
\begin{tabular}{@{}p{0.09\textwidth}p{0.12\textwidth}p{0.14\textwidth}p{0.15\textwidth}p{0.19\textwidth}p{0.20\textwidth}@{}}
\toprule
Setting (record \texttt{line}) & GSAD process state & Routing direction (base $\to$ candidate) & Pre-registration (date, commit) & $D_{\mathrm{val}}$ rule, operating point, harm & Pre-registered final test \\
\midrule
\texttt{v1\_cgr} (first) &
degraded, unrecognised (frozen outputs) &
CIDNet ($R$) $\to$ GSAD ($G$) &
2026-07-12, \texttt{97bb770} &
binary; $\lambda^\star{=}0.00691$; \textbf{7/65 $=$ 10.8\%} FAIL, target exceeded by one image &
not consumed \\
\addlinespace[1pt]
\texttt{v1\_cgr\_det} (second) &
still degraded; deterministically regenerated &
CIDNet ($R$) $\to$ GSAD ($G$), unchanged &
2026-07-16, \texttt{2c3862c} &
three-tier (binary alongside); $\lambda^\star{=}0.00691$; \textbf{4/65 $=$ 6.2\%} PASS &
not consumed: unlocked, then aborted at the D3 basin gate (2026-07-17) \\
\addlinespace[1pt]
\texttt{v1\_cgr\_off} (third) &
official (published) basin; batch gate PASS &
\textbf{GSAD ($G$) $\to$ CIDNet ($R$)}: flipped by the registered rule &
2026-07-17, \texttt{f07ca37} &
three-tier; $\lambda^\star{=}0.01333$; \textbf{3/65 $=$ 4.6\%} PASS (LTT $0.01846$: 2/65) &
consumed once, 2026-07-17 ($n{=}100$): 0/100 harm, $+0.00023$ LPIPS; a disclosed audit outcome \\
\midrule
\multicolumn{6}{@{}l}{\emph{A different pair, not the modern pair:}}\\
\texttt{v1\_cid} &
n/a (no GSAD here) &
CIDNet ($R$) $\to$ DiffLL ($G$) &
2026-07-12, \texttt{0693efb} &
binary; $\lambda^\star{=}0.01333$; \textbf{11/65 $=$ 16.9\%} FAIL; LTT repair at $0.01717$: 3/65 $=$ 4.6\% &
not consumed; published as a negative result \\
\bottomrule
\end{tabular}
\end{table*}

Table~\ref{tab:gsad} lists, in chronological order, every controlled
experiment behind the reproducibility hazard of the main text---the
single-variable process controls E1--E15, the deterministic probe D1,
the 9-cell regeneration-equivalence audit D1.5, the deferred
wrapper-vs-official comparison D3 that exposed the shared wrong basin,
the official-form probes, the batch-level basin gate that replaced the
first-image self-check, and the consumption gate D3$'$ with its
pre-registered amendment---each taken from the dated protocol, verdict
or amendment report and its JSON. No root cause is claimed, only the
three observational coordinates registered with the official-basin setting: the official \texttt{test.py} form (cuDNN autotuning on) stays in
the published basin, the deterministic batch loop lands in the degraded
basin, and the basin can flip mid-process. The reading: a passing
first-inference self-check and a passing self-consistency audit were
both satisfied while every batch output sat in the wrong basin, which
is why the frozen setting (E3 7/65 FAIL) and the deterministic
regeneration (E3 4/65 PASS) never consumed $D_{\mathrm{test}}$ and the
official-basin setting was re-registered from the direction call onward
(flipped to G-base; $\lambda^\star=0.01333$, 3/65 on validation,
0/100 harm and $+0.00023$ on the one-shot test). The lessons recorded
with the audit: published pipeline numbers can depend on undeclared
process state; a self-check at process start proves nothing about the
batch; equivalence audits against one's own frozen outputs cannot
detect a shared wrong basin; expert-level distributional gates against
an external reference belong in evaluation protocols.

\begin{table*}[t]
\centering
\scriptsize
\setlength{\tabcolsep}{3pt}
\caption{GSAD~\cite{gsad} process-state bistability: controlled
experiments, audits and gates in chronological order. LPIPS vs.\ GT on
LOLv2-real unless stated; ``official'' = the unmodified official
\texttt{test.py} process form; ``wrapper'' = our batch generator in
deterministic configuration (\texttt{cudnn.deterministic}, no
autotuning) with a first-image self-check (00690 $<0.15$). Seed-noise
floor 0.00673 = cross-seed per-image $|\Delta\mathrm{LPIPS}|$ p95
measured on the calibration fit-half in the (then unrecognised)
degraded basin.}
\label{tab:gsad}
\begin{tabular}{@{}p{0.13\textwidth}p{0.09\textwidth}p{0.50\textwidth}p{0.21\textwidth}@{}}
\toprule
Step (date) & Images & Variable controlled / comparison and observation & Outcome \\
\midrule
E1--E15 (07-12) & 00690, $>$20 processes & One factor per independent process: I/O path, memory footprint, \texttt{cv2.cvtColor}, \texttt{ToTensor}, global seed, \texttt{cudnn.benchmark} on/off, scheduler call count, GT feeding, input source, \texttt{manual\_seed} timing, import order, argparse, early yaml import, function scope, \texttt{nerog} import chain; inputs, weights, RNG state and scheduler buffers hash-identical across basins & no single factor flips the basin; refactored wrapper class reproducibly degraded \\
Same-basin chaos check & LOL-v1 eval15 & wrapper recomputation vs.\ official run within the published basin: mean 0.1037 vs.\ 0.1036 (trajectories chaotic, metric equivalent) & metric-equivalent \\
D1 probe (07-12) & 00690 & script vs.\ class form $\times$ deterministic off/on, each an independent process with 2 in-process repeats (8 samples): 0.0996, 0.0996 / 0.0995, 0.0995 in all four cells & all published basin; degraded reproducer no longer available, so ``eliminated'' not claimed \\
D1.5 audit (07-12) & 689 $\times$ 3 seeds & deterministic regeneration vs.\ frozen outputs, 9 role$\times$seed cells ($D_{\mathrm{train}}$ 564 / $D_{\mathrm{cal}}$ 60 / $D_{\mathrm{val}}$ 65); criterion $|\Delta\mathrm{mean}|\le0.002$; observed $\le0.00001$, per-image p95 0.00011--0.00014 ($\approx1/50$ of the seed floor); self-check 3/3 at 0.0995 & GO (9/9); both sides later shown degraded (means 0.275--0.294) \\
D3 comparison (07-17) & official test 100 & wrapper (det, seed 1) vs.\ official: mean \textbf{0.2592 vs.\ 0.1131}, gap 0.1461 ($29\times$ the 0.005 tolerance, $22\times$ the seed floor); per-image $|\Delta|$ median / p95 / max 0.1432 / 0.2439 / 0.3414 & FAIL; $D_{\mathrm{test}}$ endpoint not consumed \\
Isolation (07-17) & 00690 & one process, three inferences with two bit-identical input constructions: all 0.1886; the earlier $D_{\mathrm{test}}$-generation process had self-check 0.0995 then loop output 0.1886 for the same image & basin flipped mid-process; self-check gate constrains first inference only \\
Official-form probe (07-17) & 10 train-pool & official \texttt{test.py} on train-pool scenes: mean 0.0625 (10/10 published basin) vs.\ wrapper 0.2963 & with the 100 test images: 110/110 published basin \\
Reseed probe (07-17) & 10 $\times$ 4 outputs & official form + per-image reseeding (seeds 1/2/3 + seed-1 repeat): 40/40 published basin (mean 0.0617, max 0.1137); cross-seed $|\Delta|$ max 0.0048; same seed across processes not bit-identical (autotuning) but $|\Delta\mathrm{LPIPS}|\le6.8{\times}10^{-5}$ & official form adopted for production \\
Batch basin gate (07-17) & 689 $\times$ 3 seeds & pre-registered criteria on all production outputs: per-seed mean $\le0.15$ and fraction with LPIPS $>0.25$ $\le5\%$; observed means 0.10395 / 0.10457 / 0.10485, tails 3.19 / 3.19 / 3.63\% & PASS (replaces first-image self-check) \\
Official-basin seed noise & 689 & same-scheme seed pairs in the published basin: p95 0.00794 (s1--s2), 0.01226 (s1--s3), 0.00879 (s2--s3) & floor itself $\approx0.012$ \\
D3$'$ original gate (07-17) & $D_{\mathrm{test}}$ 100 & official-form seed-1 output vs.\ the SOTA official run: mean gap $5.1{\times}10^{-5}$ ($\le0.005$ (pass)); per-image p95 \textbf{0.01438} vs.\ criterion 0.0135 ($=2\times$ seed floor) (fail) & FAIL (stands as recorded) \\
D3$'$ amendment (07-17, committed before the run) & $D_{\mathrm{test}}$ 100 & yardstick recalibrated: official form rerun with global seed 4357$\to$4358 on the same 100 images gives $\mathrm{p95_{ref}}=\mathbf{0.01516}$; revised criterion p95 $\le\max(0.0135,1.25\,\mathrm{p95_{ref}})=\mathbf{0.01895}$; $0.01438\le0.01895$ & PASS; both verdicts reported \\
G5 control (07-25) & official test 100 & deterministic negative control with HVI-CIDNet, wrapper vs.\ official evaluator (Table~\ref{tab:g5}) & PASS: harness not the culprit \\
\bottomrule
\end{tabular}
\end{table*}

\paragraph{Deterministic negative control.}
Table~\ref{tab:g5} asks whether our batch harness itself perturbs
expert outputs, by running a deterministic expert (HVI-CIDNet,
LOLv2-real checkpoint) through the same wrapper path and through the
official evaluator on the same 100 LOLv2-real test images, with
criteria pre-registered before the run. The two paths agree to
$6.8{\times}10^{-5}$ mean LPIPS, about $2100\times$ below the GSAD basin
gap of 0.1461, and the per-image p95 of $8.1{\times}10^{-4}$ is $8\times$
below GSAD's own seed-noise floor; the bistability is a property of the
sampler's process state, not of the repository's wrapping.

\begin{table}[t]
\centering
\scriptsize
\setlength{\tabcolsep}{3pt}
\caption{G5 deterministic negative control: HVI-CIDNet~\cite{cidnet}
through our batch wrapper vs.\ the official evaluator on the 100
LOLv2-real official test images (criteria pre-registered). Mean-LPIPS
gap $6.8{\times}10^{-5}$; per-image statistics are over the 100 images.
Reference scales: GSAD basin gap 0.1461, GSAD seed-noise floor p95 0.00673.}
\label{tab:g5}
\begin{tabular}{@{}lccc@{}}
\toprule
Quantity & Wrapper & Official & Criterion \\
\midrule
Mean LPIPS$\downarrow$ & 0.122007 & 0.121939 & $|\Delta|<10^{-3}$ (pass) \\
$|\Delta\mathrm{LPIPS}|$ median & \multicolumn{2}{c}{$2.5{\times}10^{-4}$} & --- \\
$|\Delta\mathrm{LPIPS}|$ p95 & \multicolumn{2}{c}{$8.07{\times}10^{-4}$} & $<5{\times}10^{-3}$ (pass) \\
$|\Delta\mathrm{LPIPS}|$ max & \multicolumn{2}{c}{$1.67{\times}10^{-3}$} & --- \\
Mean PSNR$\uparrow$ & 23.888 & 23.904 & --- \\
$|\Delta\mathrm{SSIM}|$ p95 & \multicolumn{2}{c}{$1.96{\times}10^{-3}$} & --- \\
\bottomrule
\end{tabular}
\end{table}

\section{Additional qualitative cases}
\label{supp:cases}
\begin{figure*}[t]
  \centering
  \includegraphics[width=\textwidth]{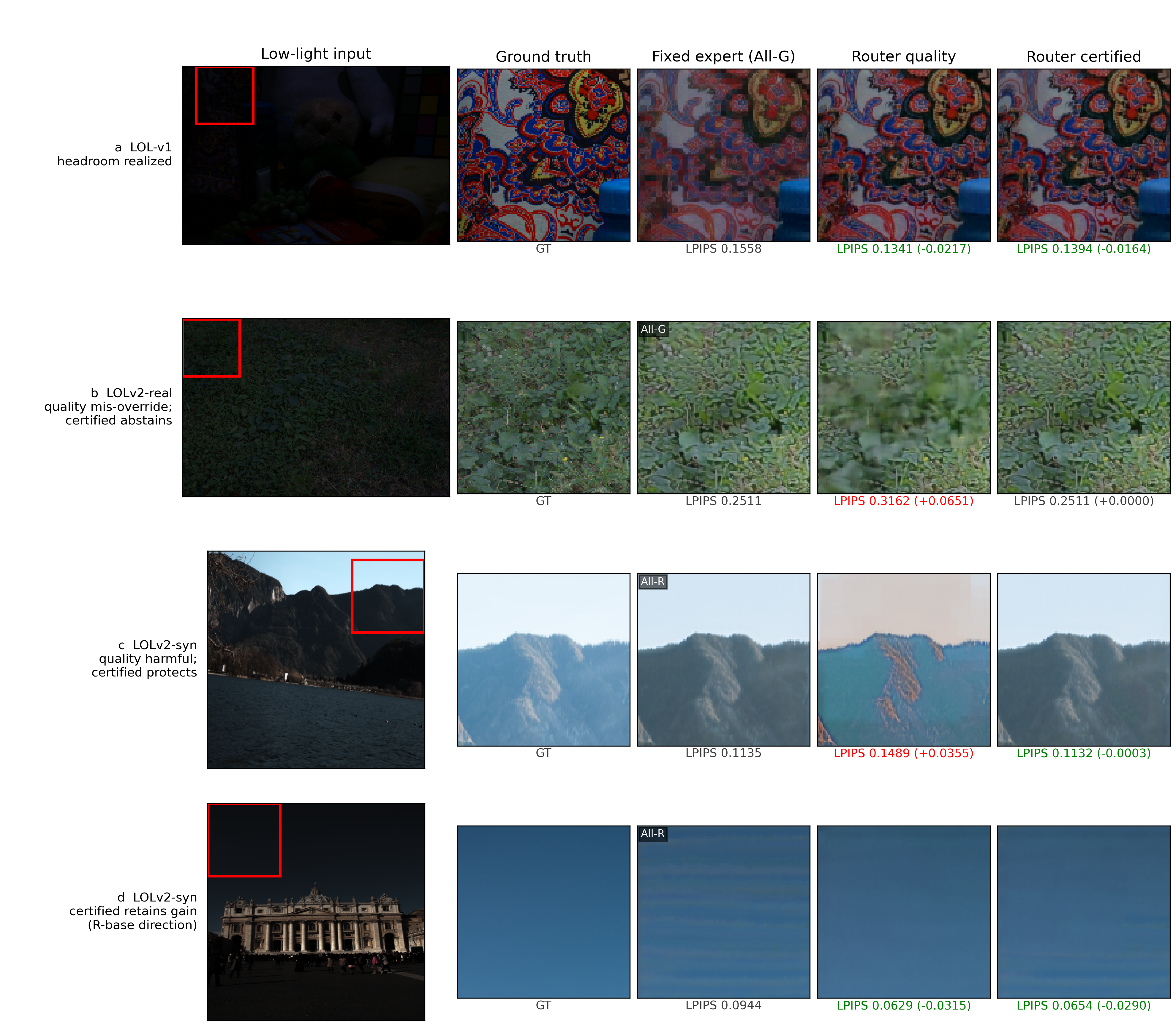}
  \caption{The four certified behaviours on final-test images. (a)~LOL-v1,
  headroom realized: both modes improve. (b)~LOLv2-real, the quality mode
  mis-overrides by $+0.065$ LPIPS while the certificate abstains and
  returns the base expert bit-exactly ($\pm0.0000$). (c)~LOLv2-synthetic,
  the quality mode is harmful ($+0.036$, visible colour shift) and the
  certificate protects. (d)~LOLv2-synthetic, certified gain retained.
  Cases were chosen to illustrate the four behaviours, one per behaviour;
  crops are the automatic maximum-difference window.}
  \label{fig:cases_supp}
\end{figure*}

Figure~\ref{fig:cases_supp} shows the four certified behaviours on
final-test images of the three earliest settings (LPIPS read from the frozen
final-test records, not recomputed).

\section{Pre-registration timeline and per-setting protocol provenance}
\label{supp:provenance}

Table~\ref{tab:provenance} lists, for each of the eight experimental settings,
the document whose commit timestamps the pre-registration, the
$\lambda$-grid cap, the fit-half/cert-half split of the calibration set,
the two certified operating points, the validation verdict rule, and
whether the pre-registered final test was consumed. All eight settings share
the grid $\{0\}\cup\mathrm{round}(\mathrm{linspace}(5\times10^{-4},
\lambda_{\max},24),5)$ ($K{=}25$ points), $\alpha{=}0.10$,
$\varepsilon{=}0$, $\delta{=}0.05$, budget 20\%, and the fit-half /
cert-half rule (direction oracle, isotonic fitting and noise floors on
the first half only; conformal calibration on the second). Reading:
every setting after v1 was registered before its first downstream
computation, and no final-test endpoint was consumed more than once. v1 is
labelled \emph{exploratory-then-ratified} because its grid commit
(153b87d, 2026-07-05) already carried cert-half results and an earlier
commit (2f94be5) fitted isotonic on the full calibration set, so the
2026-07-12 certificate-v2 programme (b9a8798) ratified, rather than
pre-registered, its operating point $\lambda^\star{=}0.01974$. SICE is the
only setting on which the pre-registered cap-correction clause was exercised:
the default cap 0.03 sat below the largest calibration score (0.0429),
11 of the 76 cert-half images received $\tilde\lambda_i{=}{+}\infty$ and
the first round abstained as a grid artefact; the clause re-ran the
certificate at 0.045; both rounds were executed and logged (the first-round abstention and the second-round pass appear in sequence in the chain log), and the released certificate file holds the second round. Deployed
behaviour is identical under either cap (certified $\equiv$ fixed expert
on $D_{\mathrm{val}}$ and on $D_{\mathrm{test}}$).

\begin{table*}[t]
\centering
\scriptsize
\setlength{\tabcolsep}{3pt}
\caption{Per-setting protocol provenance. Fit/cert = fit-half / cert-half
sizes of the calibration set; $\lambda^\star_{\mathrm{ss}}$ /
$\lambda^\star_{\mathrm{LTT}}$ = safe-suffix / valid-LTT operating points
at $\alpha{=}0.10$, $\varepsilon{=}0$; verdict = the registered
$D_{\mathrm{val}}$ safety rule and its outcome (binary: empirical harm
$\le\alpha$; three-tier at $n{=}65$: PASS $k\le7$, INCONCLUSIVE
$8\le k\le12$, FAIL $k\ge13$; at $n{=}45$: PASS $k\le5$, INC $6$--$9$,
FAIL $k\ge10$); final test = whether the single pre-registered
$D_{\mathrm{test}}$ run was executed. Documents live under
\texttt{docs/reports/} or \texttt{configs/locked/}; commits are in the
project repository.}
\label{tab:provenance}
\begin{tabular}{@{}p{1.7cm}p{3.3cm}p{1.3cm}cp{1.1cm}p{1.2cm}p{1.8cm}p{2.0cm}p{2.2cm}@{}}
\toprule
Setting (pair, base) & Pre-registration document (date, commit) & $\lambda_{\max}$ & Fit/cert & $\lambda^\star_{\mathrm{ss}}$ & $\lambda^\star_{\mathrm{LTT}}$ & $D_{\mathrm{val}}$ rule / harm & Final test & Provenance label \\
\midrule
v1 (Rf+DiffLL, $G$-base) & \path{stage1_protocol.yaml} (2026-07-04, e4ddb9d); \path{stage3_gates.yaml} (2026-07-06, 661e692); ratified by \path{2026-07-12-certificate-v2-preregistration.md} (b9a8798) & 0.03 & 30/30 & 0.01974 & abstains ($\alpha{=}0.15$: 0.02359) & binary; 4/65 $=$ 6.2\% & consumed once, 2026-07-06 ($n{=}15$; harm 1/15) & exploratory-then-ratified \\
v1\_syn\_full (Rf+DiffRet, $R$-base) & \path{stage1_protocol_syn.yaml} + \path{2026-07-07-lolv2syn-direction-probe.md} (2026-07-07, 57f9c6c) & 0.03 & 30/30 & 0.01333 & 0.01717 & binary; 0/65 & consumed once, 2026-07-07 ($n{=}100$; harm 9/100) & pre-registered; pre-lock $D_{\mathrm{val}}$ oracle probe and full-$D_{\mathrm{cal}}$ direction oracle disclosed; 775-image retrain pre-declared \\
v1\_real\_full (Rf+DiffRet, $G$-base) & \path{stage1_protocol_lolv2real.yaml} + \path{2026-07-07-lolv2real-full-pipeline.md} (2026-07-07, 0ce530d) & 0.03 & 30/30 & 0.00563 (abstaining) & 0.00563 & binary; 0/65 & consumed once, 2026-07-07 ($n{=}100$; harm 0/100) & pre-registered before any $D_{\mathrm{val}}$ access \\
v1\_cid (CIDNet+DiffLL, $R$-base) & \path{2026-07-12-cidnet-expert-line-protocol.md} (2026-07-12, 0693efb) & 0.03 & 30/30 & 0.01333 & 0.01717 & binary; 11/65 $=$ 16.9\% FAIL & never consumed (E3, E4 failed) & pre-registered; published as a negative result \\
v1\_cgr (CIDNet+GSAD, $R$-base; degraded basin) & \path{2026-07-12-cidnet-gsad-real-line-protocol.md} (2026-07-12, 97bb770) & 0.03 & 30/30 & 0.00691 & 0.00691 & binary; 7/65 $=$ 10.8\% FAIL (one image) & never consumed (E3 failed) & pre-registered; pair chosen on $D_{\mathrm{test}}$ statistics (disclosed leak) \\
v1\_cgr\_det (det-path regeneration) & \path{2026-07-16-v1cgr-det-regeneration-preregistration.md} (2026-07-16, 2c3862c) & 0.03 & 30/30 & 0.00691 & 0.00691 & three-tier (binary reported side by side); 4/65 $=$ 6.2\% PASS & never consumed: E5 unlocked, aborted at the D3 basin gate (2026-07-17) & re-registered, integrity-driven \\
v1\_cgr\_off (CIDNet+GSAD official basin, $G$-base) & \path{2026-07-17-v1cgr-off-line-preregistration.md} (2026-07-17, f07ca37) & 0.03 & 30/30 & 0.01333 & 0.01846 & three-tier; 3/65 $=$ 4.6\% PASS & consumed once, 2026-07-17 ($n{=}100$; harm 0/100) & re-registered; direction flipped by the registered rule \\
sice (CIDNet+DiffLL, $R$-base) & \path{2026-07-13-sice-line-joint-preregistration.md} + A1 (2026-07-13, 1b21973); replay gate (2026-07-16, 7dfc181) & 0.03 $\to$ 0.045 & 28/76 & 0.04307 & 0.04307 & three-tier; 0/45 PASS & consumed once, 2026-07-16 ($n{=}40$; harm 0/40) & pre-registered before any metric; cap-correction clause exercised once \\
\bottomrule
\end{tabular}
\end{table*}

\section{Construction--validation boundary audit}
\label{supp:greyzones}

The calibration-split audit (2026-08-18) found no validation-driven choice
among certificate-derived policies, but records three places where the
boundary between construction and validation is softer than the main-text
discipline paragraph alone conveys.
(i) The move from a window-level to an image-level certificate on v1
(2026-07-05) was driven by $D_{\mathrm{val}}$ empirical-coverage tests
(the window-level constructions failed at 18.4\% and 15.5\% against the
10\% target), so v1's validation harm of 6.2\% is a post-selection number;
the v1 final test itself was single-shot.
(ii) The LOLv2-synthetic retrain from 300 to 775 training images was
triggered by a $D_{\mathrm{val}}$ result (the 300-image router's raw gain
was not significant), but the 775-image run was declared the setting's sole
main result unconditionally, in writing, before its numbers existed; both
versions are reported.
(iii) The E3 safety rule evolved from a binary threshold to a three-tier
rule after v1\_cgr failed the binary rule by one image (7/65 $=$ 10.77\%);
the three-tier rule was registered on SICE (amendment A1) before any SICE
metric and copied verbatim into the det and off re-registrations, the
binary verdict on v1\_cgr was never retroactively changed, and on every later setting the two rules agree (4/65, 3/65, 0/45), so the rule change
never decided a final-test consumption.

\section{Consumption ledger}
\label{supp:ledger}

Table~\ref{tab:ledger} reproduces the fifteen-row consumption ledger
(\path{outputs/logs/consumption_ledger.json}, gate G8) that backs the
discipline sentence of the main text: each pre-registered final-test
endpoint was executed exactly once, and the underlying image sets were read
a further, disclosed number of times for operations that decide nothing.
Per image set the ledger counts 1 endpoint + 2 non-endpoint reads (LOL-v1,
$n{=}15$), 1 + 1 (LOLv2-synthetic, $n{=}100$), 2 + 7 (LOLv2-real,
$n{=}100$; two distinct settings on byte-identical splits), and 1 + 0 (SICE
$-$1EV, $n{=}40$). The lower panel gives the role-membership hashes (md5
prefix of the id list) of every split manifest; the three
identical-$D_{\mathrm{test}}$ groups (1b2ac6d6: LOL-v1 main / CIDNet /
cross-fit aliases; c15ed57a: LOLv2-real full / CIDNet+GSAD alias;
dc8f3315: SICE $-$1EV / retired minimum-exposure manifest) are the formal
statement that ``two settings on the same test images'' means the same
ids, not a re-draw. The ledger reports no missing manifests and no
missing evidence files.

\begin{table*}[t]
\centering
\scriptsize
\setlength{\tabcolsep}{3pt}
\caption{Consumption ledger (all 15 rows of
\texttt{outputs/logs/consumption\_ledger.json}) and split-role membership
hashes. Endpoint = a pre-registered single-shot final test; non-endpoint =
a read of the same images that decides nothing. Dates are those of the
evidence file or of the dated report.}
\label{tab:ledger}
\begin{tabular}{@{}cp{1.8cm}lcp{1.5cm}p{5.0cm}p{2.2cm}p{4.0cm}@{}}
\toprule
\# & Image set & Role & $n$ & Kind & Read for & Setting / date & Evidence \\
\midrule
1 & LOL-v1 eval15 & $D_{\mathrm{test}}$ & 15 & endpoint & v1 E5 single-shot final test & v1 / 2026-07-06 & \path{outputs/logs/stage4/dtest_final.json} \\
2 & LOL-v1 eval15 & $D_{\mathrm{test}}$ & 15 & non-endpoint & same-pipeline SOTA inference (CIDNet / GSAD / ModalFormer~\cite{cidnet,gsad,modalformer}) & SOTA protocol / 2026-07-10 & \path{outputs/logs/stage5/sota_LOLv1.json} \\
3 & LOL-v1 eval15 & $D_{\mathrm{test}}$ & 15 & non-endpoint & qualitative panel rendering (fig.\ cases; per-image LPIPS read from frozen records) & fig.\ cases / 2026-08-18 & \path{outputs/dtest_composites} \\
4 & LOLv2-syn test & $D_{\mathrm{test}}$ & 100 & endpoint & v1\_syn\_full E5 single-shot final test & v1\_syn\_full / 2026-07-07 & \path{outputs/logs/stage4/dtest_final_diff_retinex_LOLv2syn.json} \\
5 & LOLv2-syn test & $D_{\mathrm{test}}$ & 100 & non-endpoint & same-pipeline SOTA inference & SOTA protocol / 2026-07-10 & \path{outputs/logs/stage5/sota_LOLv2syn.json} \\
6 & LOLv2-real test & $D_{\mathrm{test}}$ & 100 & endpoint & v1\_real\_full E5 single-shot final test & v1\_real\_full / 2026-07-07 & \path{outputs/logs/stage4/dtest_final_diff_retinex_LOLv2real.json} \\
7 & LOLv2-real test & $D_{\mathrm{test}}$ & 100 & endpoint & v1\_cgr\_off E5 single-shot final test (same images, second setting) & v1\_cgr\_off / 2026-07-17 & \path{outputs/logs/stage4/dtest_final_gsad_off_LOLv2real_cg.json} \\
8 & LOLv2-real test & $D_{\mathrm{test}}$ & 100 & non-endpoint & same-pipeline SOTA inference (CIDNet / GSAD / ModalFormer) & SOTA protocol / 2026-07-10 & \path{outputs/logs/stage5/sota_LOLv2real.json} \\
9 & LOLv2-real test & $D_{\mathrm{test}}$ & 100 & non-endpoint & modern-pair selection diagnostic (disclosed leak, downgraded to a post-hoc diagnostic) & v1\_cgr / 2026-07-12 & \path{docs/reports/2026-07-12-cidnet-gsad-real-line-protocol.md} \\
10 & LOLv2-real test & $D_{\mathrm{test}}$ & 100 & non-endpoint & D3 wrapper-vs-official basin comparison (aborted the v1\_cgr\_det final test) & v1\_cgr\_det / 2026-07-17 & \path{outputs/logs/stage_minus1/d3_wrapper_vs_official_LOLv2real_cg_Dtest.json} \\
11 & LOLv2-real test & $D_{\mathrm{test}}$ & 100 & non-endpoint & gsad\_det expert outputs on $D_{\mathrm{test}}$ (E5 aborted; final test never run) & v1\_cgr\_det / 2026-07-17 & \path{outputs/expert_outputs/LOLv2real_cg/gsad_det/seed1} \\
12 & LOLv2-real test & $D_{\mathrm{test}}$ & 100 & non-endpoint & D3$'$ basin gate, original verdict FAIL (per-image p95 $0.01438>$ registered 0.0135) & v1\_cgr\_off / 2026-07-17 & \path{outputs/logs/stage_minus1/d3prime_gsad_off_Dtest.json} \\
13 & LOLv2-real test & $D_{\mathrm{test}}$ & 100 & non-endpoint & D3$'$ yardstick recalibration (amendment A1, rule locked before running; official form, seed 4358) $\to$ p95$_{\mathrm{ref}}$ 0.01516, revised gate PASS ($0.01438\le1.25\times0.01516=0.01895$) & v1\_cgr\_off / 2026-07-17 & \path{outputs/logs/stage_minus1/d3prime_yardstick_recal.json} \\
14 & LOLv2-real test & $D_{\mathrm{test}}$ & 100 & non-endpoint & G5 deterministic negative control (reads existing per-image records) & submission gate / 2026-07-25 & \path{outputs/logs/stage_minus1/g5_deterministic_control_cidnet_LOLv2real.json} \\
15 & SICE $-$1EV holdout & $D_{\mathrm{test}}$ & 40 & endpoint & SICE E5 single-shot final test (after the bit-exact replay gate) & sice / 2026-07-16 & \path{outputs/logs/stage4/dtest_final_SICE_minus1ev.json} \\
\bottomrule
\end{tabular}

\vspace{4pt}
\begin{tabular}{@{}p{4.4cm}ccccc@{}}
\toprule
Split manifest & $D_{\mathrm{train}}$ & $D_{\mathrm{select}}$ & $D_{\mathrm{cal}}$ & $D_{\mathrm{val}}$ & $D_{\mathrm{test}}$ \\
\midrule
\path{lolv1_splits} / \path{lolv1_cidnet} / \path{lolv1_xfit} & 300 / e18d75ab & 60 / ae924ed2 & 60 / cfb63d18 & 65 / c22bc393 & 15 / 1b2ac6d6 \\
\path{lolv2real_full} / \path{lolv2real_cg} & 564 / ec7ef7d6 & 0 / d41d8cd9 & 60 / b89228b0 & 65 / 0ae1b08a & 100 / c15ed57a \\
\path{lolv2syn_full} & 775 / ebab7908 & 0 / d41d8cd9 & 60 / cf5d56ef & 65 / 76ee0a76 & 100 / d829cbea \\
\path{sice_minus1ev} / \path{sice} & 360 / 9a1e4527 & 40 / ed3b3aa7 & 104 / 15388f1f & 45 / bfd1c42c & 40 / dc8f3315 \\
\bottomrule
\multicolumn{6}{@{}p{16.8cm}@{}}{\scriptsize Cells are $n$ / md5 prefix of the role's id list (d41d8cd9 = empty list). Identical-$D_{\mathrm{test}}$ groups: 1b2ac6d6 $=$ \{lolv1\_splits, lolv1\_cidnet, lolv1\_xfit\}; c15ed57a $=$ \{lolv2real\_full, lolv2real\_cg\}; dc8f3315 $=$ \{sice\_minus1ev, sice\}.}
\end{tabular}
\end{table*}

%% file: sec/0_abstract.tex
\begin{abstract}
Generative low-light enhancers recover plausible texture but may hallucinate structure; restoration networks preserve structure but can leave regions under-enhanced, and neither advantage is spatially uniform. Therefore, we formulate selective generative enhancement as a decision problem: determine where overriding a frozen base expert is beneficial and refuse to switch when the evidence is insufficient. We propose VetoNet, an auditable regional routing framework centered on a certified veto. It first measures the complementary strengths and routing direction of a frozen restoration–generative pair, then predicts the utility of regional overrides from deployment-available signals alone and admits an override only when a distribution-free harm-budget certificate says the perceptual (LPIPS) risk budget allows it. Experiments across multiple expert pairs and domains show that VetoNet adapts its routing behavior to the strength and reliability of the available evidence. When expert complementarity is sufficiently predictable, it achieves statistically significant certified gains; when the relative strengths of the experts change, it correspondingly reverses the routing direction; and when routing evidence is weak or affected by cross-domain shift, it conservatively vetoes switching. On every pre-registered final test the certified mode stayed within its harm budget; per-image records and a consumption ledger will be released.
\end{abstract}

%% file: sec/1_intro.tex
\section{Introduction}
\label{sec:intro}

\begin{figure}[t]
    \centering
    \includegraphics[width=\linewidth]{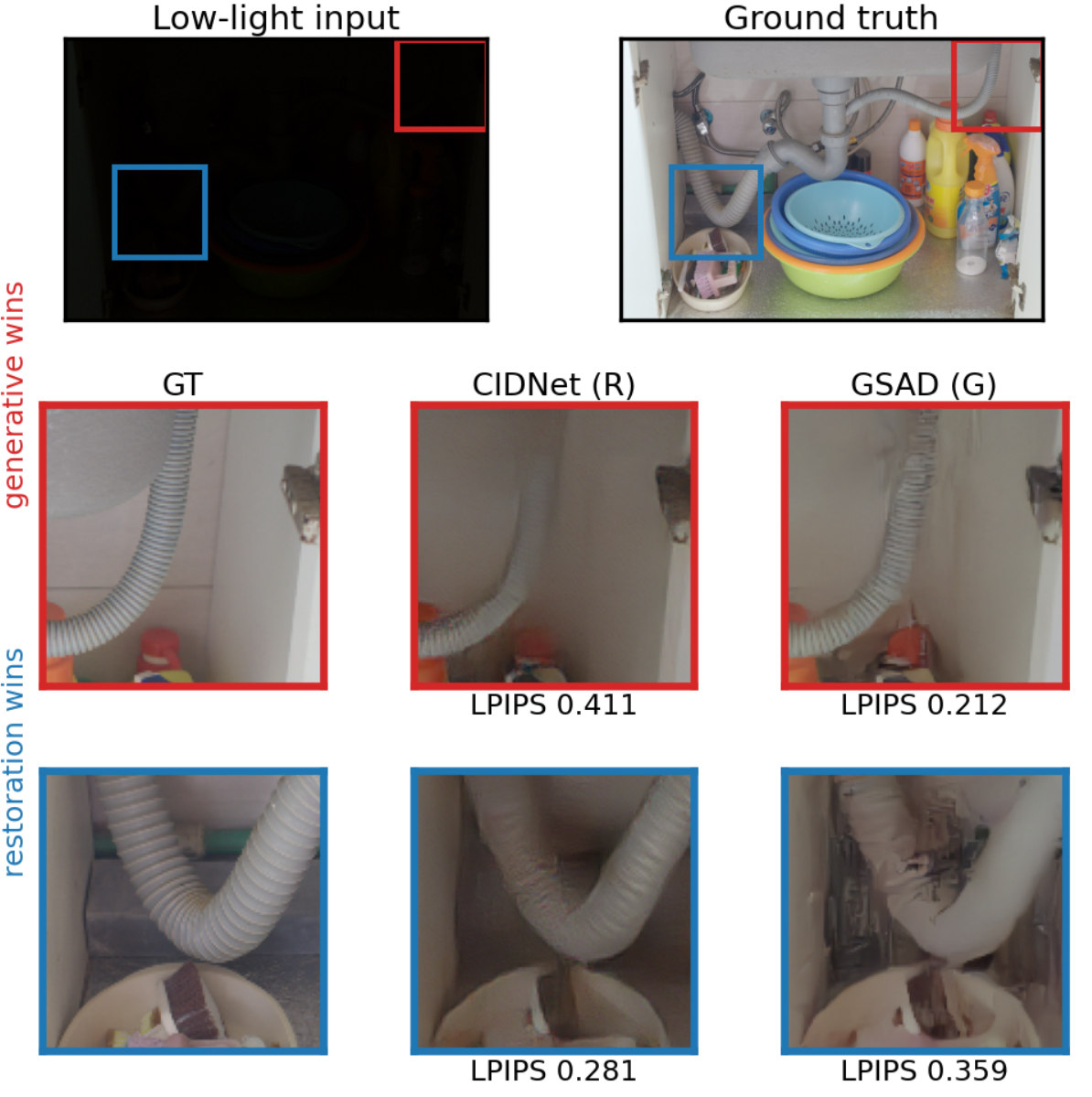}
    \vspace{-6mm}
    \caption{An example illustrating that different LLIE methods may exhibit complementary effects.}
    \label{fig:motivation}
    \vspace{-5mm}
\end{figure}

Low-light image enhancement (LLIE) has commonly been considered a restoration problem, aiming to recover the illumination conditions of a given degraded low-light image while preserving the underlying scene content and structure \cite{Chen2018Retinex, 9956107}. Recently, more generative methods have been proposed, as they can produce perceptually plausible textures and details that may be difficult to recover with restoration methods \cite{Yi2023DiffRetinexRL, Zhou2023PyramidDM}. However, most existing LLIE methods enhance the entire image with a
single model, implicitly assuming that the selected enhancement algorithm is equally suitable for different regions. They may overlook the possibility that restoration and generative models may have complementary effects within the same image. For example, as shown in Fig.~\ref{fig:motivation}, the generative model produces richer details in the red-boxed region, the folds on the pipe, while it introduces noticeable hallucinated structures in the blue-boxed region. Thus, the generative output is preferred for the red region, while the restoration output is selected for the blue region to better preserve reliable scene content. Therefore, the key question is not which
expert is overall better but whether overriding one frozen expert with the other (candidate) has positive gain and whether that decision can be made safely, in a defined perceptual sense, without ground truth at test time.

A simple idea is to selectively switch experts only where doing so is beneficial, but this raises three difficulties. First, the routing direction cannot be fixed in advance: depending on the dataset and expert pair, either the generative or restoration expert may serve as the stronger base. We observe that the preferred base/override direction can reverse when the dataset or expert pair changes. Second, it is hard to judge the benefit of an override directly. Simply comparing the outputs of two experts on a local patch may be misleading because it ignores color differences, boundary interactions, overlapping regions, and the perceptual effect of being placed back into the full image. Third, a router optimized only for average quality can make harmful local decisions. A selective pipeline therefore needs not only to predict where switching may help but also to determine when the evidence is insufficient to justify the switch.

To address the problems, we propose VetoNet, an auditable framework for selective generative enhancement built around a certified veto. Given two frozen experts, VetoNet first compares their outputs to decide which expert should be the fixed base and whether the other candidate can improve specific regions. Candidate regions are then compared with the output of the base model and evaluated after insertion. These comparisons provide supervision for training a lightweight router. At deployment, the router predicts which regions may benefit from an expert override without using the ground truth. Importantly, the area budget is an upper bound rather than a quota. If the candidate is not reliable enough, VetoNet will leave the base output unchanged.

In our implementation, we use two deployment modes with different objectives. Both modes admit an override when the calibrated router score exceeds a threshold $\lambda$; increasing $\lambda$ produces increasingly more conservative routing policies. The quality mode ($\lambda=0$) accepts windows with positive predicted utility, aiming to improve average perceptual quality without an explicit harm constraint. The certified mode instead calibrates $\lambda$ to control the probability of degrading an image relative to the fixed base expert. Because an image that is safe at one threshold need not remain safe at all more conservative thresholds, we assign each image the smallest threshold above which it remains unharmed for every more conservative policy. We call this quantity the safe-suffix score (Sec.~\ref{sec:framework}). Under exchangeability, split conformal prediction over these scores provides finite-sample marginal control of LPIPS-defined harm at level $\alpha$, and vetoes all overrides when no threshold can be certified. The guarantee is narrow by design: marginal rather than per-image, perceptual rather than semantic, and silent under arbitrary distribution shift.

This paper treats the question as an evaluation problem before a modeling problem, and its contributions are methodological:
\begin{itemize}\setlength{\itemsep}{1pt}\setlength{\parskip}{0pt}
\item A \textbf{counterfactual measurement protocol}: overlapping regions are counterfactually swapped between two frozen experts and scored against paired ground truth, giving a per-region net-utility label that a 4.1M U-Net router learns from deployment-available inputs alone (Sec.~\ref{sec:framework});
\item \textbf{VetoNet, a certified veto}: an image-level, distribution-free risk-control rule for LPIPS-defined non-degradation rule that admits a routed override only when a conformally calibrated threshold says the risk budget allows it, and otherwise vetoes generation entirely, returning the fixed base expert (Sec.~\ref{sec:framework});
\item An \textbf{audit trail as a first-class result}: four selection-independent pre-registered single-shot final tests, together with a disclosed modern-pair audit; an adversarial audit that retired our own earlier calibration rule and repaired a real 16.9\% coverage failure, with rule baselines showing why naive calibration fails there (Sec.~\ref{sec:eii}); stress tests mapping where the guarantee breaks and a label-free guard for the detectable part; a process-state reproducibility hazard in a public diffusion enhancer, caught by a deferred audit and repaired in the official process form; and a leak-free cross-domain terminal test in which calibrated abstention generalizes exactly (Sec.~\ref{sec:eiii}).
\end{itemize}

%% file: sec/2_relatedworks.tex
\section{Related Work}
\label{sec:relatedworks}

Low-light image enhancement has been studied through Retinex decomposition \cite{Chen2018Retinex,kind,uretinex}, end-to-end regression \cite{mirnet,snr,llflow}, transformer-based architectures \cite{retinexformer,restormer}, color-space formulations \cite{cidnet}, unpaired and zero-reference learning \cite{enlightengan,zerodce}, and raw-image pipelines \cite{sid}. More recently, diffusion-based methods such as DiffLL \cite{diffll}, Diff-Retinex \cite{Yi2023DiffRetinexRL}, and GSAD \cite{gsad}, building on the DDPM family \cite{ddpm,sr3}, have reported strong perceptual performance. We take five public checkpoints as frozen experts.

\subsection{Perception distortion and generative fidelity}
Perceptual quality and reconstruction fidelity are different operating objectives \cite{perception}, and generative models can produce visually convincing outputs without preserving the underlying structure, as illustrated by PULSE \cite{pulse}. Our evaluation therefore requires a full-reference metric when determining whether an intervention improves or degrades a paired sample. We use LPIPS \cite{lpips} as the primary perceptual measure and consider DISTS \cite{dists} as an alternative. In an additional no-reference evaluation using \cite{niqe,brisque,musiq,maniqa,clipiqa,nima}, generative outputs can be ranked above ground truth on two of the three examined datasets. Thus, we do not use these no-reference metrics to define fidelity-sensitive routing decisions in the present protocol.

\subsection{Regional routing and mixtures}
Mixture-of-experts methods \cite{jacobs1991,shazeer2017} and spatial routing approaches for restoration \cite{classsr,pathrestore} typically decide where computation or model capacity should be allocated. Our setting instead treats routing as a decision about which expert output should be trusted. The router predicts counterfactual regional utility measured against paired ground truth during development, while the deployed decision is additionally constrained by a statistical certificate rather than by an unconstrained heuristic threshold.

\subsection{Conformal prediction and risk control}
Split conformal prediction provides finite-sample marginal guarantees under exchangeability \cite{vovk,lei2018,gentle}, while risk-controlling prediction sets \cite{rcps}, Learn-then-Test \cite{ltt}, and conformal risk control \cite{crc} extend related ideas to broader risk functions. Distribution-free uncertainty has also been studied for image-to-image regression \cite{im2imuq}. We adapt this machinery to a selective deployment decision between override and veto. A safe-suffix split-conformal rule is used as the deployed certificate, with a valid Learn-then-Test procedure serving as an auxiliary $(\alpha,\delta)$ check. As in selective classification \cite{chow1970,elyaniv2010,geifman2017}, abstention is a valid output: when no non-trivial intervention can be certified, the system retains the fixed expert.

Our contribution is that we study how finite-sample risk-control machinery behaves when applied to regional generative enhancement, including cases in which simpler constructions fail, and how the resulting evaluation procedure can be audited under reproducibility and distribution-shift stress tests. The overall protocol follows reproducibility practices motivated by prior work \cite{lucic2018,bouthillier2019,pineau2021}, including pre-registration, separated data roles, single-shot final tests, disclosed deviations, and explicit evaluation conventions.

%% file: sec/3_framework.tex
\section{Measurement Framework and Certificates}
\label{sec:framework}

\subsection{Problem Formulation}
\begin{figure}[ht]
    \centering
    \includegraphics[width=\linewidth]{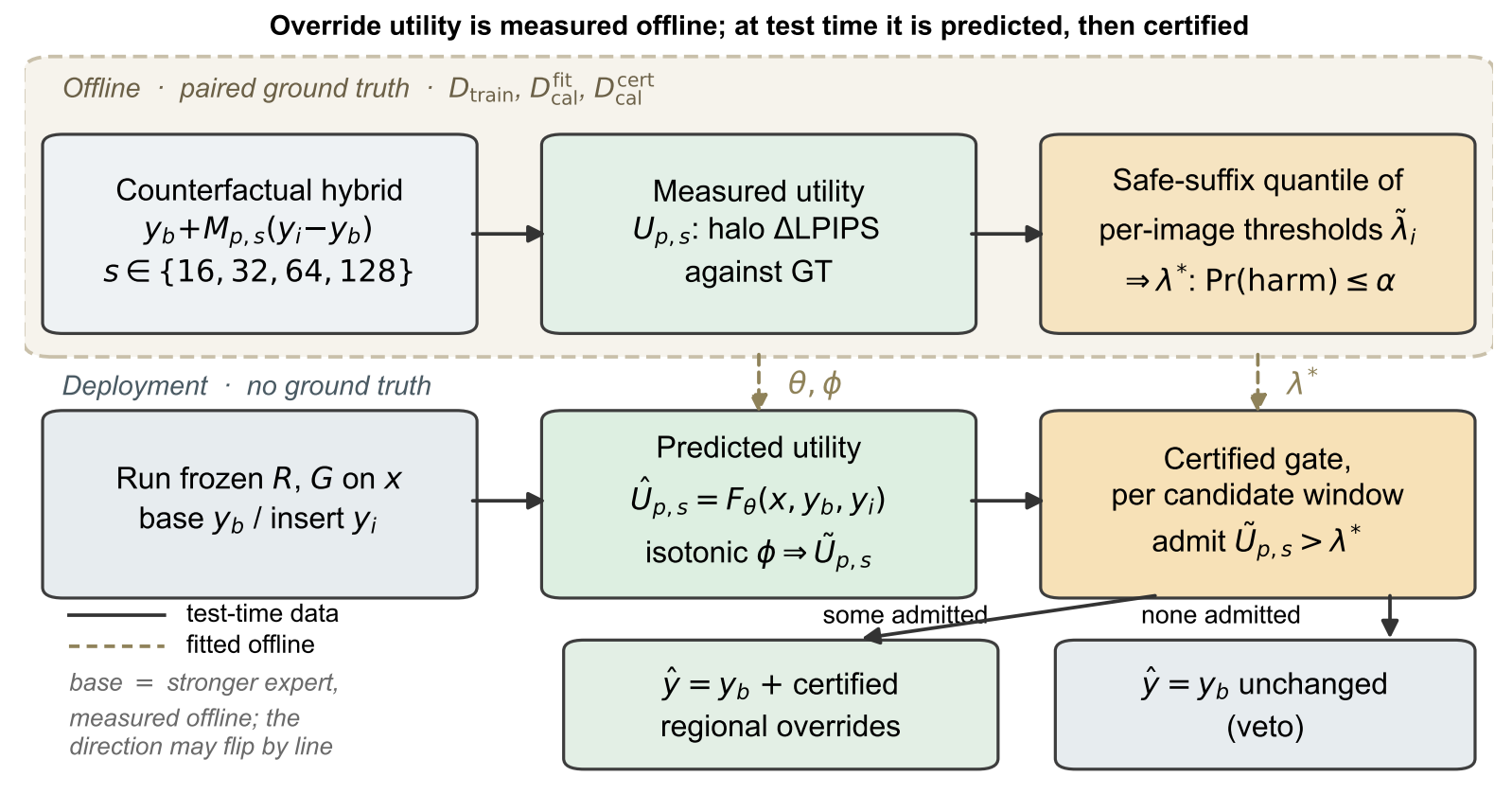}
    \vspace{-5mm}
    \caption{Overview of the VetoNet pipeline.}
    \label{fig:overview}
    \vspace{-5mm}
\end{figure}

A frozen restoration expert $R$ and a frozen generative expert $G$ process the same low-light input $x$ (Fig.~\ref{fig:overview}). For each dataset line, a pre-registered rule selects one expert as the base, while the other provides candidate regional overrides. For overlapping windows $(p,s)$ with side $s\in\{16,32,64,128\}$ and stride $s/2$, a raised-cosine mask $M_{p,s}$ defines the local counterfactual
\begin{equation}
y^{\mathrm{hyb}}_{p,s} \;=\; y_{\mathrm{base}} + M_{p,s}\odot
\bigl(y_{\mathrm{ins}}-y_{\mathrm{base}}\bigr),
\end{equation}
after deployment-consistent color alignment, and the utility $U_{p,s}$ is defined as the halo-restricted LPIPS change against paired ground truth and normalized separately at each scale. For stochastic experts, we average three seeds. Candidate regions are pre-screened using an integral-image approximation, which achieves Spearman correlations of 0.57–0.94 with exact counterfactual scores. A greedy oracle then composes overrides under a 20\% area budget to recompute marginal utilities as regions are added and to estimate the routing upper bound.

\subsection{Direction by Measurement}
We evaluate both possible routing directions, Restoration (R) $\rightarrow$ Generative (G) and $G \rightarrow R$, where the first model is the base and the second provides regional overrides. We applied a dual-direction oracle calibration on the first half of the calibration split $D_{\mathrm{cal}}$ to  select the direction whose oracle result outperforms both experts on more images.

\subsection{Router}
A U-Net (4.09M parameters) $F_\theta$ takes the low-light image and the two expert outputs as input,
\[
F_\theta(x,y_{\mathrm{base}},y_{\mathrm{ins}})
=
\bigl(\{\hat u_s\}_{s\in\mathcal S},\hat D\bigr),
\quad
\mathcal S=\{16,32,64,128\}
\]

where $\hat u_s$ is the predicted utility at scale $s$ and $\hat D$ is an auxiliary dense map. Each scale uses a $1{\times}1$ prediction head and is resized to the corresponding window grid.

Let $u_{p,s}=\frac{U_{p,s}}{r_{H,s}}$ be the normalized utility. The router is trained with
\begin{equation}
\label{eq:router-loss}
\begin{split}
&\mathcal L_{\mathrm{router}}
=\;
\frac{1}{|\mathcal S|}
\sum_{s\in\mathcal S}
\mathbb E_p
\Big[
(|u|+0.1)\,
\ell_1^{\mathrm{sm}}(\hat u,u)
\\[-5pt]
&
+0.5\,|u|\,
\bigl[0.5-\operatorname{sign}(u)\hat u\bigr]_+
\Big]
+0.3\,\|\hat D-D\|_1 ,
\end{split}
\end{equation}
where $\ell_1^{\mathrm{sm}}$ is the smooth $L_1$ loss and $D$ is the normalized spatial LPIPS-difference map. The $|u|$ weighting gives more importance to regions with larger utility. Before routing, the predicted utility is mapped back to the original scale and calibrated separately for each scale:
\[
\hat U_{p,s}=r_{H,s}\hat u_{p,s},
\qquad
\widetilde U_{p,s}=g_s(\hat U_{p,s}),
\]
where $g_s$ is an isotonic regressor fitted on $D_{\mathrm{cal}}^{\mathrm{fit}}$. The dense head is used only during training.

\subsection{The Certified Veto (VetoNet)}
We control deployment risk at the image level. An image is considered harmed if the routed output has higher LPIPS than the fixed base expert, with a strict tolerance of $\varepsilon=0$ (see Sec.~\ref{sec:discussion} for limitations). Using the second half of $D_{\mathrm{cal}}$, each calibration image $i$ defines the smallest threshold $\tilde{\lambda}_i$ such that all more conservative thresholds remain unharmed; if no such threshold exists, $\tilde{\lambda}_i=+\infty$. The deployed threshold is
\begin{equation}
\lambda^\star
=
\tilde{\lambda}_{(\lceil(1-\alpha)(n+1)\rceil)} .
\end{equation}

Under image-level exchangeability, this gives the marginal guarantee $\Pr(\mathrm{harm})\leq\alpha$ without assuming monotonic harm in $\lambda$ ~\cite{vovk,lei2018}. This safe-suffix construction is important because per-image harm can be non-monotone in $\lambda$, and simpler rules that ignore such recurrences may over-certify unsafe thresholds (Sec.~\ref{sec:eii}; per-line counts in the supplementary).

We also report an auxiliary Learn-then-Test (LTT) procedure~\cite{ltt}, using exact binomial $p$-values, a fixed sequence from the most conservative threshold, and $\delta=0.05$. It either certifies the same or a more conservative threshold or abstains (Sec.~\ref{sec:eii}). At deployment, only windows whose calibrated score exceeds $\lambda^\star$ are accepted. If no window is admitted, the system returns the base expert unchanged.

\subsection{What the audit changed}
Two earlier certificate designs were retired during the audit. A window-level certificate failed to control image-level harm, motivating the image-level formulation above. A later audit also found our previous $(V{+}1)/(N{+}1)$ threshold-selection rule invalid as a risk-control procedure, so it was replaced by the safe-suffix rule. Details of the retired certificate designs are given in the supplementary material, while their audit outcomes are reported in Sec.~\ref{sec:eii}.

\subsection{Evaluation Protocol}
Each dataset was split once at the scene level (seed 2026) into mutually exclusive training, expert-selection, calibration, validation, and final-test roles. The calibration split was further partitioned into disjoint fit and certification subsets: direction selection, score calibration, and noise-floor estimation used only the former, while conformal certification used only the latter. Protocols, threshold grids, budgets, and endpoints were fixed before validation access on each line, except for two disclosed protocol deviations on LOLv2-synthetic. A pre-registered grid-cap correction clause was exercised once on SICE. Each pre-registered final-test endpoint was executed once. The LOLv2-synthetic deviations are detailed in the supplementary: the router never saw $D_{\mathrm{val}}$ before the lock, and a fit-half-only recomputation returned the same routing direction. Every non-endpoint read of the underlying image sets is itemized with role-membership hashes in the supplementary consumption ledger.

%% file: sec/4_evidence1.tex
\section{Result I: Routing Learns Complementarity}
\label{sec:ei}

\begin{table*}[t]\centering
\caption{Final-test outcomes on the LOL family. The first three rows are selection-independent pre-registered endpoints; the modern-pair row is reported as a disclosed audit outcome because the official test set informed expert-pair selection. Gains are mean per-image LPIPS relative to the fixed base expert (paired-bootstrap 95\% CIs). Rf: Retinexformer; DiffRet: Diff-Retinex. SICE terminal test: Sec.~\ref{sec:eiii}.Sec.~\ref{sec:eiii}.}
\label{tab:cross}
\scriptsize
\setlength{\tabcolsep}{4.5pt}
\begin{tabular}{llccc}
\toprule
Dataset (test $n$) & Pair (base first) & Quality-mode gain & Certified gain & Certified harm\\
\midrule
LOL-v1 (15) & DiffLL\,($G$)\,$|$\,Rf & $+0.0065\;[+0.0030,+0.0102]$ & $+0.0028\;[+0.0007,+0.0054]$ & $6.7\%$\\
LOLv2-syn (100) & Rf\,($R$)\,$|$\,DiffRet & $-0.0002\;[-0.0018,+0.0014]$ &
$\mathbf{+0.0014\;[+0.0005,+0.0025]}$ & $9.0\%$\\
LOLv2-real (100) & DiffRet\,($G$)\,$|$\,Rf & $+0.0006\;[-0.0026,+0.0039]$ &
$0$ (abstains) & $\mathbf{0.0\%}$\\
LOLv2-real, modern pair modern-pair audit$^\dagger$ (100) & GSAD\,($G$)\,$|$\,CIDNet &
$-0.0043\;[-0.0058,-0.0027]$ &
$\mathbf{+0.00023\;[+0.00007,+0.00044]}$ & $\mathbf{0.0\%}$\\
\bottomrule
\end{tabular}
\end{table*}

\begin{table}[t]\centering
\caption{Evidence-I premises (LOL-v1 validation, $n=65$, 20\% area
budget, paired-bootstrap 95\% CIs; last row from LOLv2-real validation).
Cue-by-cue tables are in the supplementary.}
\label{tab:premises}
\scriptsize
\setlength{\tabcolsep}{2pt}
\begin{tabular}{@{}ll@{}}
\toprule
Premise & Value\\
\midrule
Oracle headroom vs.\ fixed expert & $+0.0106$ $[+0.0092,+0.0121]$\\
Area at which the oracle already wins & 2.5\% (saturates near 20\%)\\
Router margin over random / darkness & $+0.0109$ / $+0.0114$ (CI lower bound $>0$)\\
\quad over residual / uncertainty & $+0.0053$ / $+0.0101$ (CI lower bound $>0$)\\
Gain capture / mean regret & 57.2\% / 0.0046\\
Darkness cue vs.\ no routing & LPIPS 0.1164 vs.\ 0.1110 (worse)\\
LOLv2-real: beats cues / raw beats fixed & 75--88\% / 11\% of images\\
\bottomrule
\end{tabular}
\end{table}

\begin{figure}[t]
  \centering
  \includegraphics[width=0.68\columnwidth]{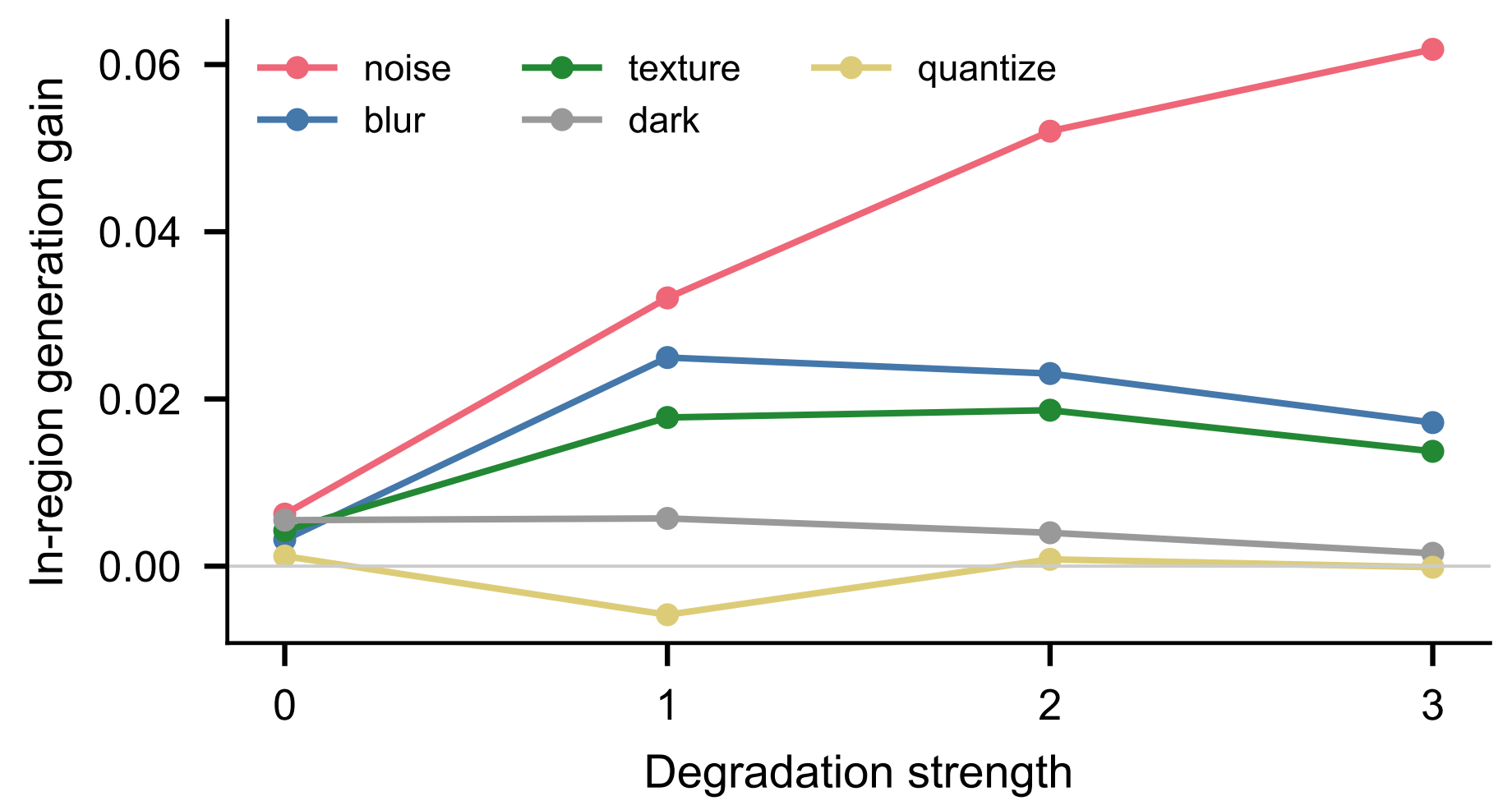}
  \vspace{-2mm}
  \caption{Counterfactual gain under syn degradations of LOL-v1 inputs.}
  \vspace{-5mm}
  \label{fig:dose}
\end{figure}

\paragraph{Premises.}
The following analysis is based on three observations (Table~\ref{tab:premises}). First, oracle headroom exists on LOL-v1 but is pair-dependent: the budget curve is nearly flat on LOLv2-syn and negative on LOLv2-real. Second, the learned router outperforms trivial cues at equal budget and captures 57\% of the available headroom, while darkness-based routing can be worse than no routing. On LOLv2-real, the headroom is too small to support deployment. Third, the routing signal tracks recoverable information loss; among five local synthetic degradations applied at three strengths, only some increase the in-region gain (Fig.~\ref{fig:dose}; protocol in the supplementary).

\paragraph{Final tests.}
Table~1 reports the three pre-registered LOL-family final tests. On LOL-v1, quality-mode routing improves LPIPS from 0.1183 to 0.1118 ($+0.0065$, 95\% CI $[+0.0030,+0.0102]$), with 80\% of images improved. The certified mode is more conservative, reaching 0.1156 with 6.7\% empirical harm. On LOLv2-syn, raw routing does not retain its validation gain at test ($-0.0002$), while the certified mode provides the only significant deployable improvement ($+0.0014$, 95\% CI $[+0.0005,+0.0025]$) while keeping harm at 9\%. On LOLv2-real, raw routing already showed significant harm on validation ($-0.0051$, 95\% CI $[-0.0069,-0.0035]$). The certificate therefore does not allow overrides, making the test output the same as the fixed expert with 0\% harm. The certified mode can remain consistent across datasets, seeds, and splits.

\paragraph{A modern pair under a disclosed selection leak.}
We additionally studied the HVI-CIDNet~\cite{cidnet} and GSAD~\cite{gsad} pair on LOLv2-real as an audit case. This pair was initially chosen after diagnostics on the official test set showed different metric preferences: GSAD achieved lower LPIPS on 65\% of the images, whereas CIDNet achieved higher PSNR on 75\%. Because the test set influenced the choice of expert pair, this experiment contains a disclosed selection leak, and we do not use the initial test-set diagnostic as evidence of expert complementarity.

We therefore repeated the same comparison on the disjoint calibration $\cup$ validation dataset (125 images), with zero test overlap. In this leak-free analysis, correctly generated GSAD outperforms CIDNet on both LPIPS ($0.107$ vs.\ $0.143$) and PSNR ($23.8$ vs.\ $18.2$), showing that the apparent complementarity was specific to the data used for the initial diagnostic.

A second issue affected the first two of three runs: GSAD outputs had been generated from a degraded process basin (Sec.~\ref{sec:eiii}). With these incorrect outputs, the rule selected CIDNet as the base, and the pre-registered binary safety endpoint exceeded its target by one image (7/65 $=10.8\%$); deterministic regeneration reduced this to 4/65. After GSAD was regenerated using the verified official procedure, the relative expert strengths reversed, and the rule correspondingly selected GSAD as the base with CIDNet providing candidate overrides. Under this corrected setting, both pre-registered safety endpoints pass, with validation harm of 3/65 ($4.6\%$). The certified mode improves over both experts by $+0.00298$ LPIPS (95\% CI $[+0.00033,+0.00670]$), captures 38.5\% of the available oracle headroom, and leaves 54/65 validation images unchanged.

For completeness, on the official test split ($n=100$), the corrected certified policy produces zero empirical harm and a mean gain of $+0.00023$ LPIPS (95\% CI $[+0.00007,+0.00044]$), whereas raw routing is harmful ($-0.0043$, 95\% CI $[-0.0058,-0.0027]$) and improves only 20\% of images. Because this test set influenced the original pair selection, we treat these numbers as audit outcomes rather than selection-independent confirmatory evidence. Despite the weak router correlation (Spearman $\rho=0.116$), the certificate restricts intervention to a small subset of candidate overrides.

%% file: sec/5_evidence2.tex
\section{Result II: The Certificate Holds and Repairs}
\label{sec:eii}

\begin{table*}[t]\centering
\caption{Certificate comparison on the LOL-family validation sets ($n=65$, $\alpha=0.10$, $\varepsilon=0$). We compare the deployed safe-suffix (SS) rule, valid LTT ($\delta=0.05$), and the 'first-safe' conformal baseline. Bold highlights the LTT repair of v1\_cid and the corrected official-state modern pair; degraded-basin runs are retained for audit completeness. v1 cid = CIDNet+DiffLL; v1 cgr = CIDNet+GSAD.}
\vspace{-3mm}
\label{tab:cert}
\scriptsize
\setlength{\tabcolsep}{3.5pt}
\begin{tabular}{lcccccc}
\toprule
Setting & SS $\lambda^\star$ & SS harm (CP95) & LTT $\lambda^\star$ & LTT harm & first-safe $\lambda^\star$ & harm\\
\midrule
v1 (Rf+DiffLL) & 0.0197 & 4/65 $=$ 6.2\% [1.7,15.0] & abstains ($\alpha{=}.15$: 0.0236) & --- & 0.0197 & 4/65\\
v1\_syn (Rf+DiffRet) & 0.0133 & 0/65 & 0.0172 & 0/65 & 0.0133 & 0/65\\
v1\_real (Rf+DiffRet) & 0.0056 & 0/65 (abstains) & 0.0056 & 0/65 & 0.0056 & 0/65\\
v1\_cid (CIDNet+DiffLL) & 0.0133 & 11/65 $=$ 16.9\% & $\mathbf{0.0172}$ & $\mathbf{3/65=4.6\%}$ & 0.0108 & 13/65 $=$ 20.0\%\\
v1\_cgr (CIDNet+GSAD, degraded basin) & 0.0069 & 7/65 $=$ 10.8\% [4.4,20.9] & 0.0069 & 7/65 & 0.0069 & 7/65\\
v1\_cgr det-path regeneration & 0.0069 & 4/65 $=$ 6.2\% & 0.0069 & 4/65 & 0.0069 & 4/65\\
\textbf{v1\_cgr official state ($G$ base)} & 0.0133 & $\mathbf{3/65=4.6\%}$ & 0.0185 & 2/65 $=$ 3.1\% & 0.0120 & 3/65\\
\bottomrule
\end{tabular}
\end{table*}

\begin{figure}[t]
  \centering
  \includegraphics[width=0.66\columnwidth]{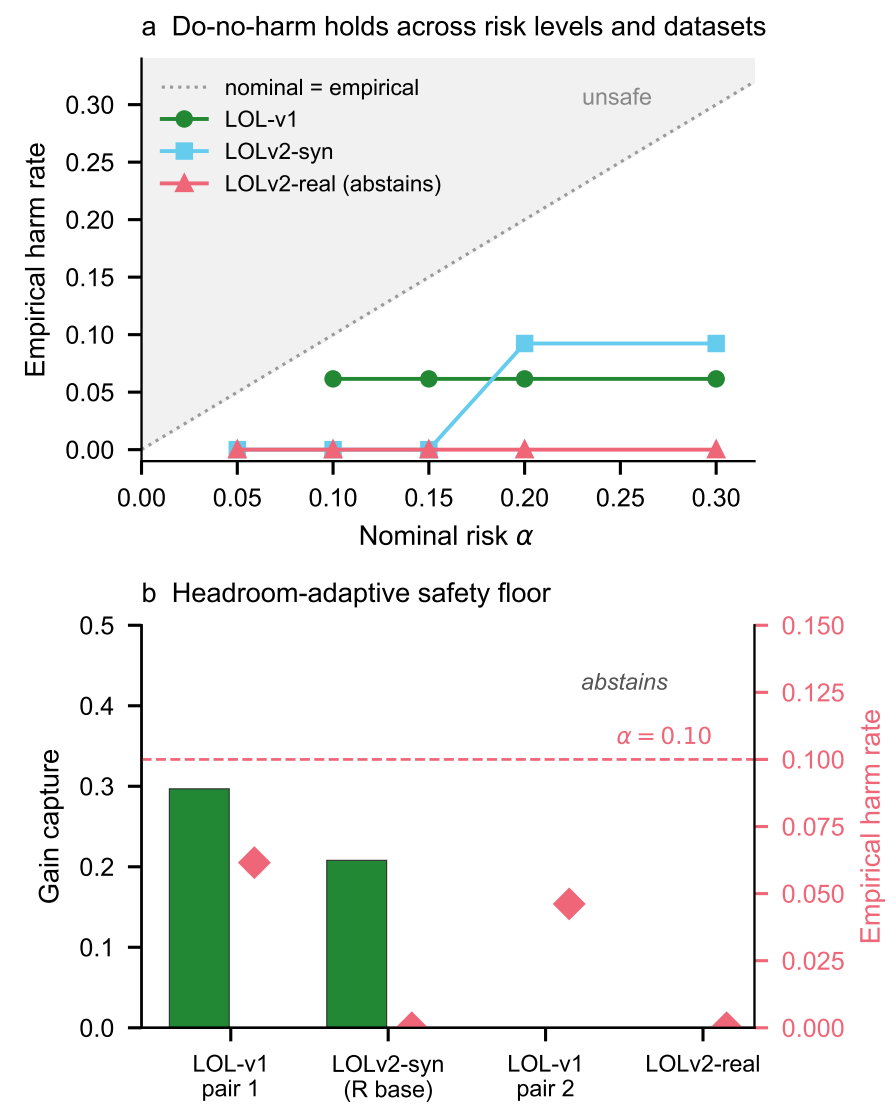}
  \vspace{-3mm}
  \caption{Certificate behavior across risk budgets. (a) Empirical harm versus nominal risk $\alpha$ for the three earliest settings using frozen calibration scores. (b) Certified capture and empirical harm at $\alpha=0.10$ for the four early LOL-family settings; v1\_cid is shown at the valid-LTT repaired point, and its negative capture ($-0.04$) is clipped. Full sweeps are provided in the supplementary; modern-pair results are reported in Table~\ref{tab:cert}.}
  \label{fig:cert}
\end{figure}

\paragraph{Overall certificate behavior.}
The results of safe-suffix and valid-LTT certificates on the LOL-family settings at $\alpha=0.10$ and $\varepsilon=0$ are tabulated in Table~\ref{tab:cert}. All four selection-independent final tests satisfy the nominal harm target. The disclosed modern-pair audit also yields zero empirical test harm but is treated separately because its test set informed expert-pair selection. On validation, two settings exceed 10\% harm: v1 cid reaches 16.9\% and is reduced to 4.6\% by valid LTT, while the degraded-basin modern pair reaches 10.8\% (7/65) and falls to 6.2\% (4/65) after deterministic regeneration. Fig.~\ref{fig:cert} illustrates that sweeping $\alpha$ from 0.05 to 0.30 with frozen calibration scores and no refitting shows the same general behavior: capture increases only when usable headroom exists. It rises from 10\% to 25\% on LOLv2-synthetic and from 9.5\% to 45.5\% on the corrected modern pair, while remaining flat on
LOL-v1 because tied scores keep $\lambda^\star=0.0197$ over $\alpha\in[0.10,0.30]$. Detailed sweep results are reported in the supplementary.

\paragraph{The audit repairs a real failure.}
On v1\_cid, the safe-suffix operating point produces 11/65 $=16.9\%$ validation harm against the 10\% target. Valid LTT rejects this point and moves to a more conservative threshold, reducing harm to 3/65 $=4.6\%$ without reducing the measured quality gain. On the remaining settings, LTT either certifies the recorded point or selects a more conservative one, never with higher validation harm; on LOL-v1 it abstains entirely at $\delta=0.05$. The audit therefore moves operating points only toward safety and repairs the one point that was not safe. It showed that the audit can change, rather than just
confirm, the deployment decision.

\paragraph{Why simpler calibration is insufficient.}
We then use two simpler rules to re-evaluate the same frozen calibration scores, which are a plug-in threshold based directly on empirical calibration harm and a 'first-safe' conformal rule that ignores later harm recurrences. When harm is approximately monotone, these rules select the same or nearby operating points. The difference appears when harm recurs as the threshold changes. On v1\_cid, the plug-in and first-safe rules produce 30.8\% and 20.0\% validation harm, respectively, compared with 16.9\% for safe-suffix and 4.6\% after the valid-LTT repair (Table~\ref{tab:cert}). On SICE, both simpler rules also release overrides
that harm 11--13\% of validation images, whereas safe-suffix and LTT abstain (0/45). These cases show why calibration must account for non-monotone harm rather than selecting the first apparently safe threshold.

\paragraph{Abstention when headroom vanishes.}
On LOLv2-real, the certificate admits no overrides at $\alpha=0.10$ and continues to abstain throughout the sweep up to $\alpha=0.30$. This is because every threshold that allows routing is harmful on validation (61/65 images; mean gain $-0.024$), leaving no useful operating point. SICE also abstains at $\alpha=0.10$, but it begins to allow overrides at larger risk budgets. The reason for zero routing at $\alpha=0.10$ is simple: the calibrated routing score has a fixed upper bound, and this maximum score is still below the certified threshold $\lambda^\star$ on both settings. Therefore, no window can be selected. These results show that abstention is the intended behavior when the available routing gain is too small to justify the risk.

\paragraph{Noise floors and $\varepsilon$-sensitivity.}
Finally, stochastic experts produce small run-to-run LPIPS variations. The p95 $|\Delta\mathrm{LPIPS}|$ noise floors are $4{\times}10^{-4}$ for DiffLL, $6.7{\times}10^{-3}$ for GSAD, and $3.0$--$3.7{\times}10^{-2}$ for Diff-Retinex. Therefore, $\varepsilon=0$ is a strict choice, because some small measured harms may come from the experts' own randomness rather than routing itself. The registered $\varepsilon$ sweep (supplementary) shows that allowing a small tolerance can account for this variability, but it does not remove the empirical validation target miss on v1 cid.

%% file: sec/6_evidence3.tex
\section{Result III: Boundaries and Reproducibility}
\label{sec:eiii}

\begin{table}[t]\centering
\caption{Stress test on LOL-v1 validation under full-image distribution shifts with zero recalibration. The full per-condition grid is provided in the supplementary.}
\vspace{-2mm}
\label{tab:stress}
\scriptsize
\setlength{\tabcolsep}{2.2pt}
\begin{tabular}{@{}lccccc@{}}
\toprule
Condition & Cert.\ harm & Area & Qual.\ harm & Guard det. & After guard\\
\midrule
clean & 6.2\% & 0.043 & 18.5\% & 6.2\% & 4.6\%\\
noise s1--s3 & 4.6--10.8\% & 0.004--0.006 & 86--99\% & 51--100\% & 0--1.5\%\\
blur s1--s3 & 10.8--56.9\% & 0.012--0.050 & 25--92\% & 100\% & \textbf{0\%}\\
dark s2--s3 & 15.4--32.3\% & 0.031--0.092 & 29--39\% & 86--100\% & 0--4.6\%\\
dark s1 (residual) & 27.7\% & 0.087 & 35.4\% & 32\% & 20.0\%\\
\bottomrule
\end{tabular}
\end{table}

\begin{figure}[t]
  \centering
  \includegraphics[width=0.64\columnwidth]{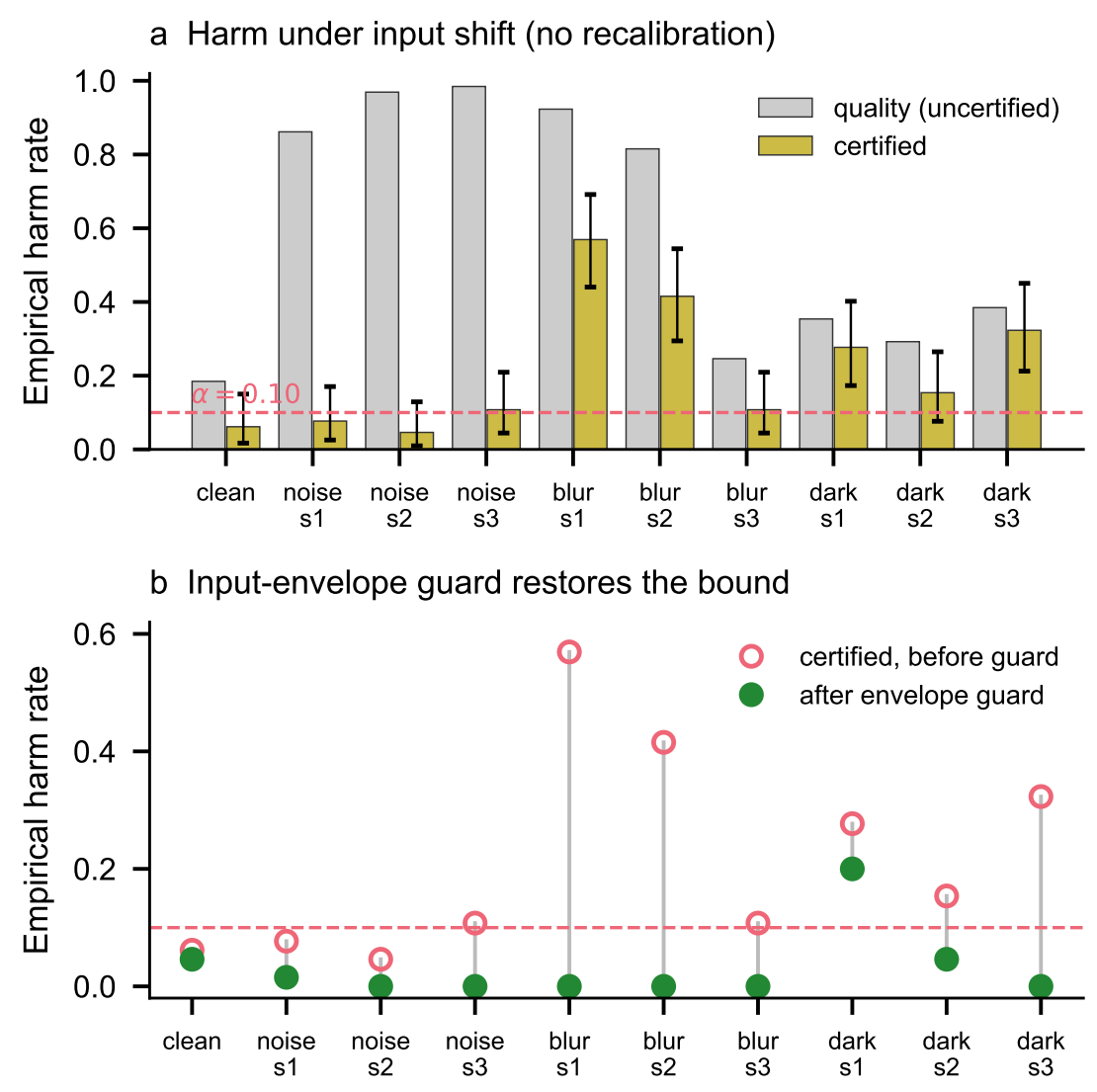}
  \vspace{-3mm}
  \caption{Stress test under input shift on LOL-v1 validation with zero recalibration. (a) Harm of the certified and quality modes with CP95 intervals. (b) Harm after applying the input-envelope guard.}
  \label{fig:stress}
\end{figure}

\paragraph{Where the guarantee breaks.}
Conformal validity relies on exchangeability. We therefore stress-tested the frozen LOL-v1 certificate under a pre-registered grid of full-image noise, blur, and darkening shifts at three strengths, without recalibration (Table~\ref{tab:stress}, Fig.~\ref{fig:stress}a). As the results show, under-score-collapsing shifts such as noise, calibrated scores decrease, and most candidate overrides fall below $\lambda^\star$. The routed area therefore collapses, and harm remains close to the nominal level. In this situation, the routing rule becomes empirically self-protective by abstaining. In contrast, score-preserving shifts such as blur and darkening can leave the router confident while changing which expert is actually better. The certificate then fails, with harm reaching 56.9\% under blur s1. Thus, abstention protects against shifts that are visible in the router scores but not against shifts that preserve those scores while changing the underlying utility. The uncertified quality mode is less stable under every tested shift, with harm ranging from 25\% to 99\%.

\paragraph{A label-free guard for detectable shifts.}
Therefore, we applied a simple input-envelope guard using three statistics, which are log-Laplacian energy, mean luminance, and their ratio. The accepted range of each statistic is defined only from the 60 clean calibration images, requiring neither labels nor retraining. Inputs outside this envelope are forced to abstain. The guard detects all tested blur shifts and 86--100\% of the stronger darkening shifts, reducing post-guard harm to 0\% for blur and 0--4.6\% for dark s2-s3 (Fig.~\ref{fig:stress}b). It also reduces noise harm to 0--1.5\%. On clean data, the guard has a 6.2\% false-alarm rate while retaining 90.5\% of the original certified gain. The main residual failure is mild darkening (dark s1): only 32\% of these inputs fall outside the clean envelope, leaving 20.0\% post-guard harm. This is a limitation of input-only detection when the shifted input distribution substantially overlaps the clean one.

\begin{figure*}[t]
\centering
\includegraphics[width=\textwidth]{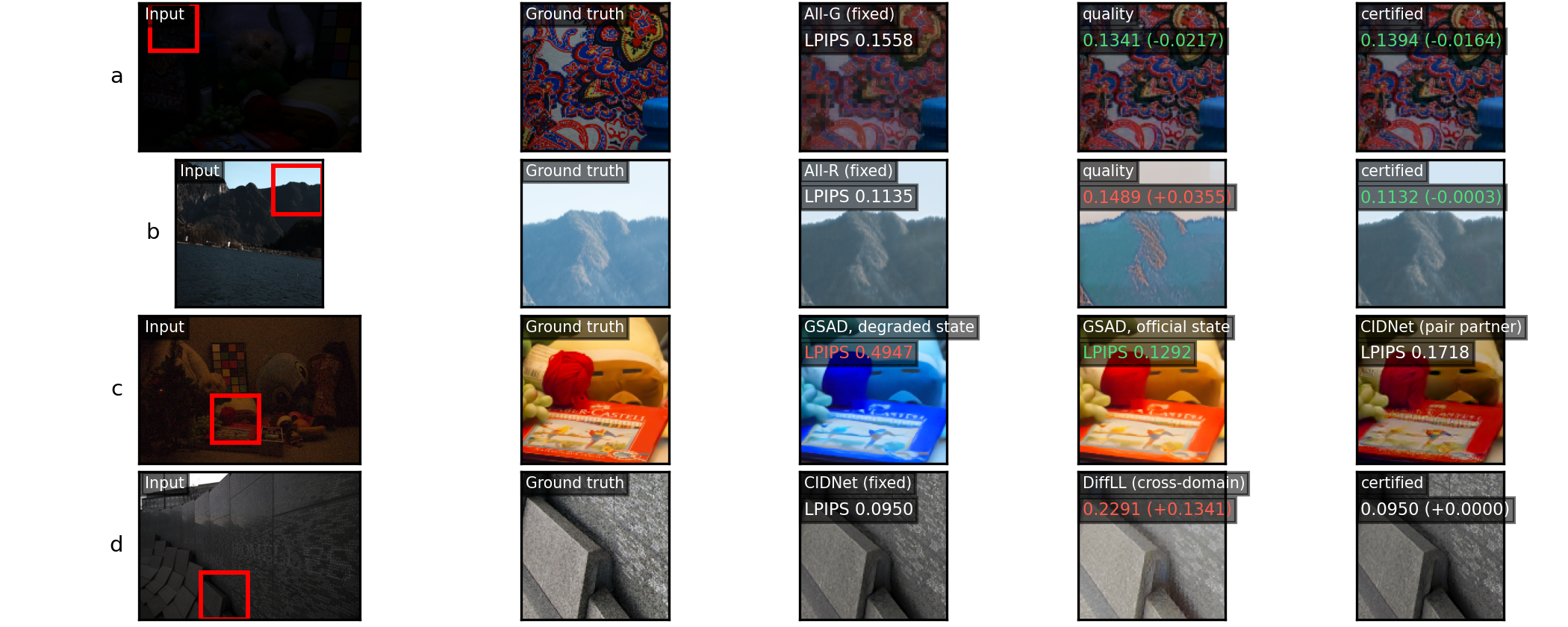}
\caption{Qualitative examples (red boxes indicate the displayed crops; stochastic experts are shown at seed~1). (a) LOL-v1: the certificate preserves useful refinement. (b) LOLv2-syn: the quality mode hallucinates structure and color, while the certificate vetoes the harmful override. (c) GSAD process-state bistability: the same input and weights produce a degraded state and the published state. (d) SICE: the certificate admits no override, so the output matches the fixed CIDNet expert.}
\label{cases}
\end{figure*}

\paragraph{Expert reproducibility is part of the audit.}
In the experiment, we found that GSAD~\cite{gsad} could enter two different process-dependent output states despite identical inputs, weights, and RNG states: the published state achieves LPIPS $\approx0.11$ on LOLv2-real, whereas the degraded state gives $\approx0.26$--$0.30$. Our original first-image self-check did not detect this failure because the state could change later during batch inference. A subsequent comparison against the official evaluator revealed that our stored batch outputs were indeed in the degraded state (mean LPIPS 0.259 vs.\ 0.113 on the same 100 images).

We verified that the difference was not caused by our evaluation harness: a deterministic HVI-CIDNet negative control differed from its official evaluator by only $7{\times}10^{-5}$ mean LPIPS. We therefore regenerated GSAD using the verified official process configuration and replaced the single-image check with a batch-level distributional comparison against the official evaluator. The full regeneration audit, gate criteria, and amendment history are provided in the supplementary. Thus, expert-level reproducibility checks should be considered as part of an auditable evaluation protocol.

\paragraph{Cross-domain terminal test: exact abstention.}
To test the protocol beyond the LOL family, we pre-registered a complete setting on SICE~\cite{sice} Part 2, using 229 sequences split into 40/104/45/40 for selection, calibration, validation, and test. Pair selection used only the 40 selection images and chose CIDNet+DiffLL, which showed 9.2\% whole-image complementarity, while CIDNet+GSAD showed almost none. The selected pair retained oracle headroom ($0.0177$ at a 20\% area budget), and the router outperformed all four trivial cues (73–80\%). However, at $\alpha=0.10$, no safe non-trivial threshold could be certified, so the certified output was identical to the fixed CIDNet expert on validation (harm 0/45). The sealed one-shot test reproduced the same behavior: the certified output was identical to CIDNet on all 40 test images, with zero routed area and harm $0/40$ (Fig.~\ref{cases}d). We therefore report SICE as an abstention replication, rather than an enhancement gain.

%% file: sec/7_discussion.tex
\section{Discussion and Limitations}
\label{sec:discussion}
\paragraph{Evaluation conventions and comparison scope.}
For context, same-pipeline comparisons with representative LLIE methods are reported in the supplementary using a single evaluation stack and without GT-mean brightness correction for any method. Published results that use different brightness-correction conventions are listed separately and are not mixed into the same comparison. These comparisons are provided for context rather than as a state-of-the-art claim; our contribution is the certification and audit protocol, not the experts themselves.

\paragraph{Scope of the guarantee.}
The certificate controls harm marginally over exchangeable deployment images. Thus, it is not a per-image or intervention-conditional guarantee: an individual routed output may still be worse than the fixed base expert. The guarantee also relies on deployment remaining sufficiently close to the calibrated regime; under substantial distribution shift, it no longer formally applies. Our stress tests characterize this boundary empirically, but the guard can only detect some forms of shift. In addition, harm is defined using LPIPS against paired ground truth. The certificate therefore controls perceptual degradation under this metric, not semantic hallucination.

\paragraph{Finite-sample certification.}
The available LLIE benchmarks provide relatively small calibration and test sets. For example, the main certificate uses only $n=30$ calibration images, while some final test sets are also small. Harm-rate estimates therefore have substantial statistical uncertainty, and a small number of harmful calibration examples can make the certificate conservative or force it to abstain. Larger calibration sets would allow more precise and potentially less conservative deployment decisions.

\paragraph{Magnitude of the gains.}
The certified improvements are modest when averaged over the full test set. This is partly expected because the certified policy modifies only a small fraction of images and abstains whenever the evidence for a safe improvement is insufficient. We therefore view the main contribution as reliable selective intervention under an explicit harm budget, rather than large average metric gains. Whether these small LPIPS improvements are consistently perceptible to human observers remains a question and would require a larger dedicated perceptual study.

\paragraph{Dependence on expert complementarity.}
VetoNet is useful only when the two frozen experts provide complementary outputs. If one expert consistently dominates the other, there is little useful routing opportunity, and the certified policy reduces to the base expert. Thus, the framework does not create complementarity; it identifies when available complementarity can be exploited without exceeding the prescribed harm budget. The current implementation also uses hard regional replacement and requires running both experts, which increases inference cost. Future work could extend the same risk-control framework to softer blending strategies and more efficient expert selection.

%% file: sec/8_conclusion.tex
\section{Conclusion}
\label{sec:conclusion}

Selective generative enhancement can be evaluated counterfactually, routed adaptively, and controlled with a certified veto. Across expert pairs and domains, the proposed protocol adapted to the available evidence: it
produced small but consistent gains when the experts were complementary, reversed the routing direction when their relative strengths changed, and abstained when no safe override could be certified. Across all four selection-independent pre-registered final tests, the certified mode remained within its nominal harm budget; the disclosed modern-pair audit also showed zero empirical test harm. The certified gains are modest, and abstention is frequent, but this is also the central finding: a selective enhancement system should be allowed to refuse intervention when the evidence is insufficient.